\RequirePackage{fix-cm} %
\documentclass[10pt]{article} %
\usepackage[accepted]{tmlr}

\usepackage{amsmath,amsfonts,bm}

\def\eqref#1{equation~\ref{#1}}

\def\1{\bm{1}}

\def\vzero{{\bm{0}}}

\def\vtheta{{\bm{\theta}}}

\def\vb{{\bm{b}}}

\def\vh{{\bm{h}}}

\def\vr{{\bm{r}}}
\def\vs{{\bm{s}}}

\def\vx{{\bm{x}}}

\def\vz{{\bm{z}}}

\def\mA{{\bm{A}}}

\def\mH{{\bm{H}}}
\def\mI{{\bm{I}}}

\def\mQ{{\bm{Q}}}

\def\mU{{\bm{U}}}

\def\mW{{\bm{W}}}
\def\mX{{\bm{X}}}

\DeclareMathAlphabet{\mathsfit}{\encodingdefault}{\sfdefault}{m}{sl}
\SetMathAlphabet{\mathsfit}{bold}{\encodingdefault}{\sfdefault}{bx}{n}

\def\gD{{\mathcal{D}}}
\def\gE{{\mathcal{E}}}

\def\gL{{\mathcal{L}}}

\def\gN{{\mathcal{N}}}

\newcommand{\E}{\mathbb{E}}

\newcommand{\R}{\mathbb{R}}

\newcommand{\KL}{D_{\mathrm{KL}}}

\usepackage{bold-extra}

\usepackage{amsmath}
\usepackage{amssymb}
\usepackage{mathtools}
\usepackage{amsthm}

\definecolor{mydarkblue}{rgb}{0,0.08,0.45}
\usepackage[colorlinks,citecolor=mydarkblue,urlcolor=mydarkblue,linkcolor=mydarkblue]{hyperref}
\usepackage{url}

\usepackage[capitalize,noabbrev]{cleveref}
\crefformat{equation}{(#2#1#3)}

\usepackage{caption} %
\usepackage{subcaption} %
\usepackage{sidecap} %
\sidecaptionvpos{figure}{t}
\usepackage{graphicx} %
\usepackage{booktabs} %
\usepackage{layouts} %
\usepackage{multirow} %
\usepackage{marvosym} %
\usepackage{pdflscape} %
\usepackage{wrapfig} %
\usepackage{placeins} %
\usepackage{nth} %

\theoremstyle{plain}

\theoremstyle{definition}

\theoremstyle{remark}

\usepackage{tikz} %
\usetikzlibrary{arrows, chains, positioning, decorations.pathreplacing, fit, calc, decorations.pathmorphing, patterns}
\usepackage{scalerel} %
\usepackage[outline]{contour} %

\newcommand{\circlemarker}[2][75]{\scalerel*{\tikz \draw[#2, fill=#2!#1, line width=0.4pt] (0,0) circle (2pt);}{\circ}}

\newcommand{\diamondmarker}[2][75]{\scalerel*{\tikz \draw[#2, fill=#2!#1, rounded corners=0.02pt, line width=0.4pt] (-3pt,0)--++(45:3pt)--++(-45:3pt)--++(-180+45:3pt)--cycle;}{\square}}

\newcommand{\triangleleftmarker}[2][75]{\scalerel*{\tikz \draw[#2, fill=#2!#1, rounded corners=0.02pt, line width=0.4pt] (0,-2.5pt)--++(0,5pt)--++(-180+30:5pt)--cycle;}{\circ}}

\definecolor{mplblue}{rgb}{0.12156862745098039, 0.4666666666666667, 0.7058823529411765}
\definecolor{mplorange}{rgb}{1.0, 0.4980392156862745, 0.054901960784313725}
\definecolor{mplgreen}{rgb}{0.17254901960784313, 0.6274509803921569, 0.17254901960784313}
\definecolor{mplred}{rgb}{0.8392156862745098, 0.15294117647058825, 0.1568627450980392}
\definecolor{mplpurple}{rgb}{0.5803921568627451, 0.403921568627451, 0.7411764705882353}
\definecolor{mplbrown}{rgb}{0.5490196078431373, 0.33725490196078434, 0.29411764705882354}
\definecolor{mplpink}{rgb}{0.8901960784313725, 0.4666666666666667, 0.7607843137254902}
\definecolor{mplgrey}{rgb}{0.4980392156862745, 0.4980392156862745, 0.4980392156862745}
\definecolor{mplyellow}{rgb}{0.7372549019607844, 0.7411764705882353, 0.13333333333333333}
\definecolor{mplcyan}{rgb}{0.09019607843137255, 0.7450980392156863, 0.8117647058823529}

\definecolor{viridisyellow}{HTML}{EED809}
\definecolor{viridispurple}{HTML}{440154}

\newcommand{\dark}[1]{#1!50!black}
\newcommand{\light}[1]{#1!50!white}
\tikzset{tile1/.style={draw=\dark{#1}, fill=\light{#1}}}
\tikzset{tile2/.style={preaction={fill=\light{#1}}, draw=\dark{#1}, pattern=north east lines, pattern color=#1!95!black}}
\tikzset{mlpout/.style={}}
\tikzset{softmaxin/.style={double}}

\colorlet{color1}{mplblue}
\colorlet{color2}{mplorange}
\colorlet{color3}{mplgreen}

\tikzset{patch/.style={rectangle, minimum width=3.75ex, minimum height=3.75ex, inner sep=-0.5ex}}
\tikzset{smallpatch/.style={rectangle, minimum width=1.0ex, minimum height=1.0ex}}
\tikzset{tinypatch/.style={rectangle, minimum width=1.ex, minimum height=1.ex, inner sep=0.ex}}
\tikzset{bar/.style={rectangle, minimum width=0.8ex, inner sep=0}}
\tikzset{op/.style={rectangle, draw, minimum height=3.ex, inner sep=0.375ex}}

\newcommand{\patchmarker}[1][mplgrey]{\scalerel*{\tikz \node[smallpatch, tile1=#1] {};}{\square}}
\newcommand{\tiledpatchmarker}[1][mplgrey]{\scalerel*{\tikz \node[smallpatch, tile2=#1] {};}{\square}}
\newcommand{\doublepatchmarker}[1][mplgrey]{\scalerel*{\tikz \node[smallpatch, double, tile1=#1] {};}{\square}}

\tikzset{line/.style={draw=gray!90!black}}
\tikzset{arrow/.style={-stealth, line}}
\tikzset{part1/.style={}}
\tikzset{part2/.style={}}

\usepackage{xspace}
\newcommand{\ack}{\textsc{e}$^3$\xspace}

\newcommand{\mname}{efficient ensemble of experts\xspace}
\newcommand{\MNAME}{Efficient Ensemble of Experts\xspace}
\newcommand{\MName}{Efficient ensemble of experts\xspace}

\def\rvvarepsilon{{\mathbf{\varepsilon}}}

\rmfamily %
\DeclareFontShape{T1}{lmr}{b}{sc}{<->ssub*cmr/bx/sc}{}
\DeclareFontShape{T1}{lmr}{bx}{sc}{<->ssub*cmr/bx/sc}{}

\title{Sparse MoEs meet Efficient Ensembles}

\author{\name James Urquhart Allingham$^{1,}$\thanks{Work done as a Google Research intern.} \email jua23@cam.ac.uk
      \AND
      \name Florian Wenzel$^{2,}$\thanks{Work done while at Google Research.} \email fln.wenzel@gmail.com
      \AND
      \name Zelda E Mariet, Basil Mustafa \email \{zmariet,basilm\}@google.com
      \AND
      \name Joan Puigcerver, Neil Houlsby \email \{jpuigcerver,neilhoulsby\}@google.com
      \AND
      \name Ghassen Jerfel$^{3,}$\footnotemark[2] \email ghassen@google.com 
      \AND
      \name Vincent Fortuin$^{1,4,}$\footnotemark[1] \email vbf21@cam.ac.uk
      \AND
      \name Balaji Lakshminarayanan, Jasper Snoek \email \{balajiln,jsnoek\}@google.com
      \AND
      \name Dustin Tran, Carlos Riquelme, Rodolphe Jenatton \email \{trandustin,rikel,rjenatton\}@google.com
      \AND
      \addr Google Research, Brain Team; $^1$University of Cambridge; $^2$no affiliation; $^3$Waymo; $^4$ETH Zürich
      }

\def\month{09}  %
\def\year{2022} %
\def\openreview{\url{https://openreview.net/forum?id=i0ZM36d2qU}} %

\begin{document}

\maketitle

\begin{abstract}
Machine learning models based on the aggregated outputs of submodels, either at the activation or prediction levels, %
often exhibit strong performance compared to individual models.
We study the interplay of two popular classes of such models: ensembles of neural networks and sparse mixture of experts (sparse MoEs).
First, we show that the two approaches have complementary features whose combination is beneficial.
This includes a comprehensive evaluation of sparse MoEs in uncertainty related benchmarks.
Then, we present \emph{\mname} (\ack), a scalable and simple ensemble of sparse MoEs that takes the best of both classes of models, while using up to 45\% fewer FLOPs than a deep ensemble.
Extensive experiments demonstrate the accuracy, log-likelihood, few-shot learning, robustness, and uncertainty improvements of \ack over several challenging vision Transformer-based baselines.
\ack not only preserves its efficiency while scaling to models with up to 2.7B parameters, but also provides better predictive performance and uncertainty estimates~for~larger~models.
\end{abstract}

\section{Introduction}

Neural networks (NNs) typically use all their parameters to process an input.
Sustaining the growth of such models---reaching today up to 100B+ parameters~\citep{brown2020language}---is challenging, e.g., due to their high computational and environmental costs~\citep{strubell2019energy, patterson2021carbon}.
In this context, sparse mixtures of experts (sparse MoEs) employ \textit{conditional computation}~\citep{bengio2013estimating} to combine multiple submodels %
and route examples to %
specific ``expert'' submodels \citep{shazeer2017outrageously,lepikhin2020gshard,fedus2021switch,riquelme2021scaling}. Conditional computation can decouple the growth of the number of parameters from the training and inference costs, by only activating a subset of the overall model in an input-dependent fashion.

Paralleling this trend, the deployment of ML systems in safety-critical fields, e.g., medical diagnosis~\citep{dusenberry2020analyzing} and self-driving cars~\citep{levinson2011towards}, has motivated the development of reliable deep learning, e.g., for \emph{calibrated and robust predictions} \citep{snoek2019can}. Among the existing approaches, ensembles of NNs have remarkable performance for
calibration and accuracy
under dataset shifts~\citep{snoek2019can}.
These methods 
improve reliability by aggregating the predictions of individual submodels, referred to as ensemble members.
However, this improvement comes at a significant computational cost. Hence, naively ensembling NNs that continue to grow in size becomes less and less feasible.
In this work, we try to overcome this limitation. Our core motivation is to improve the robustness and uncertainty estimates of large-scale fine-tuned models through ensembling, but to do so in a tractable---and thus practically useful---manner, by carefully developing a hybrid approach using advances in sparse MoEs.

While sharing conceptual similarities, these two classes of models---MoEs and ensembles---have different
properties.
Sparse MoEs adaptively combine their experts depending on the inputs, and the combination generally happens at internal activation levels.
Ensembles typically combine several models in a static way and at the prediction level. Moreover, these two classes of models tend to be benchmarked on different tasks: few-shot classification for MoEs~\citep{riquelme2021scaling} and 
uncertainty-related evaluation for ensembles~\citep{snoek2019can, gustafsson2019evaluating}. 
For example, sparse MoEs are seldom, if ever, applied to the problems of calibration.

\begin{figure*}[t]
    \vspace{-2.em}
    \centering
    \resizebox{\textwidth}{!}{
    \begin{tikzpicture}[
        every node/.append style={font=\tiny}
    ]
    \input{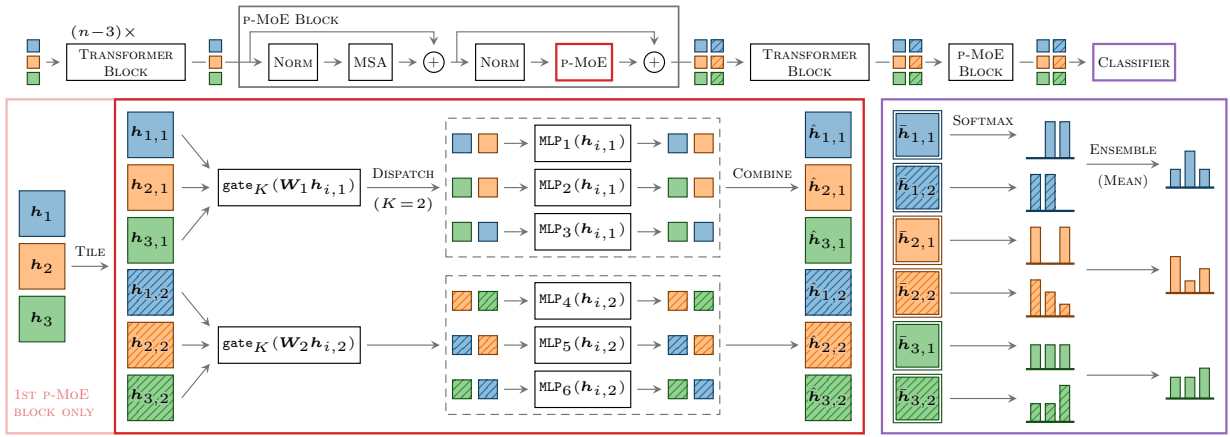}
    \end{tikzpicture}
    }
    \vspace{-1.em}
    \caption{
    End-to-end overview of \ack with $E\!=\!6$ experts, partitioned into $M\!=\!2$ groups, with sparsity of $K\!=\!2$, and a ``last-2'' configuration. \textbf{Top}: \ack contains a sequence of transformer blocks, followed by alternating transformer and p(artitioned)-MoE blocks. As in ViT, images are split into patches whose embeddings are processed by each block. Here, we show 1 embedding for each of three images (\patchmarker[color1], \patchmarker[color2], \patchmarker[color3]). \textcolor{mplred}{\textbf{Bottom left}}: In a p-MoE block, we replace the transformer block's MLP with parallel partitioned expert MLPs, see \cref{eq:gating_pbatch_ensemble}.
    The effect of the routing weights is not depicted.
    Embeddings are tiled (\tiledpatchmarker) in the first p-MoE block only. \textcolor{mplpurple}{\textbf{Bottom right}}: The classifier averages predictions from the final tiled representations (\doublepatchmarker).
    }
    \label{fig:E3_diagram}
    \vspace{-1.em}
\end{figure*}

Here, we study the interplay between sparse MoEs and ensembles. This results in two sets of contributions:

\textbf{Contribution 1: Complementarity of sparse MoEs and ensembles.} We show that sparse MoEs and ensembles have complementary features and benefit from each other.
Specifically:
\begin{itemize}
\item The adaptive computation in sparse MoEs and the static combination in ensembles are orthogonal, with additive benefits when associated together.
At the intersection of these two model families is an exciting trade-off between performance and compute (FLOPs). That is, the frontier can be mapped out by varying the ensemble size and the sparsity of MoEs.
\item Over tasks where either sparse MoEs or ensembles are known to perform well, naive---and computationally expensive---ensembles of MoEs
provide the best predictive performance.
Our benchmarking effort includes the first evaluation of sparse MoEs on uncertainty-related vision tasks, which builds upon the work of \citet{riquelme2021scaling}.
\end{itemize}

\textbf{Contribution 2: \MName.} We propose \MNAME (\ack), see \cref{fig:E3_diagram}, an efficient ensemble approach tailored to sparse MoEs:
\begin{itemize}
\item \ack improves over sparse MoEs across few-shot error, likelihood and calibration error. \ack matches the performance of deep ensembles 
while using from 30\% to 45\% fewer FLOPs.
\item \ack gracefully scales up to 2.7B parameter models.
\item \ack is both simple---requiring only minor implementation changes---and
convenient---\ack models can be fine-tuned directly from standard sparse-MoE checkpoints. Code can be found at \url{https://github.com/google-research/vmoe}.
\end{itemize}

\section{Preliminaries}
We focus on classification tasks where we learn classifiers of the form $f(\vx; \vtheta)$ based on some training data $\gD = \{(\vx_n, y_n)\}_{n=1}^N$. A pair $(\vx_n, y_n)$ corresponds to an input $\vx_n \in \R^P$ together with its label $y_n \in \{1,\dots, C\}$ belonging to one of the $C$ classes. The model $f(\cdot; \vtheta)$ is parametrized 
by $\vtheta$ and
outputs a $C$-dimensional probability vector.
We use $\circ$ to refer to the matrix element-wise product.

\subsection{Vision Transformers and Sparse MoEs}\label{sec:vit_and_vmoe}
\paragraph{Vision Transformers.} Throughout the paper, we choose the model $f$ to be a vision Transformer (ViT)~\citep{dosovitskiy2020image}.
ViT is growing in popularity for vision, especially in transfer-learning settings where it was shown to outperform convolutional networks while requiring fewer pre-training resources. 
ViT operates at the level of patches. An input image is split into equal-sized patches (e.g., $32\times32$, $16\times16$, or $14\times14$ pixels) whose resulting sequence is (linearly) embedded and processed by a Transformer~\citep{vaswani2017attention}. The operations in the Transformer then mostly consist of a succession of multiheaded self-attention (MSA) and MLP layers.
ViT is defined at different scales~\citep{dosovitskiy2020image}: S(mall), B(ase), L(arge) and H(uge); see specifications in~\Cref{sec:vit_model_specifications}. For example, ViT-L/16 stands for a large ViT with patch size $16 \times 16$.

\paragraph{Sparse MoEs and V-MoEs.} The main feature of sparsely-gated mixture-of-experts models (sparse MoEs) lies in the joint use of sparsity and \textit{conditional computation}~\citep{bengio2013estimating}. In those models, we only activate a small subset of the network parameters \textit{for a given input}, which allows the total number of parameters $\vtheta$ to grow while keeping the overall computational cost constant.
The \textit{experts}  are the subparts of the network activated on a per-input fashion.

Central to our study, \citet{riquelme2021scaling} recently extended ViT to sparse MoEs. Their extension, referred to as V-MoE, follows the successful applications of sparse models in NLP~\citep{shazeer2017outrageously}. \citet{riquelme2021scaling} show that V-MoEs dominate their ``dense'' ViT counterparts on a variety of tasks for the same computational cost. 
In the specific case of V-MoEs, the experts are placed in the MLP layers of the Transformer, a design choice reminiscent of~\citet{lepikhin2020gshard} in NLP. Given the input $\vh \in \R^D$ of such a layer, the output of a single $\texttt{MLP}(\vh)$ is replaced by
\begin{flalign}\label{eq:gating}
    &\texttt{MoE}(\vh) = \sum_{e=1}^E g_e(\vh) \cdot \texttt{MLP}_e(\vh) \ \ \text{with} \ \
    \{g_e(\vh)\}_{e=1}^E = \texttt{top}_K(\texttt{softmax}(\mW \vh)),
\end{flalign}
where the \textit{routing} weights $\{g_e(\vh)\}_{e=1}^E$ combine the outputs of the $E$ different experts  $\{\texttt{MLP}_e\}_{e=1}^E$. To sparsely select the experts, $\texttt{top}_K$ sets all but the $K$ largest weights to zero.
The router parameters $\mW \in \R^{E \times D}$ are trained together with the rest of the network parameters. We call the layer defined by~\cref{eq:gating} an MoE layer. In practice, the weights $\{g_e(\vh)\}_{e=1}^E$ are obtained by a noisy version of the routing function $\texttt{top}_K(\texttt{softmax}(\mW \vh + \sigma \rvvarepsilon))$ with $\rvvarepsilon \sim \gN(\vzero, \mI)$, which 
mitigates the non-differentiability of $\texttt{top}_K$ when combined with auxiliary losses~(see Appendix A in \citet{shazeer2017outrageously}).
Making non-differentiable operators smooth with some noise injection is an active area of research \citep{berthet2020learning,abernethy2016perturbation,duchi2012randomized}.
We use the shorthand 
$\texttt{gate}_K(\vz) = \texttt{top}_K(\texttt{softmax}(\vz + \sigma \rvvarepsilon))$ and take $\sigma = 1/E$ as in \citet{riquelme2021scaling}.

In this paper, we consider the ``last-$n$'' setting of \citet{riquelme2021scaling} wherein only a few MoE layers are placed at the end of the Transformer ($n=2$ for the \{S, B, L\} scale and $n=5$ for H). This setting retains most of the performance gains of V-MoEs while greatly reducing the training cost.

\subsection{Ensembles of Neural Networks}
\paragraph{Ensembles.} 
We build on the idea of ensembles, which is a known scheme to improve the performance of individual models~\citep{hansen1990neural,geman1992neural, krogh1995neural,opitz1999popular,dietterich2000ensemble,Lakshminarayanan2017}. 
Formally, we assume a set of $M$ model parameters $\Theta = \{\vtheta_m\}_{m=1}^M$. We refer to $M$ as the \textit{ensemble size}. Prediction proceeds by computing $\frac{1}{M} \sum_{\vtheta \in \Theta} f(\vx; \vtheta)$, i.e., the average probability vector over the $M$ models.
To assess the diversity of the predictions in the ensemble, we will use the KL divergence $\KL(f(\vx_t;\vtheta_m)\|f(\vx_t;\vtheta_{m'}))$ between the predictive distributions $f(\vx_t;\vtheta_m)$ and $f(\vx_t;\vtheta_{m'})$, averaged over the test input $\vx_t$ and all pairs $(m, m')$ of ensemble members.

\paragraph{Batch ensembles.} 
\citet{wen2020batchensemble} construct a \emph{batch ensemble} (BE) as a collection of submodels, with the parameters $\vtheta_m \in \Theta$ sharing components.
This mitigates the computational and memory cost of ensembling, while still improving performance.
We focus on the example of a single dense layer in $f$ with parameters $\mU \in \R^{D \times L}$, assuming no bias. BE defines $M$ copies of parameters $\{\mU_m\}_{m=1}^M $ so that
$\mU_m = \mU \circ (\vr_m \vs_m^\top)$,
where $\mU$ are parameters shared across ensemble members, and $\vr_m$ and $\vs_m$ are separate $D$- and $L$-dimensional vectors for ensemble member $m$. Given an input, BE produces $M$ outputs, which are averaged after applying all layers.
Despite the simple rank-1 parametrization, BE leads to remarkable predictive performance and robustness~\citep{wen2020batchensemble}.
Notably, the efficiency of BE relies on both the parameter sharing and the tiling of the inputs to %
predict with the $M$ ensemble members, two insights that we exploit in our paper.

\subsection{Pre-training and Fine-tuning}
Large-scale Transformers pre-trained on \textit{upstream} tasks were shown to have strong performance when fine-tuned on smaller \textit{downstream} tasks, across a variety of domains~\citep{devlin2018bert, dosovitskiy2020image, radford2021learning}. We follow this paradigm and focus on the fine-tuning of models pre-trained on JFT-300M~\citep{sun2017revisiting}, similar to \citet{riquelme2021scaling}. We will thus assume the availability of already pre-trained ViT and V-MoE model checkpoints.
Our assumption relies on the growing popularity of transfer learning, e.g.~\citet{kolesnikov2020big}, and the increasing accessibility of pre-trained models in repositories such as 
\href{https://www.tensorflow.org/hub}{\url{www.tensorflow.org/hub}} or \href{https://www.pytorch.org/hub}{\url{www.pytorch.org/hub}}.
The fine-tuning of all the approaches we study here, including extensions of ViT and V-MoE, will be either directly compatible with those checkpoints or require only mild adjustments, e.g., reshaping or introducing new downstream-specific parameters (see \cref{sec:upstream_ckpt_compatibility}). Also, unless otherwise mentioned, the performance we report will always be downstream, e.g., for ImageNet~\citep{deng2009imagenet} or CIFAR10/100~\citep{krizhevsky2009learning}.
In all our comparisons, we will use the downstream training floating point operations per second (FLOPs), or GFLOPs (i.e.,~$10^9\times$FLOPs), to quantify the computational cost of the different methods.

\begin{table*}[tbp]
\vspace{-2.em}
\caption{
Overview of key properties of sparse MoEs, ensembles, and \ack. \ack acheives the best of both worlds. \texttt{dense} is a base model upon which we add the sparse MoE or ensemble logic, e.g., a ViT model in this paper.
}
\vspace{-1em}
\label{tab:sparse_moes_vs_ensembles}
\begin{center}
\resizebox{\textwidth}{!}{%
\begin{tabular}{ccccc}
\toprule
\multicolumn{1}{c}{}  & \multicolumn{1}{c}{\textsc{Predictions}}  & \multicolumn{1}{c}{\textsc{Combinations}} & \multicolumn{1}{c}{\textsc{Conditional Computation}} & \multicolumn{1}{c}{\textsc{Cost}}
\\ \midrule
{\bf Sparse MoEs} & Single & Activation level &  Yes, adaptively per-input & $\approx$ \texttt{dense}  \\
{\bf Ensembles}   & Multiple & Prediction level & No, static & $>$ \texttt{dense} \\ \cmidrule{1-5}
\textbf{\ack} & Multiple & Activation \& prediction level & Yes, adaptively per-input & $\approx$ \texttt{dense} \\
\bottomrule
\end{tabular}%
}
\end{center}
\vspace{-1.em}
\end{table*}

\section{Sparse MoEs meet Ensembles}\label{sec:sparse_moes_meet_efficient_ensembles}

As illustrated in \cref{tab:sparse_moes_vs_ensembles}, sparse MoEs and ensembles have different properties. For instance, ensembles typically do not use conditional computation and just statically combine members at the prediction level. This contrasts with sparse MoEs where the different experts are combined at internal activation levels while enjoying per-input adaptivity through the routing logic, see~\cref{eq:gating}. 
In terms of cost, sparse MoEs are usually designed to match the inference time of their dense counterparts whereas ensembles, in their simplest forms, will typically lead to a substantial overhead.  
In this section, we study the extent to which these properties are complementary and may benefit from each other. In \cref{sec:experiments}, we further evaluate this complementarity on tasks where either sparse MoEs or ensembles are known to perform well, e.g., few-shot and out-of-distribution (OOD) evaluations, respectively.

\begin{figure*}[tbp]
    \vspace{-2.em}
     \centering
     \begin{subfigure}[b]{0.32\textwidth}
        \centering
        \includegraphics[width=0.85\textwidth]{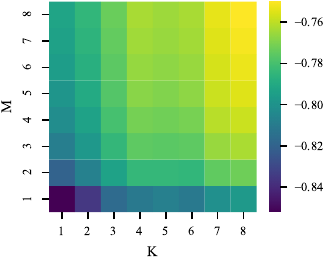}
        \caption{Log-Likelihood (LL) \textcolor{white}{$\left[\frac{\text{I}_{(}}{\text{I}_{(}}\right]$}}
        \label{fig:ll_heatmap}
     \end{subfigure}
     \hfill
     \begin{subfigure}[b]{0.32\textwidth}
        \centering
        \includegraphics[width=0.85\textwidth]{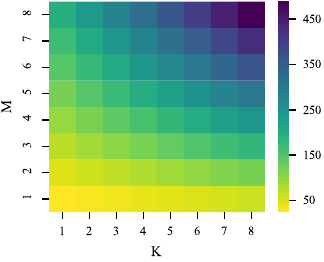}
        \caption{(downstream) GFLOPs \textcolor{white}{$\left[\frac{\text{I}_{(}}{\text{I}_{(}}\right]$}}
        \label{fig:flops_heatmap}
     \end{subfigure}
     \hfill
     \begin{subfigure}[b]{0.32\textwidth}
        \centering
        \includegraphics[width=0.85\textwidth]{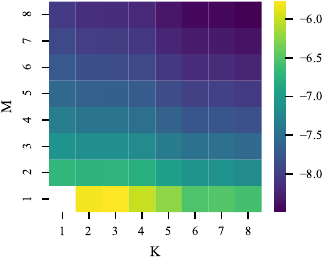}
        \caption{
        $\text{log}\!\left[\!\frac{\text{LL}_{(K,M)} - \text{LL}_{(1,1)}}{\text{GFLOPs}_{(K,M)} - \text{GFLOPs}_{(1,1)}}\!\right]$
        }
        \label{fig:norm_ll_heatmap}
     \end{subfigure}
    \vspace{-1.em}
\caption{Increasing static ($M$) and adaptive ($K$) ensembling on ImageNet, for V-MoE-S/32.
\textbf{\textcolor{viridisyellow}{Yellow}}/\textbf{\textcolor{viridispurple}{purple}} indicates better/worse performance. Increasing \emph{both} static and adaptive ensembling is beneficial, the latter being more efficient.
}
\vspace{-1.em}
\label{fig:static-vs-adaptive}
\end{figure*}

To investigate the interactions of the properties in \cref{tab:sparse_moes_vs_ensembles}, we study the performance of \textit{downstream} deep ensembles (i.e., with all ensemble members having the same upstream checkpoint) formed by $M$ independent V-MoEs with $E$ experts per MoE layer and a sparsity level $K$ (the larger $K$, the more selected experts). %
$M$  controls the static combination, while $K$ and $E$ impact the adaptive combination of experts in each sparse MoE model.
We report in \cref{fig:static-vs-adaptive} the ImageNet performance and compute cost for ensembles with varying choices of $K$ and $M$, while keeping $E=32$ fixed. 
We focus on $K$ rather than $E$ to explore adaptive computation, as we found the performance quickly plateaus with $E$
(see \cref{fig:e_k_m_plot} in the Appendix). Also, by fixing $E=32$, we match more closely the 
setup of \citet{riquelme2021scaling}. The architecture of the V-MoE is ViT-S/32, see details in \cref{sec:sparse_moes_meet_efficient_ensembles_appendix}.
We make the following observations:

\paragraph{Investigating the cumulative effects of adaptive and static ensembling.} In the absence of ensembles (i.e., when we consider $M=1$), and given a fixed number of experts,
the authors of 
\citet{riquelme2021scaling} already reported an increase in performance as $K$ gets larger.~Interestingly, we observe that for each value of $K$, it is also beneficial to increase the ensemble size $M$. 
\begin{wrapfigure}[21]{r}{0.333\textwidth}
  \vspace{-0.5em}
\begin{center}
\includegraphics[width=0.3\textwidth]{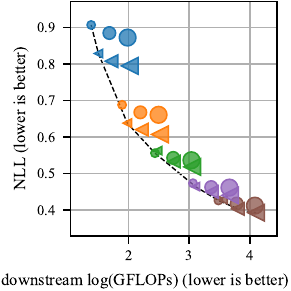}
\end{center}
\caption{
ImageNet evaluation for ViT~(\circlemarker{black}) and V-MoE~($K\!\!=\!\!1$)~(\triangleleftmarker{black}) ensembles of size $\{1,\,2,\,4\}$ ({\tiny\diamondmarker{black}},\,\diamondmarker{black},\,{\Large \diamondmarker{black}}). Model sizes:~S/32~(\diamondmarker{mplblue}), B/32~(\diamondmarker{mplorange}),~L/32~(\diamondmarker{mplgreen}), L/16~(\diamondmarker{mplpurple}),~and~H/14~(\diamondmarker{mplbrown}). ViT and V-MoE benefit from ensembling equally, at all scales.
}
\label{fig:vit_vmoe_ens}
\end{wrapfigure}
In other words, the static combination of ensembles is beneficial when applied to sparse MoEs.
This observation is perhaps surprising
since adaptive combination may already encapsulate the effect of static combination.
\cref{fig:vit_vmoe_ens}, and \cref{sec:sva_part2}, show that the combination of static ensembling and adaptivity is beneficial to NLL for a range of ViT families.
We also see that the benefits of static ensembling are similar for V-MoE and ViT (which does not have any adaptivity).

\paragraph{Investigating ensembles of sparse MoEs with fewer experts.}
In \cref{sec:more_motivation}, we compare the performance of a V-MoE with $E=32$ experts and ensembles of V-MoEs with fewer experts, namely ($M=2$, $E=16$) and ($M=4$, $E=8$). We see that the performance---e.g., as measured by NLL---is better for ($M=4$, $E=8$) than ($M=2$, $E=16$) which is in turn better than ($M=1$, $E=32$). Thus, we conclude that \emph{reducing the number of experts only mildly affects the combination of adaptive and static ensembling.}

\paragraph{Investigating the trade-off between FLOPs and performance.} Without any computational constraints, the previous observation would favor approaches with the largest values of $K$ and $M$.
However, different values of $(K, M)$ lead to different computational costs, as measured here by FLOPs, with $(K, M)=(1, 1)$ being the cheapest.
\cref{fig:flops_heatmap} shows, as expected, that the number of FLOPs grows more quickly along the $M$ axis than along the $K$ axis.
To capture the various trade-offs at play, in \cref{fig:norm_ll_heatmap} we report the logarithm of the normalized gains in log likelihood
$\frac{\text{LL}_{(K,M)} - \text{LL}_{(1,1)}}{\text{GFLOPs}_{(K,M)} - \text{GFLOPs}_{(1,1)}}$ 
when going from $(K, M)=(1, 1)$ to other choices of $(K, M)$. Interestingly, it appears more advantageous to first grow $K$, i.e., the adaptive combination, before growing $M$.

\paragraph{Summary.} While simple ensembling of sparse MoEs results in strong predictive performance, we lose their computational efficiency. 
We next show how to efficiently combine ensembling and sparse MoEs, exploiting the fact that statically combining sparse MoEs with fewer experts remains effective.

\section{\MNAME} \label{sec:method}

Equipped with the insights from \cref{sec:sparse_moes_meet_efficient_ensembles}, we describe \mname (\ack), with the goal of keeping the strengths of both sparse MoEs and ensembles. Conceptually, %
\ack %
jointly %
learns an ensemble of smaller sparse MoEs, 
where all layers without experts (e.g., attention layers) are shared across the members.

\subsection{The Architecture}\label{sec:E3_architecture}
There are two main components in \ack:

\textbf{Disjoint subsets of experts.
}
We change the structure of~\cref{eq:gating} by \textit{partitioning} the set of $E$ experts into $M$ subsets of $E/M$ experts (%
assuming that $E$ is a multiple of $M$). 
We denote the subsets by $\gE_m$. For example, $\gE_1 = \{1, 2, 3\}$ and $\gE_2=\{4, 5, 6\}$ for $E=6$ and $M=2$. 
Intuitively, the ensemble members have separate parameters for independent predictions, while efficiently sharing parameters among all non-expert layers.
Instead of having a single routing function $\texttt{gate}_K(\mW \cdot)$ as in~\cref{eq:gating}, we apply separate routing functions $\texttt{gate}_K(\mW_m \cdot)$ to each subset $\gE_m$.
Note that this does not affect the total number of parameters since $\mW$ has $E$ rows while each $\mW_m$ has $E/M$ rows.
A similar partitioning of the experts was done in~\citet{yang2021exploring} but not exploited to create different ensemble members, in particular not in conjunction with tiled representations, which we show to be required to get performance gains
(see~\cref{sec:partitioning_without_tiling}).
    
\textbf{Tiled representation.} To jointly handle the predictions of the $M$ ensemble members, we tile the inputs by a factor $M$, inspired by~\citet{wen2020batchensemble}.
This enables a simple implementation of \ack on top of an existing MoE.
In \cref{sec:batch_ensembles_as_sparse_moes}, we connect sparse MoEs and BE, illustrating that tiling naturally fits into the formalism of sparse MoEs.
Because of the tiling, a given image patch has $M$ different representations that, when entering an MoE layer, are each routed to their respective expert subsets $\gE_m$.
Formally, consider some tiled inputs 
$\mH\in\R^{B \times M\times D}$
where $B$ refers to the batch size
(a batch contains image patches)
and $\vh_{i,m} \in \R^D$ is the representation of the $i$-th input for the $m$-th member.
The routing logic in \ack can be written as
\begin{flalign}\label{eq:gating_pbatch_ensemble}
    &\texttt{p-MoE}(\vh_{i, m}) = \sum_{e \in \gE_m} g_{e}(\vh_{i, m}) \cdot \texttt{MLP}_e(\vh_{i, m})
    \ \
    \text{with} \ \ \{g_{e}(\vh_{i, m})\}_{e\in \gE_m}\!\!\! = \texttt{gate}_K(\mW_m \vh_{i, m}),
\end{flalign}
where the routing weights are now $\mW_m \in \R^{(E/M) \times D}$; see \Cref{fig:E3_diagram}.

To echo the observations from \cref{sec:sparse_moes_meet_efficient_ensembles}, we can first see that \ack brings together the static and adaptive combination of ensembles and sparse MoEs, which we found to be complementary. However, we have seen that static ensembling comes at the cost of a large increase in FLOPs, thus we opt for an efficient ensembling approach. Second, we ``split'' the MoE layers along the axis of the experts, i.e., from $E$ experts to $M$ times $E/M$ experts. We do so since we observed that the performance of sparse MoEs tends to plateau quickly for more experts.
We note that \ack retains the property of ensembles to output multiple predictions per input.

In a generic implementation, we tile a batch of $B$ inputs $\mX \in \R^{B \times P}$ by a factor $M$ to obtain the tiled inputs $\mX_\text{tiled} = [\mX;\dots; \mX] \in \R^{(M\cdot B) \times P}$  and the model processes $f(\mX_\text{tiled}; \vtheta)$. 
Since tiling in \ack has an effect only from the first MoE layer onwards, we postpone the tiling operation to that stage, 
thus saving otherwise redundant prior computations in non-MoE-layers.
For example, for L/16 and $K=M=2$, we can save about 47\% of the FLOPs.
Further implementation details of \ack, and a discussion of the increased memory consumption due to tiling, are in~\Cref{sec:E3_implementation_details}. Code can be found at \url{https://github.com/google-research/vmoe}.
Finally, we note that although \ack and BE share conceptual design similarities---tiled representation and sharing of parameters---they differ in fundamental structural ways, see \cref{sec:e3_vs_batch_ensembles}.

\subsection{Ablation Studies: Partitioning and Tiling} \label{sec:pandt_ablations}

Our method introduces two changes to V-MoEs: (a) the partitioning of the experts and (b)~the tiling of the representations. In this section, we assess the separate impact of each of those changes and show that it is indeed their combination that explains the performance gains.
We summarize the results of this study in \cref{tab:partitioning_vs_tiliing} which shows ImageNet performance---negative log likelihood (NLL), classification error, expected calibration error (ECE)~\citep{guo2017calibration}, and Kullback–Leibler divergence---for different %
ablations of \ack.
See \cref{sec:additional_overview_diagrams}, for end-to-end overview diagrams in the style of \cref{fig:E3_diagram}, for the %
these ablations. We use $E = 32$ experts. We provide FLOP measurements for these ablations in \cref{sec:flops_tables}.

\begin{table}[t]
\vspace{-2.0em}
\centering
\caption{ImageNet performance (means $\pm$ standard errors over 8 seeds) of \ack-B/32 ($K=M=2$), V-MoE ($K=4$), and two ablations: \emph{only} tiling and \emph{only} partitioning. The noise in $\texttt{gate}_K$ is denoted by $\sigma$. 
}
\vspace{-.5em}
\label{tab:partitioning_vs_tiliing}
\begin{tabular}{c@{\hskip 3ex}cccc>{\color{red}}c}
\toprule
 &                      \textsc{NLL} $\downarrow$ &                   \textsc{Error} $\downarrow$ &                      \textsc{ECE} $\downarrow$ &                       \textsc{KL} $\uparrow$ \\
\midrule
 V-MoE  &  0.636 {\tiny \color{gray} $\pm$ 0.001} &  16.70 {\tiny \color{gray} $\pm$ 0.04} &           0.034 {\tiny \color{gray} $\pm$ 0.001} &               --- \\ 
 \ack &  \textbf{0.612} {\tiny \color{gray} $\pm$ 0.001} &  \textbf{16.49} {\tiny \color{gray} $\pm$ 0.02} &  \textbf{0.013} {\tiny \color{gray} $\pm$ 0.000} &  \textbf{0.198} {\tiny \color{gray} $\pm$ 0.003} \\
\cmidrule{1-5}
 Tiling &  0.637 {\tiny \color{gray} $\pm$ 0.002} &  16.74 {\tiny \color{gray} $\pm$ 0.06} &  0.028 {\tiny \color{gray} $\pm$ 0.001} &  0.000 {\tiny \color{gray} $\pm$ 0.000} \\
 Tiling ($\sigma \times 2$) &  0.638 {\tiny \color{gray} $\pm$ 0.001} &  16.72 {\tiny \color{gray} $\pm$ 0.03} &  0.033 {\tiny \color{gray} $\pm$ 0.001} &  0.001 {\tiny \color{gray} $\pm$ 0.000} \\
 Tiling ($\sigma \times 4$) &  0.638 {\tiny \color{gray} $\pm$ 0.001} &  16.74 {\tiny \color{gray} $\pm$ 0.03} &  0.033 {\tiny \color{gray} $\pm$ 0.001} &  0.002 {\tiny \color{gray} $\pm$ 0.000} \\
 \cmidrule{1-5}
Partitioning &  0.640 {\tiny \color{gray} $\pm$ 0.001} &  16.72 {\tiny \color{gray} $\pm$ 0.05} &  0.034 {\tiny \color{gray} $\pm$ 0.001} &                                 --- \\
\bottomrule
\end{tabular}%
\vspace{-1.em}
\end{table}

\subsubsection{Partitioning without Tiling}\label{sec:partitioning_without_tiling}

We first compare \ack with a variant of V-MoE where we only partition the set of experts (\emph{Partitioning}).
In this variant, each input $\vh_{i} \in \R^D$ (note the dropping of the index $m$ due to the absence of tiling) can select $K$ experts in subset $\gE_m$, resulting in a total of $K \times M$ selected experts per input. Formally, \cref{eq:gating_pbatch_ensemble} becomes
\begin{equation*}
  \texttt{partitioning-only-MoE}(\vh_{i}) = \sum_{m=1}^M \sum_{e \in \gE_m} g_{e}(\vh_{i}) \cdot \texttt{MLP}_e(\vh_{i}) \ \
    \text{with} \ \ \{g_{e}(\vh_{i})\}_{e\in \gE_m}\!\!\! = \texttt{gate}_K(\mW_m \vh_{i}).
\end{equation*}
The \textit{expert prototyping} of~\citet{yang2021exploring} leads to a similar formulation.
As shown in \cref{tab:partitioning_vs_tiliing}, across all metrics, \emph{Partitioning} is not competitive with \ack. We do not report KL since without tiling, \emph{Partitioning} does not output multiple predictions per input.

\subsubsection{Tiling without Partitioning} \label{sec:tiling_without_partitioning}

We now compare \ack with the variant where only the tiling is enabled (\emph{Tiling}). In this case, we have tiled inputs
$\mH\in\R^{B \times M\times D}$ 
applied to the standard formulation of~\cref{eq:gating}.
Compared with~\cref{eq:gating_pbatch_ensemble}, there is no mechanism to enforce the $M$ representations of the $i$-th input across the ensemble members, i.e., $\{\texttt{MoE}(\vh_{i, m})\}_{m=1}^M$, to be different. Indeed, without partitioning, each $\vh_{i, m}$ could select $K$ identical experts.
As a result, we expect \emph{Tiling} to output $M$ similar predictions across ensemble members. This is confirmed in \cref{tab:partitioning_vs_tiliing} where we observe that the KL for~\emph{Tiling} is orders of magnitude smaller than for \ack. 
To mitigate this effect, we also tried to increase the level of noise $\sigma$ in $\texttt{gate}_K$ (by a factor $\{2, 4\}$), to cause the expert assignments to differ across $\{\vh_{i, m}\}_{m=1}^M$. While we do see an increase in KL, \emph{Tiling} still performs worse than \ack across all metrics.

Interestingly, we can interpret \emph{Tiling} as an approximation, via $M$ samples, of the marginalization $\E_{\rvvarepsilon_1,\dots, \rvvarepsilon_\ell} [f(\vx; \vtheta)]$ with respect to the noise $\{\rvvarepsilon_l\}_{l=1}^\ell$ in the $\ell$ MoE layers of $f(\cdot; \vtheta)$ (further assuming the capacity constraints of the experts, as described in~\citet{riquelme2021scaling}, do not bias the $M$ samples).

\subsubsection{Tiling with Increasing Parameter Sharing} \label{sec:importance_of_partitioning}

The results in \Cref{tab:partitioning_vs_tiliing} (as well as \cref{sec:diversity_and_ens_members}) suggest that the strong performance of \ack is related to its high-diversity predictions.  We hypothesise that this diversity is %
a result of the large number of non-shared parameters in each ensemble member, i.e., the partitioning of the experts. 
To test this hypothesis, we allow the subsets $\mathcal{E}_m$ in \ack to have \emph{some degree of overlap} (i.e., ensemble members share some experts), thus interpolating between \ack and \emph{Tiling}.
For example, with total experts $E=32$ and an ensemble size $M=2$, an overlap of 8 shared experts means that each ensemble member has $(32 - 8)/2 = 12$ unique experts, and $12 + 8 = 20$ in total. \cref{tab:partitioning_importance} shows that increasing the number of shared experts directly decreases diversity and thus NLL, Error, and ECE.
We see the same trends for $K = 1$ (rather than $K = 2$; see \cref{tab:partitioning_importance_part2}).

\begin{table}[t]
\vspace{-2.em}
\centering
\caption{ImageNet performance (means $\pm$ standard errors over 8 seeds) of \ack-B/32 ($K=M=2$), $E=32$ total experts, and varying expert overlap between subsets $\mathcal{E}_m$.
}
\vspace*{-0.5em}
\label{tab:partitioning_importance}
\begin{tabular}{c@{\hskip 3ex}cccc}
\toprule
\textsc{Overlap} &                      \textsc{NLL} $\downarrow$ &                   \textsc{Error} $\downarrow$ &                      \textsc{ECE} $\downarrow$ &                       \textsc{KL} $\uparrow$ \\
\midrule
 0 (=\ack) &  \textbf{0.612} {\tiny \color{gray} $\pm$ 0.001} &  \textbf{16.49} {\tiny \color{gray} $\pm$ 0.02} &  \textbf{0.013} {\tiny \color{gray} $\pm$ 0.000} &  \textbf{0.198} {\tiny \color{gray} $\pm$ 0.003} \\
 2 & 0.617 {\tiny \color{gray} $\pm$ 0.003} & 16.55 {\tiny \color{gray} $\pm$ 0.09} & 0.016 {\tiny \color{gray} $\pm$ 0.001} & 0.167 {\tiny \color{gray} $\pm$ 0.005} \\
 4 & 0.622 {\tiny \color{gray} $\pm$ 0.001} & 16.62 {\tiny \color{gray} $\pm$ 0.02} & 0.017 {\tiny \color{gray} $\pm$ 0.001} & 0.148 {\tiny \color{gray} $\pm$ 0.003} \\
 8 & 0.627 {\tiny \color{gray} $\pm$ 0.002} & 16.67 {\tiny \color{gray} $\pm$ 0.07} & 0.021 {\tiny \color{gray} $\pm$ 0.001} & 0.122 {\tiny \color{gray} $\pm$ 0.010} \\
 16 & 0.639 {\tiny \color{gray} $\pm$ 0.004} & 16.82 {\tiny \color{gray} $\pm$ 0.07} & 0.030 {\tiny \color{gray} $\pm$ 0.003} & 0.077 {\tiny \color{gray} $\pm$ 0.011} \\
 32 (=Tiling) &  0.637 {\tiny \color{gray} $\pm$ 0.002} &  16.74 {\tiny \color{gray} $\pm$ 0.06} &  0.028 {\tiny \color{gray} $\pm$ 0.001} &  0.000 {\tiny \color{gray} $\pm$ 0.000} \\

\bottomrule
\end{tabular}%
\vspace{-0.5em}
\end{table}

\subsubsection{Multiple Predictions without Tiling or Partitioning} \label{sec:multipred_ablation}

As highlighted in \cref{tab:sparse_moes_vs_ensembles}, an ensemble of size $M$ outputs $M$ predictions for a given input (thereafter, averaged) while sparse MoEs only produce a single prediction. 
Thus, a natural question is how much the gains of \ack are simply due to its ability to produce multiple predictions, rather than its specific tiling and partitioning mechanisms?
To answer this, we investigate a simple multi-prediction variant of sparse MoEs (\emph{Multi-pred}) wherein the last MoE layer of the form~\cref{eq:gating} is replaced by
\begin{flalign} \label{eq:multihead_gating}
    &\texttt{Multi-pred-MoE}(\vh) = \{ g_e(\vh) \cdot \texttt{MLP}_e(\vh) \}_{g_e(\vh) > 0} \in \R^{K \times Q}
    \ \
    \text{with}\ \ \{g_e(\vh)\}_{e=1}^E = \texttt{gate}_K(\mW \vh),
\end{flalign}
where we have assumed $\texttt{MLP}_e(\vh) \in \R^Q$. 
Instead of \textit{summing} the expert outputs like in~\cref{eq:gating}, we \textit{stack} the $K$ selected expert contributions (as a reminder, $\texttt{gate}_K$ zeroes out the $E-K$ smallest weights). Keeping track of those $K$ contributions makes it possible to generate $K$ different predictions per input as in the classifier of \cref{fig:E3_diagram},
thus aiming at capturing model uncertainty around the true prediction.
\begin{table}[t]
    \centering
        \caption{
        ImageNet performance (means $\pm$ standard errors over 8 seeds) of V-MoE-B/32 and a
        simple multi-prediction variant (\emph{Multi-pred})
        whose last MoE layer is changed as in~\cref{eq:multihead_gating}.
        }
    \label{tab:feature_vs_pred}
    \vspace{-0.5em}
\begin{tabular}{cccccc}
\toprule
     & \textsc{K} &                        \textsc{NLL} $\downarrow$ &                     \textsc{Error} $\downarrow$ &                        \textsc{ECE} $\downarrow$ &                           \textsc{KL} $\uparrow$ \\
\midrule
 \ack ($M=2$)&          1 &  \textbf{0.622} {\tiny \color{gray} $\pm$ 0.001} &           \textbf{16.70} {\tiny \color{gray} $\pm$ 0.03} &           \textbf{0.018} {\tiny \color{gray} $\pm$ 0.000} & \textbf{0.217} {\tiny \color{gray} $\pm$ 0.003} \\ 
 Multi-pred &          2 &  0.636 {\tiny \color{gray} $\pm$ 0.001} &           17.16 {\tiny \color{gray} $\pm$ 0.02} &           0.024 {\tiny \color{gray} $\pm$ 0.000} & 0.032 {\tiny \color{gray} $\pm$ 0.001} \\ 
 V-MoE &          2 &  0.638 {\tiny \color{gray} $\pm$ 0.001} &  \textbf{16.76} {\tiny \color{gray} $\pm$ 0.05} &           0.033 {\tiny \color{gray} $\pm$ 0.001} &               --- \\ \cmidrule{1-6}
  \ack ($M=2$)&          2 &  \textbf{0.612} {\tiny \color{gray} $\pm$ 0.001} &           \textbf{16.49} {\tiny \color{gray} $\pm$ 0.02} &           \textbf{0.013} {\tiny \color{gray} $\pm$ 0.000} & \textbf{0.198} {\tiny \color{gray} $\pm$ 0.003} \\ 
 Multi-pred &          4 &           0.645 {\tiny \color{gray} $\pm$ 0.001} &           17.39 {\tiny \color{gray} $\pm$ 0.04} &  0.021 {\tiny \color{gray} $\pm$ 0.000} &           0.011 {\tiny \color{gray} $\pm$ 0.001} \\ 
 V-MoE &          4 &  0.636 {\tiny \color{gray} $\pm$ 0.001} &  16.70 {\tiny \color{gray} $\pm$ 0.04} &           0.034 {\tiny \color{gray} $\pm$ 0.001} &               --- \\ \cmidrule{1-6}
 \ack ($M=2$)&          4 &  \textbf{0.611} {\tiny \color{gray} $\pm$ 0.001} &           \textbf{16.45} {\tiny \color{gray} $\pm$ 0.03} &           \textbf{0.014} {\tiny \color{gray} $\pm$ 0.000} & \textbf{0.193} {\tiny \color{gray} $\pm$ 0.003} \\ 
 Multi-pred &          8 &           0.650 {\tiny \color{gray} $\pm$ 0.001} &           17.50 {\tiny \color{gray} $\pm$ 0.03} &  0.021 {\tiny \color{gray} $\pm$ 0.000} &           0.005 {\tiny \color{gray} $\pm$ 0.000} \\
 V-MoE &          8 &  0.635 {\tiny \color{gray} $\pm$ 0.002} &  16.72 {\tiny \color{gray} $\pm$ 0.06} &           0.028 {\tiny \color{gray} $\pm$ 0.001} &               --- \\
\bottomrule
\end{tabular}
\vspace{-1.em}
\end{table}

\cref{tab:feature_vs_pred} compares the ImageNet performance of this simple multiple-prediction method with the standard V-MoE and \ack.
In all cases, \emph{Multi-pred}
under performs relative to \ack.
Indeed, despite improvements in ECE, it is only for $K=2$, that \emph{Multi-pred} 
provides small gains in NLL relative to V-MoE, while its classification error is always worse.
In fact, 
\emph{Multi-pred}
for $K=4$ performs \emph{worse} in terms of NLL, classification error, and diversity than for $K=2$.
The KL diversity metric indicates that the
\emph{Multi-pred}
is unable to provide diverse predictions.
This indicates that it is specifically tiling and partitioning in \ack that provide good performance.

\subsection{Comparison with Alternative Efficient Ensembling Strategies}

\begin{table}[tbp]
\vspace{-2.0em}
\centering
\caption{ImageNet performance (means $\pm$ standard errors over 8 seeds) of different efficient ensemble approaches based on a ViT-B/32 architecture.
}
\label{tab:short_comparison_efficient_ensembles_B32}
\vspace{-0.5em}
\begin{tabular}{cccccccc}
\toprule
 & \textsc{K} & \textsc{M} &                      \textsc{NLL} $\downarrow$ &                   \textsc{Error} $\downarrow$ &                      \textsc{ECE} $\downarrow$ &                       \textsc{KL} $\uparrow$  & \textsc{GFLOPs} $\downarrow$ \\
\midrule
ViT &         -- &         -- &  0.688 {\tiny \color{gray} $\pm$ 0.003} &  18.65 {\tiny \color{gray} $\pm$ 0.08} &  0.022 {\tiny \color{gray} $\pm$ 0.000} &                                -- &  \textbf{78.0} \\
BE ViT &         -- &          2 &  0.682 {\tiny \color{gray} $\pm$ 0.003} &  18.47 {\tiny \color{gray} $\pm$ 0.05} &  0.021 {\tiny \color{gray} $\pm$ 0.000} &  0.040 {\tiny \color{gray} $\pm$ 0.001} & 97.1\\
\cmidrule{1-8}
V-MoE &          2 &         -- &  0.638 {\tiny \color{gray} $\pm$ 0.001} &  \textbf{16.76} {\tiny \color{gray} $\pm$ 0.05} &  0.033 {\tiny \color{gray} $\pm$ 0.001} &                                -- & 94.9\\
MC Dropout V-MoE &          1 &          2 &  0.648 {\tiny \color{gray} $\pm$ 0.002} &  17.10 {\tiny \color{gray} $\pm$ 0.05} &  \textbf{0.019} {\tiny \color{gray} $\pm$ 0.001} &  0.046 {\tiny \color{gray} $\pm$ 0.000} & 97.2 \\
\multirow{2}*{MIMO V-MoE} &          2 &          2 &  0.636 {\tiny \color{gray} $\pm$ 0.002} &  16.97 {\tiny \color{gray} $\pm$ 0.04} &  0.028 {\tiny \color{gray} $\pm$ 0.001} &  0.000 {\tiny \color{gray} $\pm$ 0.000} & 96.3  \\
 &          2 &          4 &  0.672 {\tiny \color{gray} $\pm$ 0.001} &  17.72 {\tiny \color{gray} $\pm$ 0.04} &  0.037 {\tiny \color{gray} $\pm$ 0.000} &  0.001 {\tiny \color{gray} $\pm$ 0.000} & 99.0 \\
 \cmidrule{1-8}
\ack &          1 &          2 &  \textbf{0.622} {\tiny \color{gray} $\pm$ 0.001} &  \textbf{16.70} {\tiny \color{gray} $\pm$ 0.03} &  \textbf{0.018} {\tiny \color{gray} $\pm$ 0.000} &  \textbf{0.217} {\tiny \color{gray} $\pm$ 0.003} & 105.9\\
\bottomrule
\end{tabular}
\end{table}

In the previous subsection we saw that a simple approach to multiple predictions in~a~V-MoE~model~is~unable to achieve good diversity in predictions and thus strong predictive performance. Following~\citet{havasi2020training, soflaei2020aggregated}, a possible fix to this problem would be to 
have a multi-prediction
\textit{and multi-input} approach. Furthermore, other efficient ensembling strategies could provide alternative solutions to this problem. 
Unfortunately, as we show in \cref{tab:short_comparison_efficient_ensembles_B32}, while common efficient ensembling strategies such as BE \citep{wen2020batchensemble}, MC Dropout \citep{gal2016dropout}, and MIMO \citep{havasi2020training}, do improve slightly on ViT/V-MoE, they are unable to match the performance of \ack.
In \cref{sec:efficient_ensemble_comparison}, we provide a detailed description of how we carefully apply these methods to ViT/V-MoE to ensure a fair evaluation (sometimes even designing extensions of these methods). We also give results for additional $K$ and $M$ values.

\section{Evaluation}\label{sec:experiments}

We now benchmark \ack against V-MoE. As a 
baseline we also include results for \emph{downstream} ensembles of V-MoE and ViT. These ensembles offer a natural baseline against \ack as they also use a single upstream checkpoint, are easy to implement, and provide consistent improvements upon V-MoE. In \cref{sec:up_and_down_vs_down_only_ens}, we compare with \emph{upstream} ensembles that require multiple upstream checkpoints \citep{mustafa2020deep}. 
All results correspond to the average over 8 (for \{S, B, L\} single models) or 5 (for H single models and all up/downstream ensembles) replications.
In \cref{sec:additional_results,sec:tabular_results} we provide results for additional datasets and metrics as well as standard errors.
Following \citet{riquelme2021scaling}, we compare the predictive-performance vs.~compute cost trade-offs for each method across a range of ViT families. 
In the results below,
\ack uses $(K, M) = (1, 2)$, single V-MoE models use $K=2$, V-MoE ensembles use $K=1$, and all use $E = 32$.
Experimental details, including those for our upstream training, downstream fine-tuning, hyperparameter sweeps and (linear) few-shot evaluation can 
be found in~\Cref{sec:experiment_settings}.
Our main findings are as follows:

\begin{figure*}[tbp]
    \centering
    \includegraphics[width=\textwidth]{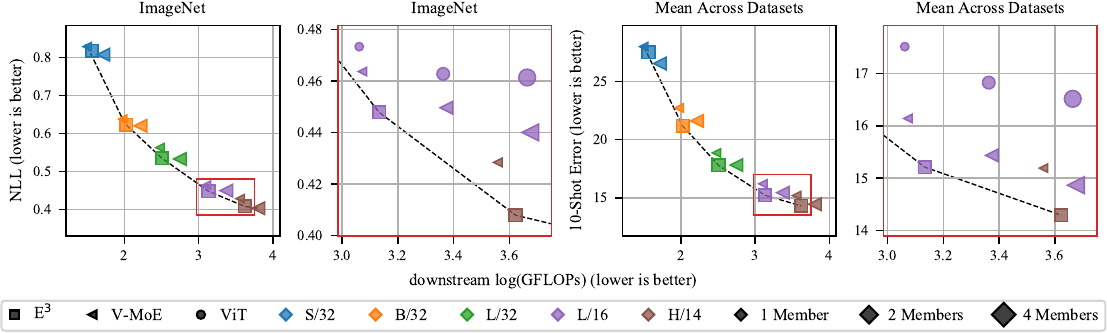}
    \vspace{-1.25em}
    \caption{ImageNet NLL (left, center left) and mean 10-shot error across datasets (center right, right), with zoomed-in plots of highlighted areas. Zoomed-in plots include additional ensemble baselines for comparison. Dashed lines show Pareto frontiers which tend to be defined by \ack.
    }
    \label{fig:nll_and_fewshot_zooms}
    \vspace{-1.25em}
\end{figure*}

\subsection{V-MoE vs. ViT}
\begin{figure*}[tbp]
    \vspace{-2.0em}
    \centering
    \includegraphics[width=\textwidth]{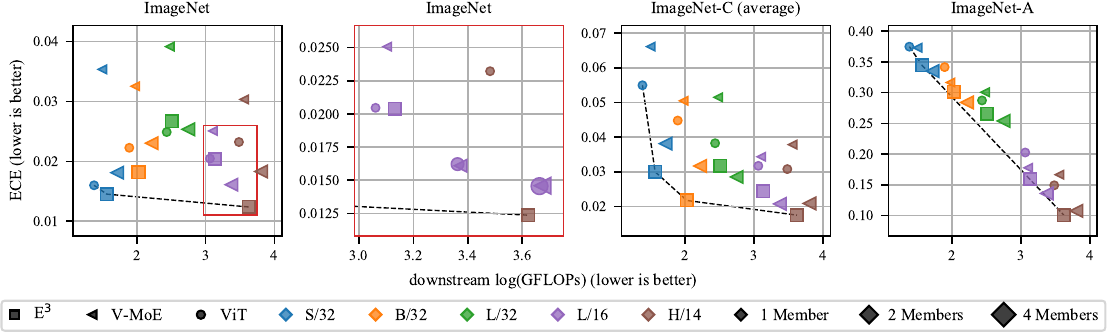}
    \vspace{-1.em}
    \caption{
    ECE for ImageNet (left), with a zoomed-in plot of the highlighted area (center left), and under distribution shift (right, center right). This is a metric for which ensembles are known to perform well whereas, to the best of our knowledge, the performance of V-MoE has not been evaluated. The zoomed-in plot includes additional ensemble baselines for comparison. Dashed lines show Pareto frontiers.
    }
    \label{fig:ece_dist_shift}
    \vspace{-1.em}
\end{figure*}

\begin{figure*}[tbp]
    \centering
    \includegraphics[width=\textwidth]{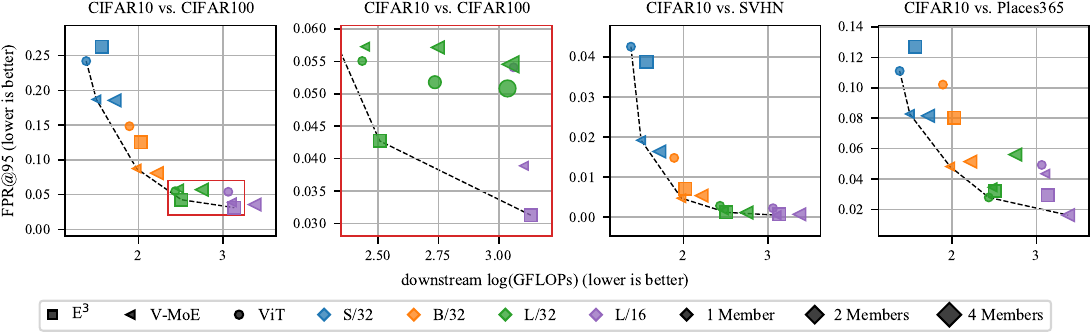}
    \vspace{-0.5cm}
    \caption{
    OOD detection, measured by false positive rate at 95\% precision~\citep{fort2021explore}, for models fine-tuned on CIFAR10, with a zoomed-in plot of the highlighted area. This is a metric for which ensembles are known to perform well whereas, to the best of our knowledge, the performance of V-MoE has not been evaluated. The zoomed-in plot includes additional ensemble baselines for comparison. Dashed lines show Pareto frontiers.
    }
    \label{fig:ood_cifar10}
    \vspace{-0.2cm}
\end{figure*}

$\bullet$
\textbf{Ensembles help V-MoE just as much as ViT.} Ensembling was expected to benefit ViT. However, \cref{fig:vit_vmoe_ens,fig:nll_and_fewshot_zooms,fig:dist_shift,fig:ece_dist_shift,fig:ood_cifar10,fig:cifar_nll} suggest that ensembling provides similar gains for V-MoE in terms of few-shot performance, NLL, ECE, OOD detection, and robustness to distribution shift. We believe this has not been observed before.
Moreover, a downstream ensemble with four H/14 V-MoEs leads to a 88.8\% accuracy on ImageNet (even reaching an impressive 89.3\% for an upstream ensemble
that further benefits from multiple upstream checkpoints, 
see \cref{tab:upstream_vs_downstream_ensembles}).
\\
$\bullet$
\textbf{ViT consistently provides better ECE than V-MoE.} Surprisingly, despite V-MoE tending to have better NLL than ViT (\cref{fig:vit_vmoe_ens}), their ECE is worse (\cref{fig:ece_dist_shift}). 
\\
$\bullet$
\textbf{ECE is not consistent for different ViT/V-MoE families.} We see the ECE, unlike other metrics presented in this work, tends not to provide consistent trends as we increase the ViT family size (\cref{fig:ece_dist_shift}).
This observation is consistent with \citet{minderer2021}, who noted similar behaviour for a range of models. They note that post-hoc temperature scaling can improve consistency.
\\
$\bullet$
\textbf{V-MoE outperforms ViT in OOD detection.} With L/32 being the only exception, V-MoE outperforms ViT on a range of OOD detection tasks (\cref{fig:ood_cifar10}). 
While this may seem surprising, given the opposite trend for ECE, it suggests that ViT makes more accurate predictions about the scale of the uncertainty estimates while V-MoE makes better predictions about the relative ordering of the uncertainty estimates.
\\
$\bullet$
\textbf{For smaller ViT families, V-MoE outperforms ViT in the presence of distribution shift.} In contrast to the OOD detection results, \cref{fig:dist_shift} shows that for smaller ViT families V-MoE improves on the performance of ViT. However, as the ViT family becomes larger, this trend is less consistent.

\subsection{\MNAME}
\begin{figure*}[tbp]
    \vspace{-0.3cm}
    \centering
    \includegraphics[width=\textwidth]{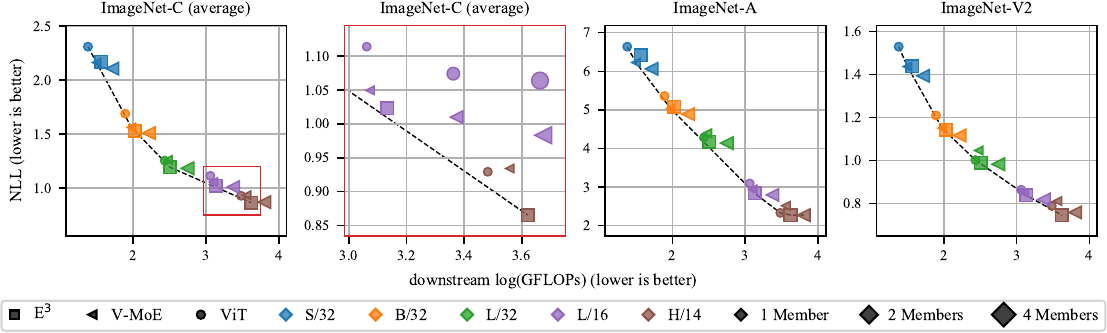}
    \vspace{-0.5cm}
    \caption{
    NLL under distribution shift for models trained on ImageNet. For ImageNet-C, we provide a zoomed-in plot of the highlighted area.
    The zoomed-in plot includes additional ensemble baselines for comparison. Dashed lines show Pareto frontiers which tend to be defined by \ack.
    }
    \label{fig:dist_shift}
\end{figure*}

\begin{figure*}[tbp]
    \centering
    \includegraphics[width=\textwidth]{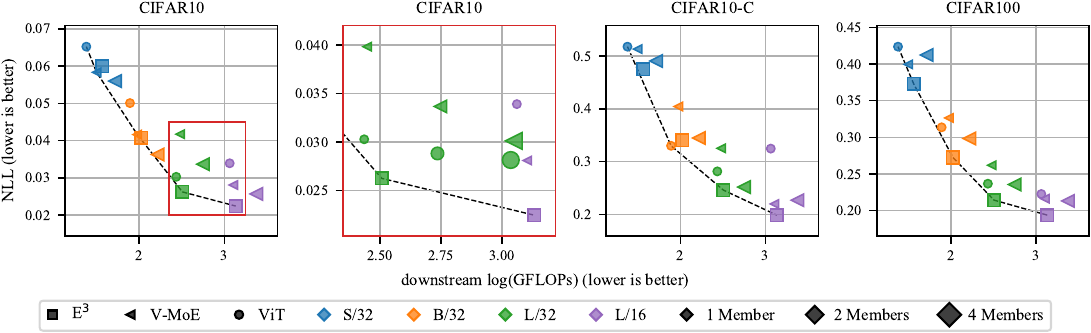}
    \vspace{-0.5cm}
    \caption{
    NLL for CIFAR10 (left), with a zoomed-in plot of the highlighted area (center left), CIFAR10-C (center right), and CIFAR100 (right). The zoomed-in plot includes additional ensemble baselines for comparison. Dashed lines show Pareto frontiers.
    }
    \label{fig:cifar_nll}
    \vspace{-0.25cm}
\end{figure*}

$\bullet$
\textbf{\ack improves classification performance.} As shown in \cref{fig:nll_and_fewshot_zooms}, \ack is either on, or very near to, the Pareto frontiers
for NLL and 10-shot classification error, despite the fact that these are metrics for which ensembles and V-MoE, respectively, are known to perform well. \Cref{fig:cifar_nll} shows that similar conclusions hold for CIFAR10/100 NLL.
\\
$\bullet$
\textbf{\ack performs best at the largest scale.} The difference in predictive performance between \ack and V-MoE---or ensembles thereof---increases as the ViT family becomes larger (\Cref{fig:nll_and_fewshot_zooms,fig:dist_shift,fig:ece_dist_shift,fig:ood_cifar10,fig:cifar_nll}, and \cref{sec:relative_gains} where we propose a scheme to normalize performances across scales).
\\
$\bullet$
\textbf{\ack becomes Pareto efficient for larger ViT families in the presence of distribution shift.} \cref{fig:dist_shift} shows that \ack is more robust to distribution shift for larger ViT families, despite less consistent V-MoE performance at scale. When averaged over the shifted datasets (ImageNet-C, ImageNet-A, ImageNet-V2), \ack improves on V-MoE in terms of NLL for all ViT families other than S/32, with improvements up to 8.33\% at the largest scale; see \cref{sec:nll_shift_summary}.\\
$\bullet$
\textbf{\ack improves ECE over ViT and V-MoE.} Despite V-MoE providing poor ECE, \ack does not suffer from this limitation (\cref{fig:ece_dist_shift}). 
For most ViT families, \ack also provides better ECE than V-MoE ensembles.\\
$\bullet$
\textbf{\ack does not provide consistent OOD detection performance.} Firstly, \cref{fig:ood_cifar10} shows that for small ViT families, \ack performs worse than V-MoE and (even ViT in some cases). Nevertheless, as above, the relative performance improves for larger ViT families such that \ack becomes Pareto efficient for two dataset pairs.
Secondly, \ack seems to perform better on the more difficult near OOD detection task (CIFAR10 vs.~CIFAR100). These results, although sometimes subtle, are consistent across OOD detection metrics and dataset pairs, as shown in \cref{sec:additional_results}.

\paragraph{Summary.} While no single model performs best in all of our evaluation settings, we do find that \ack performs well and is Pareto efficient in \emph{most} cases. This is particularly true for the larger ViT families. One exception was OOD detection, where \ack performance was somewhat inconsistent. On the other hand, while V-MoE clearly outperforms ViT in terms of accuracy, the uncertainty estimates can be better or worse depending on the downstream application.

\section{Related Work}\label{sec:related_work}

\paragraph{Mixture of Experts.}
MoEs~\citep{JacobsJordanNowlanEtAl91, jordan94, journals/nn/ChenXC99, yuksel2012twenty, EigenRS13} combine the outputs of different submodels, or \textit{experts}, 
in an input-dependent way.
Sparse MoEs only select a few experts per input, enabling to greatly scale models while keeping the prediction time constant. Sparse MoEs have been used to build large language models~\citep{shazeer2017outrageously, lepikhin2020gshard, fedus2021switch}.
Recently, sparse MoEs have been also successfully applied to vision problems~\citep{riquelme2021scaling, yang2021exploring, lou2021sparsemlp,xue2021wider}.
Our work builds on the V-MoE architecture proposed by \citet{riquelme2021scaling}, which is based on the vision Transformer (ViT)~\citep{dosovitskiy2020image} and showed improved performance for the same computational cost as ViT.
We explore the interplay between sparse MoEs and ensembles and show that V-MoEs benefit from ensembling, by improving their predictive performance and robustness.
While previous work studied ViT's calibration and robustness~\citep{minderer2021, fort2021explore, paul2021vision, mao2021robust}, we are the first to study the robustness of V-MoE models.

\paragraph{Ensembles.}
Ensemble methods combine several different models to improve generalization and uncertainty estimation.
Ensembles achieve the best performance when they are composed of diverse members that make complementary errors \citep{hansen1990neural, fort2019deep, wenzel2020hyper, DAngelo2021repulsive,lopes2021no}. 
However, standard ensembles are inefficient since they consist of multiple models, each of which can already be expensive.
To reduce test time,
\citet{Xie2013HorizontalAV} and \citet{HinVin15Distilling,tran2020hydra,nam2021diversity} use compression and distillation mechanisms, respectively.
To reduce training time, ensembles can be constructed
with cyclical learning-rate schedules to
snapshot models along the training trajectory~\citep{huang2017snapshot, zhang2019cyclical}.
Our work builds on batch ensemble \citep{wen2020batchensemble} where a \emph{single} model encapsulates an ensemble of networks,
a strategy also explored by \citet{lee2015m, havasi2020training, antoran2020depth, dusenberry2020efficient, rame2021mixmo}.
\citet{wenzel2020hyper} extended BE to models with different hyperparameters.

\paragraph{Bayesian Neural Networks.} BNNs are an alternative approach to deep ensembles for uncertainty quantification in neural networks. In a BNN the weights are treated as random variables and the uncertainty in the weights is translated to uncertainty in predictions via marginalisation. However, because the weight posterior is intractable for NNs, approximation is required. Popular approximations include variational inference \citep{hinton1993keeping,graves2011practical,blundell2015weight}, the Laplace approximation \citep{mackay1992practical,ritter2018scalable}, and MC Dropout \citep{gal2016dropout}. However, many of these approximations make restrictive mean-field assumptions which hurt performance \citep{foong2020expressiveness,coker2022wide,fortuin2021bayesian}. Unfortunately, modeling full weight correlations for even small ResNets---with~relatively few parameters compared to the ViT models considered here---is intractable \citep{daxberger2021baesyain}.

\section{Conclusions and Future Work}

Our study of the interplay between sparse MoEs and ensembles has shown that these two classes of models are symbiotic. \MName exemplifies those mutual benefits---as illustrated by its accuracy, log-likelihood, few-shot learning, robustness, and uncertainty calibration improvements over several challenging baselines in a range of benchmarks.
We have also provided the first, to the best of our knowledge, investigation into the robustness and uncertainty calibration properties of V-MoEs---showing that these models are robust to dataset shift and are able to detect OOD examples.
While our study has focused on downstream fine-tuned models, we believe that an extension to the upstream case and from-scratch training, would also result in a fruitful investigation.
In fact, in \cref{sec:imagenet_from_scratch} we provide some promising, but preliminary, results for from-scratch training.
Similarly, %
although we have focused on computer vision,
our approach should be readily applicable to the modelling of text, where sparse MoEs have been shown to be remarkably effective. With the growing prevalence of sparse MoEs in NLP~\citep{patterson2021carbon}, the questions of understanding and improving the robustness and reliability of such models become increasingly important. Furthermore, the computational scale at which those models operate make those questions challenging to tackle. We believe that our study, and approaches such as \ack, make steps in those directions.

\section*{Acknowledgements}

We would like to extend our gratitude to Josip Djolonga for having reviewed an earlier version of the draft and for helping us with \texttt{robustness\_metrics}~\citep{djolonga2020robustness}. We would also like to thank
André Susano Pinto and Cedric Renggli for fruitful comments on the project, as well as Jakob Uszkoreit for insightful discussions.
JUA acknowledges funding from the EPSRC, the Michael E. Fisher Studentship in Machine Learning, and the Qualcomm Innovation Fellowship.
VF acknowledges funding from the Swiss Data Science Center through a PhD Fellowship, as well as from the Swiss National Science Foundation through a Postdoc.Mobility Fellowship, and from St. John’s College Cambridge through a Junior Research Fellowship.

\bibliography{tmlr}
\bibliographystyle{tmlr}

\clearpage
\appendix
\section*{Appendix Structure}

This appendix is structured as follows:

\begin{itemize}
    \item We provide experimental details, such as model specifications, upstream and downstream training hyperparameters, and few-shot evaluation settings, in \cref{sec:experiment_settings}.
    \item We discuss the compatibility and adaption of upstream V-MoE and ViT checkpoints to the various models considered in this paper, in \cref{sec:upstream_ckpt_compatibility}.
    \item We provide implementation details for \ack, in \cref{sec:E3_implementation_details}.
    \item We show that \ack provides larger relative improvements versus V-MoE for larger models, in \cref{sec:relative_gains}.
    \item We draw connections between BE and sparse MoEs, in \cref{sec:batch_ensembles_as_sparse_moes}.
    \item We highlight differences between \ack and BE, in \cref{sec:e3_vs_batch_ensembles}. 
    \item We compare empirical results for \ack and various other efficient ensemble approaches applied to ViT and V-MoE, in \cref{sec:efficient_ensemble_comparison}.
    \item We provide sensitivity analyses for the load balancing loss strength, expert initialisation diversity, and expert group permutation, in \cref{sec:sensitivity}.
    \item We investigate the roles of diversity and individual model performance in overall ensemble performance, in \cref{sec:diversity_and_ens_members}.
    \item We compare ensembling only downstream with ensembling that takes place both up and downstream, in \cref{sec:up_and_down_vs_down_only_ens}.
    \item We provide preliminary results for training \ack on ImageNet from scratch, without any pre-training, in \cref{sec:imagenet_from_scratch}.
    \item We extend the results of \cref{sec:sparse_moes_meet_efficient_ensembles,sec:method,sec:experiments} to additional datasets, metrics, model sizes, and other settings, in \cref{sec:additional_results}.
    \item We provide a range of tabular results, including flops numbers, parameter counts, and standard errors for the experiments in \cref{sec:experiments}, in \cref{sec:tabular_results}.
    \item We provide additional overview diagrams, in the style of \cref{fig:E3_diagram}, for the various algorithms and ablations presented in the main text, in \cref{sec:additional_overview_diagrams}.
\end{itemize}

\section{Experiment Settings}\label{sec:experiment_settings} 

\subsection{ViT Model Specifications}\label{sec:vit_model_specifications}

Following~\citet{dosovitskiy2020image}, we recall the specifications of the ViT models of different scales in~\Cref{tab:vit_model_specifications}.
\begin{table}[h]
\centering
\caption{Specifications of ViT-S, ViT-B, ViT-L and ViT-H.}
\label{tab:vit_model_specifications}
\begin{tabular}{lccc}
\toprule
 & \textsc{Hidden dimension} & \textsc{MLP dimension} &  \textsc{\# layers} \\
 \midrule
Small & 512 & 2048 & 8 \\
Base & 768 & 3072 & 12 \\
Large & 1024 & 4096 & 24 \\
Huge & 1280 & 5144 & 32 \\
\bottomrule
\end{tabular}
\end{table}

\subsection{Upstream Setting}\label{sec:upstream_setting}

For all our upstream experiments, we scrupulously follow the setting described in~\citet{riquelme2021scaling}, see their Section B.2 in their appendix. For completeness, we just recall that S/32 models are trained for 5 epochs while B/\{16, 32\} and L/32 models are trained for 7 epochs. For L/16 models, both 7 and 14 epochs can be considered~\citep{dosovitskiy2020image, riquelme2021scaling}; we opted for 7 epochs given the breadth of our experiments. Finally, the H/14 model are trained for 14 epochs.

In particular, the models are all trained on JFT-300M~\citep{sun2017revisiting}. This dataset contains about 305M
training and 50\,000 validation images. The labels have a hierarchical structure, with a total of 18\,291 classes, leading on average to 1.89 labels per image.

\subsection{Downstream Setting}\label{sec:downstream_setting}
During fine-tuning, there is a number of \emph{common} design choices we apply. In particular:
\begin{itemize}
  \item Image resolution: 384.
  \item Clipping gradient norm at: 10.0.
  \item Optimizer: SGD with momentum (using half-precision, $\beta = 0.9$).
  \item Batch size: 512.
  \item For V-MoE models, we finetune with capacity ratio $C = 1.5$ and evaluate with $C = 8$.
\end{itemize}

We use the following train/validation splits depending on the dataset:
\begin{center}
\begin{tabular}{ ccc} 
\toprule
\textsc{Dataset} & \textsc{Train Dataset Fraction} & \textsc{Validation Dataset Fraction} \\
\midrule
ImageNet & 99\% & 1\% \\ 
CIFAR10 & 98\% & 2\% \\ 
CIFAR100 & 98\% & 2\% \\ 
Oxford-IIIT Pets & 90\% & 10\% \\ 
Oxford Flowers-102 & 90\% & 10\% \\ 
\bottomrule
\end{tabular}
\end{center}
All those design choices follow from~\citet{riquelme2021scaling} and~\citet{dosovitskiy2020image}.

\subsection{Hyperparameter Sweep for Fine-tuning}

In all our fine-tuning experiments, we use the sweep of hyperparameters described in \cref{tab:finetuning_hp_sweep}. We use the recommendations from~\citet{dosovitskiy2020image} and~\citet{riquelme2021scaling}, further considering several factors {\color{blue}\{0.5, 1.0, 1.5, 2.0\}} to sweep over different numbers of steps. \citet{riquelme2021scaling} use a half schedule (with the factor 0.5) while \citet{dosovitskiy2020image} take the factor 1.0. 

We show in \cref{tab:hp_sweep_impact_on_accuracy} the impact of this enlarged sweep of hyperparameters in the light of the results reported in~\citet{riquelme2021scaling}. We notably tend to improve the performance of ViT and V-MoE (especially for smaller models), which thus makes the baselines we compare to more competitive.

\begin{table}[t]
\centering
\caption{Hyperparameter values for fine-tuning on different datasets. Compared with~\citet{dosovitskiy2020image} and~\citet{riquelme2021scaling}, we further consider several factors {\color{blue}\{0.5, 1.0, 1.5, 2.0\}} to sweep over different numbers of steps.
}
\label{tab:finetuning_hp_sweep}
\begin{tabular}{lccc}
\toprule
\textsc{Dataset}  & \textsc{Steps} & \textsc{Base LR} & \textsc{Expert Dropout} \\
\midrule
ImageNet & 20\,000 $\times$ {\color{blue}\{0.5, 1.0, 1.5, 2.0\}} & \{0.0024, 0.003, 0.01, 0.03\} & 0.1\\
CIFAR10  & \ \ 5\,000 $\times$ {\color{blue}\{0.5, 1.0, 1.5, 2.0\}} & \{0.001, 0.003, 0.01, 0.03\} & 0.1\\
CIFAR100 & \ \ 5\,000 $\times$ {\color{blue}\{0.5, 1.0, 1.5, 2.0\}}  & \{0.001, 0.003, 0.01, 0.03\} & 0.1\\
Oxford-IIIT Pets & \ \ \ \ 500 $\times$ {\color{blue}\{0.5, 1.0, 1.5, 2.0\}} & \{0.001, 0.003, 0.01, 0.03\} & 0.1\\
Oxford Flowers-102 & \ \ \ \ 500 $\times$ {\color{blue}\{0.5, 1.0, 1.5, 2.0\}} & \{0.001, 0.003, 0.01, 0.03\} & 0.1\\
\bottomrule     
\end{tabular}%
\end{table}

\begin{table}[t]
\centering
\caption{Impact of using the enlarged sweep of hyperparameters described in \cref{tab:finetuning_hp_sweep}. We typically improve the results reported in~\citet{riquelme2021scaling}, therefore strengthening the baselines we compare to. The table displays means and standard errors over 8 replications, except for H/14 that has 4 replications. {\bf L/16$^\star$}: For L/16, we consider the setting where the upstream models are trained with 7 epochs, as opposed to 14 epochs in~\citet{riquelme2021scaling}, hence the slightly worse accuracy reported in this paper.}
\label{tab:hp_sweep_impact_on_accuracy}
\begin{tabular}{cccc}
\toprule
\textsc{Model size} &   \textsc{Model name} &            \textsc{Accuracy (this paper)} &  \textsc{Accuracy} \citep{riquelme2021scaling} \\
\midrule
      \multirow{2}{*}{S/32} &          ViT &  76.31 {\tiny \color{gray} $\pm$ 0.05} &                             73.73 \\
       &  V-MoE (K=2) &  78.91 {\tiny \color{gray} $\pm$ 0.08} &                             77.10 \\
      \multirow{2}{*}{B/32} &          ViT &  81.35 {\tiny \color{gray} $\pm$ 0.08} &                             80.73 \\
       &  V-MoE (K=2) &  83.24 {\tiny \color{gray} $\pm$ 0.05} &                             82.60 \\
      \multirow{2}{*}{L/32} &          ViT &  84.62 {\tiny \color{gray} $\pm$ 0.05} &                             84.37 \\
       &  V-MoE (K=2) &  84.95 {\tiny \color{gray} $\pm$ 0.03} &                             85.04 \\
      \multirow{2}{*}{B/16} &          ViT &  84.30 {\tiny \color{gray} $\pm$ 0.06} &                             84.15 \\
       &  V-MoE (K=2) &  85.40 {\tiny \color{gray} $\pm$ 0.04} &                             85.39 \\
      \multirow{2}{*}{L/16$^\star$} &          ViT &  86.63 {\tiny \color{gray} $\pm$ 0.08} &                             87.12 \\
       &  V-MoE (K=2) &  87.12 {\tiny \color{gray} $\pm$ 0.04} &                             87.54 \\
      \multirow{2}{*}{H/14} &  ViT &  88.01 {\tiny \color{gray} $\pm$ 0.05} &                88.08 \\
      &  V-MoE (K=2) &  88.11 {\tiny \color{gray} $\pm$ 0.13} &                             88.23 \\
      
\bottomrule
\end{tabular}
\end{table}

\subsection{Details about the (Linear) Few-shot Evaluation}\label{sec:linear_few_shot_evaluation}

We follow the evaluation methodology proposed by~\citet{dosovitskiy2020image, riquelme2021scaling} which we recall for completeness.
Let us rewrite our model $f$ with parameters $\vtheta=\{\mQ, \vtheta'\}$ as
\begin{equation*}
    f(\vx; \vtheta) = \texttt{softmax}(\mQ \phi(\vx; \vtheta'))
\end{equation*}
where $\mQ \in \R^{C \times S}$ corresponds to the parameters of the last layer of $f$ with the $S$-dimensional representation $\phi(\vx; \vtheta') \in \R^S$.

In linear few-shot evaluation, we construct a linear classifier to predict the target labels (encoded as one-hot vectors) from the $S$-dimensional feature vectors induced by $\phi(\cdot; \vtheta')$; see Chapter 5 in \citet{hastie2017elements} for more background about this type of linear classifiers.
This evaluation protocol makes it possible to evaluate the quality of the representations $\phi$ learned by $f$.

While \citet{dosovitskiy2020image, riquelme2021scaling} essentially focus on the quality of the representations learned upstream on JFT by computing the (linear) few-shot accuracy on ImageNet, we are interested in the representations after fine-tuning on ImageNet. As a result, we consider a collection of 8 few-shot datasets (that does not contain ImageNet):
\begin{itemize}
    \item Caltech-UCSD Birds 200~\citep{wah2011caltech} with 200 classes,
    \item Caltech 101~\citep{bansal2021transfer} with 101 classes,
    \item Cars196~\citep{KrauseStarkDengFei2013} with 196 classes,
    \item CIFAR100~\citep{krizhevsky2009learning} with 100 classes,
    \item Colorectal histology~\citep{kather2016multi} with 8 classes,
    \item Describable Textures Dataset~\citep{cimpoi14describing} with 47 classes,
    \item Oxford-IIIT pet~\citep{parkhi2012cats} with 37 classes and
    \item UC Merced~\citep{yang2010bag} with 21 classes.
\end{itemize}
In the experiments, we compute the few-shot accuracy for each of the above datasets and we report the averaged accuracy over the datasets, for various number of shots in $\{1, 5, 10, 25\}$. As commonly defined in few-shot learning, we understand by $s$~shots a setting wherein we have access to $s$ training images per class label in each of the dataset. 

To account for the different scales of accuracy across the 8 datasets, we also tested to compute a weighted average, normalizing by the accuracy of a reference model (ViT-B/32). This is reminiscent of the normalization carried out in~\citet{hendrycks2019benchmarking} according to the score of AlexNet. We found the conclusions with the standard average and weighted average to be similar.

\subsubsection{Specific Considerations in the Ensemble Case}

For an ensemble with $M$ members,  we have access to $M$ representations $\{\phi(\vx; \vtheta_m')\}_{m=1}^M$ for a given input $\vx$. We have explored two ways to use those representations:
\begin{itemize}
    \item \textbf{Joint}: We concatenate the $M$ representations $\{\phi(\vx; \vtheta_m')\}_{m=1}^M$ into a single ``joint'' feature vector in $\R^{M \cdot S}$, remembering that each
    $\phi(\vx; \vtheta_m') \in \R^S$. We then train a \textit{single} a linear classifier to predict the target labels from the ``joint'' feature vectors.
    \item \textbf{Disjoint}: For each of the $M$ representations $\{\phi(\vx; \vtheta_m')\}_{m=1}^M$, we separately train a linear classifier to predict the target labels from the feature vectors induced by $\phi(\vx; \vtheta_m')$. We then average the predictions of the $M$ linear classifiers trained in this fashion.
\end{itemize}
In \Cref{tab:joint_vs_disjoint_few_shot_ablation}, we report a comparison of those approaches. We aggregate the results over all ensemble models (namely, \ack and upstream ViT/V-MoE ensembles of size 2 and 4) and over 8 replications, for the ViT families S/32, B/32 and L/32.

The results indicate that ``joint'' and ``disjoint'' perform similarly. Throughout our experiments, we use the ``joint'' approach because it eased some implementation considerations.

\subsection{List of Datasets}

For completeness, in addition to the few-shot datasets listed in \cref{sec:linear_few_shot_evaluation}, we list the datasets used for downstream training and evaluation in this work.

\begin{itemize}
    \item ImageNet (ILSVRC2012)~\citep{deng2009imagenet} with 1000 classes and 1281167 training examples.
    \item ImageNet-C~\citep{hendrycks2019benchmarking}, an ImageNet test set constructed by applying 15 different corruptions at 5 levels of intensity to the original ImageNet test set. (We report the mean performance over the different corruptions and intensities.)
    \item ImageNet-A~\citep{hendrycks2019nae}, an ImageNet test set constructed by collecting new data and keeping only those images which a ResNet-50 classified incorrectly.
    \item ImageNet-V2~\citep{recht2019imagenet}, an ImageNet test set independently collected using the same methodology as the original ImageNet dataset.
    \item CIFAR10~\citep{krizhevsky2009learning} with 10 classes and 50000 training examples.
    \item CIFAR10-C~\citep{hendrycks2019benchmarking}, a CIFAR10 test set constructed by applying 15 different corruptions at 5 levels of intensity to the original CIFAR10 test set. (We report the mean performance over the different corruptions and intensities.)
    \item CIFAR100~\citep{krizhevsky2009learning} with 100 classes and training 50000 examples.
    \item Oxford Flowers 102~\citep{nilsback2008automated} with 102 classes and 1020 training examples.
    \item Oxford-IIIT pet~\citep{parkhi2012cats} with 37 classes and 3680 training examples.
    \item SVHN~\citep{netzer2011reading} with 10 classes. 
    \item Places365~\citep{zhou2017places} with 365 classes.
    \item Describable Textures Dataset (DTD)~\citep{cimpoi14describing} with 47 classes.
\end{itemize}

\subsection{Sparse MoEs meet Ensembles Experimental Details} \label{sec:sparse_moes_meet_efficient_ensembles_appendix}

The setup for the experiments in \Cref{fig:static-vs-adaptive,fig:e_k_m_plot} differs slightly the other experiments in this paper. Specifically, while for all other experiments we used upstream V-MoE checkpoints with $(K, E) = (2, 32)$, for these experiments we matched the upstream and downstream checkpoints. We did this to avoid a checkpoint mismatch as a potential confounder in our results.

\subsection{Multiple Predictions without Tiling or Partitioning Details} \label{sec:multi_pred_details}

The naive multi-pred method presented in \cref{sec:multipred_ablation} was trained in almost the same manner as the vanilla V-MoE, the only difference being the handling of multiple predictions. This was accomplished by using the average ensemble member cross entropy as described for \ack in \cref{sec:E3_implementation_details}.
In contrast, in order to compute the evaluation metrics presented in \cref{tab:feature_vs_pred}, we first averaged predictions of the model and then used the average prediction when calculating each metric.

\begin{table}[h]
\caption{Comparison of two approaches, ``joint'' and ``disjoint'', to compute the linear few-shot evaluation in the case of ensembles. For the ViT families S/32, B/32 and L/32, the mean error across datasets is averaged over 8 replications and over all the ensemble models of size 2 and 4.}
\label{tab:joint_vs_disjoint_few_shot_ablation}
\centering
\resizebox{0.8\textwidth}{!}{
\begin{tabular}{ccccccc}
\toprule
\textsc{Model size} &    \textsc{Method} &   \multicolumn{4}{c}{\textsc{Mean error across datasets}} \\
 &    &   \textsc{1 shot} &   \textsc{5 shots} &  \textsc{10 shots} &  \textsc{25 shots} \\
\midrule
      \multirow{2}{*}{S/32} &  disjoint &  51.01 {\tiny \color{gray} $\pm$ 0.43} &  32.80 {\tiny \color{gray} $\pm$ 0.34} &  26.33 {\tiny \color{gray} $\pm$ 0.26} &  20.97 {\tiny \color{gray} $\pm$ 0.18} \\
       &     joint &  51.12 {\tiny \color{gray} $\pm$ 0.42} &  32.81 {\tiny \color{gray} $\pm$ 0.30} &  26.30 {\tiny \color{gray} $\pm$ 0.24} &  20.77 {\tiny \color{gray} $\pm$ 0.17} \\
      \multirow{2}{*}{B/32} &  disjoint &  42.43 {\tiny \color{gray} $\pm$ 0.41} &  25.49 {\tiny \color{gray} $\pm$ 0.21} &  20.30 {\tiny \color{gray} $\pm$ 0.15} &  15.98 {\tiny \color{gray} $\pm$ 0.11} \\
       &     joint &  42.59 {\tiny \color{gray} $\pm$ 0.40} &  25.74 {\tiny \color{gray} $\pm$ 0.18} &  20.54 {\tiny \color{gray} $\pm$ 0.13} &  16.06 {\tiny \color{gray} $\pm$ 0.10} \\
      \multirow{2}{*}{L/32} &  disjoint &  36.41 {\tiny \color{gray} $\pm$ 0.31} &  21.49 {\tiny \color{gray} $\pm$ 0.15} &  17.13 {\tiny \color{gray} $\pm$ 0.12} &  13.56 {\tiny \color{gray} $\pm$ 0.10} \\
       &     joint &  36.48 {\tiny \color{gray} $\pm$ 0.30} &  21.66 {\tiny \color{gray} $\pm$ 0.13} &  17.34 {\tiny \color{gray} $\pm$ 0.10} &  13.56 {\tiny \color{gray} $\pm$ 0.08} \\
\bottomrule
\end{tabular}%
}
\end{table}

\section{Compatibility and Adaptation of the Upstream Checkpoints}\label{sec:upstream_ckpt_compatibility}

Throughout the paper, we make the assumption that we can start from existing checkpoints of ViT and V-MoE models (trained on JFT-300M; see \cref{sec:upstream_setting}).
We next describe how we can use those checkpoints for the fine-tuning of the extensions of ViT and V-MoE that we consider in this paper.

In all our experiments that involve V-MoEs, we consider checkpoints with $K=2$ and $E=32$, which is the canonical setting advocated by~\citet{riquelme2021scaling}.

\subsection{\MNAME}

In the case of \ack, the set of parameters is identical to that of a V-MoE model.
In particular, neither the tiled representation nor the partitioning of the experts transforms the set of parameters. 

To deal with the fact that the single routing function $\texttt{gate}_K(\mW \cdot)$ of a V-MoE becomes separate routing functions $\{\texttt{gate}_K(\mW_m \cdot)\}_{m=1}^M$, one for expert subset $\mathcal{E}_m$, we simply slice row-wise $\mW \in~\R^{E \times D}$ into the $M$ matrices $\mW_m \in \R^{(E/M) \times D}$.

\subsection{Batch Ensembles (BE)}

We train BE starting from ViT checkpoints, which requires to introduce downstream-specific parameters.
Following the design of V-MoEs, we place the batch-ensemble layers in the MLP layers of the Transformer.

Let us consider a dense layer in one of those MLPs, with parameters $\mU \in \R^{D \times L}$, in absence of bias term. In BE, the parametrization of each ensemble member has the following structure $\mU_m = \mU \circ (\vr_m \vs_m^\top)$ where $\{\vr_m\}_{m=1}^M$ and $\{\vs_m\}_{m=1}^M$ are respectively $D$- and $L$-dimensional vectors.

A standard ViT checkpoint provides pre-trained parameters for $\mU$. We then introduce $\{\vr_m\}_{m=1}^M$ and $\{\vs_m\}_{m=1}^M$ at fine-tuning time, following the random initialization schemes proposed in~\citet{wen2020batchensemble}; see details in the hyperparameter sweep for BE in~\Cref{sec:BE_in_efficient_ensemble_comparison}.

\subsection{MIMO}

We train MIMO models from V-MoE checkpoints. The only required modifications are to the input and output parameters of the checkpoints. The linear input embedding must be modified to be compatible with input images containing $M$ times as more channels, as required by the multiple-input structure of MIMO. Similarly, the final dense layer in the classification head must be modified to have $M$ times more output units, following the multiple-output structure in MIMO.

Concretely, the embedding weight $\mW_{\mathrm{in}} \in \mathbb{R}^{H \times W \times 3 \times D}$ is replicated in the third (channel) dimension, resulting in $\mW_{\mathrm{MIMO},\mathrm{in}} \in \mathbb{R}^{H \times W \times 3 \cdot M \times D}$, where $H$ and $W$ are the height and width of the convolution and $D$ is the hidden dimension of the ViT family (specified in \cref{tab:vit_model_specifications}). 
The output layer weight $\mW_{\mathrm{out}} \in \mathbb{R}^{D \times C}$ is replicated in the second (output) dimension, resulting in $\mW_{\mathrm{MIMO},\mathrm{out}} \in \mathbb{R}^{H \times C \cdot M}$, where $C$ is the number of classes. 
The output layer bias $\vb_{\mathrm{out}} \in \mathbb{R}^{C}$ is replicated resulting in $\vb_{\mathrm{MIMO},\mathrm{out}} \in \mathbb{R}^{C \times M}$.
Finally, in order to preserve the magnitude of the activation for these layers, $\mW_{\mathrm{MIMO},\mathrm{in}}$ and $\mW_{\mathrm{MIMO},\mathrm{out}}$ are scaled by $1/M$.

\section{Implementation Details of \MNAME}\label{sec:E3_implementation_details}

We provide details about the training loss and the regularizer used by \ack. We also discuss the memory requirements compared to V-MoE.

\subsection{Training Loss}

Since \ack outputs $M$ predictions $\{f(\vx; \vtheta_m)\}_{m=1}^M$ for a given input $\vx$, we need to adapt the choice of the training loss $\gL$ accordingly.
Following the literature on efficient ensembles~\citep{wen2020batchensemble, dusenberry2020efficient, wenzel2020hyper}, we choose the average ensemble-member cross
entropy
\begin{equation*}
    \gL(y, \vx; \vtheta) = \frac{1}{M} \sum_{m=1}^M \text{cross-entropy}(y, f(\vx; \vtheta_m))
\end{equation*}
instead of other alternatives such as the ensemble cross-entropy
\begin{equation*}
    \text{cross-entropy}\Big(y, \frac{1}{M} \sum_{m=1}^M f(\vx; \vtheta_m) \Big)
\end{equation*}
that was observed to generalize worse~\citep{dusenberry2020efficient}.

\subsection{Auxiliary Losses}
Inspired by previous applications of sparse MoEs in NLP~\citep{shazeer2017outrageously}, \citet{riquelme2021scaling} employ regularizers, also referred to as \textit{auxiliary losses}, to guarantee a balanced usage of the $E$ experts.
Two auxiliary losses---the importance and load losses, see Appendix A in~\citet{riquelme2021scaling} for their formal definitions---are averaged together to form the final regularization term that we denote by $\Omega$.

As a reminder, let us recall the notation of the routing function
\begin{equation*}
    \vh\in \R^D \mapsto \texttt{gate}_K(\mW \vh) = \texttt{top}_K(\texttt{softmax}(\mW \vh + \sigma \rvvarepsilon)) \in \R^E,
\end{equation*}
with $\mW \in \R^{E \times D}$ and $\rvvarepsilon \sim \gN(\vzero, \mI)$. Consider a batch of $B$ inputs $\{\vh_i\}_{i=1}^B$ that we represent by $\mH \in \R^{B \times D}$. Finally, let us define
\begin{equation*}
    \mA = \mH \mW^\top + \sigma \rvvarepsilon_{B \times E} \in \R^{B \times E},
\end{equation*}
where we emphasise that $\rvvarepsilon_{B \times E}$ is a matrix of Gaussian noise entries in $\R^{B \times E}$.
The regularization term $\Omega$ used by~\citet{riquelme2021scaling} can be seen as a function that depends on $\mA$ and $\mH \mW^\top$.

In the context of \mname, the set of $E$ experts is partitioned into $M$ subsets of $E/M$ experts, denoted by $\cup_{m=1}^M \gE_m$; see \Cref{sec:E3_architecture}.
With the introduction of the $M$ routing functions $\{\texttt{gate}_K(\mW_m \cdot)\}_{m=1}^M$ with each $\mW_m \in \R^{(E/M) \times D}$, the matrix $\mA$ becomes accordingly partitioned into $\{\mA_m\}_{m=1}^M$ where each $\mA_m \in \R^{B \times (E/M)}$.

Since we want to enforce a balanced usage of the $E/M$ experts in each subset $\gE_m$ of the experts, we thus redefine the regularization as the average regularization separately applied to each part of the partition
\begin{equation*}
    \Omega_\text{partition}(\mA, \mH \mW^\top) = \frac{1}{M} \sum_{m=1}^M \Omega(\mA_m, \mH \mW_m^\top).
\end{equation*}
We found this option to work better in practice. To guarantee a fair comparison, we also applied $\Omega_\text{partition}$ to the ``Only partitioning'' model in the ablation study of \Cref{sec:partitioning_without_tiling}. 

Following~\citet{riquelme2021scaling}, the regularization parameter controlling the strength of $\Omega_\text{partition}$ was set to 0.01 throughout the experiments.

\subsection{Memory Requirements versus V-MoE} \label{sec:memory_vs_vmoe}

Due to tiling, \ack requires more memory than V-MoE. To be concrete, the memory complexity of V-MoE can be decomposed into two terms
\begin{equation*}
    \mathcal{O}\left(\texttt{memory}_\text{params} + \texttt{memory}_\text{activations}\right),
\end{equation*}
for the forward and backward passes, respectively.
For \ack, the complexity becomes
\begin{equation*}
    \mathcal{O}\left(\texttt{memory}_\text{params} + \texttt{memory}_\text{activations} \times \frac{L_\text{before} + L_\text{after} \times M}{L_\text{total}}\right),
\end{equation*}
where $M$ is the ensemble size, and, $L_\text{before}$, $L_\text{after}$, and $L_\text{total}$ are the number of layers before tiling, after tiling, and in total, respectively. Importantly, neither $\texttt{memory}_\text{params}$ nor $\texttt{memory}_\text{activations}$ depend on $M$.
Thanks to the ``last-$n$'' setting employed in the paper, we have $L_\text{after} \ll L_\text{before}$, and thus the increase in memory due to tiling remains mild. More concretely, for ViT-L, we have $L_\text{total} = 24$, $L_\text{before} = 21$, and $L_\text{after} = 3$ (with MoE layers placed at layers 22 and 24). Thus, for an ensemble of size $M = 2$, \ack would only increase $\texttt{memory}_\text{activations}$ by 12.5\%, while leaving $\texttt{memory}_\text{params}$ unchanged.

\section{\MNAME and V-MoE Relative Improvements per ViT Family} \label{sec:relative_gains}

In \cref{sec:experiments} we claim that \ack performs best at the largest scale. In this section we motivate that claim in more detail. Specifically, we consider two metrics of improvement in performance. Firstly, we consider the percentage improvement in NLL for both \ack and V-MoE versus vanilla ViT. Secondly, we consider a normalised version of this improvement. We consider this second metric to take into account the ``difficulty'' in further improving the NLL of larger ViT family models. Intuitively, the larger the ViT family, the better the corresponding NLL will be, and the more difficult it will be to improve on that NLL.

\begin{figure}[ht]
    \centering
    \includegraphics{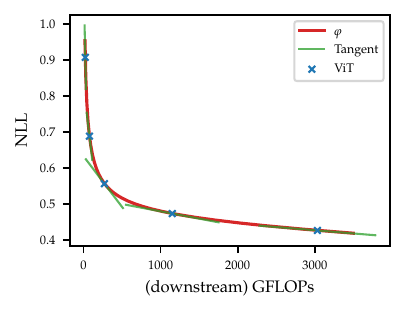}
    \caption{Estimated $\varphi$ compared to the ImageNet NLL values for our ViT models. We also include the tangent at the points corresponding to each ViT model to indicate the gradients at those points.}
    \label{fig:phi_curve}
\end{figure}

The normalisation we apply is based on the gradient of the NLL with respect to FLOPs.
Indeed, this gradient captures the typical variation of NLL at a particular amount of FLOPs.
The ratio of this gradient at the FLOPs values (i.e., the instantaneous change in NLL at those FLOPs values) for two ViT families is a measure of the relative difficulty in increasing the NLL.
Thus, we can use this ratio to normalise our results. 
To be more concrete, let us define the mapping
\begin{equation*}
   \mathrm{NLL} = \varphi(\mathrm{FLOPs})
   \ \
   \text{and its derivative}
   \ \
   \varphi'(\mathrm{FLOPs}) = \frac{d\varphi(\mathrm{FLOPs})}{d\mathrm{FLOPs}}.
\end{equation*}
We estimate $\varphi$ and its gradient by fitting a linear model to the $(\mathrm{NLL}, \mathrm{FLOPs})$ pairs for each ViT family, using the data of the standard ViT models we trained.
We use feature expansion $[1.0, \log(\mathrm{FLOPs}), \log(\mathrm{FLOPs})^2, \log(\mathrm{FLOPs})^3]$ and solve for the parameters of the linear model via ordinary least squares. We determine the gradient of this function at each FLOPs value using automatic differentiation in \texttt{JAX}~\citep{jax2018github}. See \cref{fig:phi_curve} for the resulting fit and an indication of the gradients.

The normalised values are calculated as:
\begin{equation} \label{eq:norm_flops}
    \mathrm{Normalised\ improvement}(v) = \mathrm{improvement}(v)
    \times
    \frac{\varphi'(\textrm{FLOPs}_{\ \text{H/14}})}{\varphi'(\textrm{FLOPs}_{v})},
\end{equation}
where $v$ is one of the ViT families, i.e., S/32, B/32, L/32, L/16, or H/14. 
Note that this normalisation leaves the improvement for H/14 the same. We tried to normalize with respect to other choices of ViT family, different from H/14. Our conclusions are robust, in the sense that both the ordering and the monotonic behavior with respect to scale are preserved.
Using the ratio for normalisation also has the advantage that the normalisation is less sensitive to the particular parameterisation of $\varphi$.

\begin{table}[ht]
    \centering
    \caption{Percentage improvements in NLL for \ack with $(K, M) = (1, 2)$ and V-MoE with $K = 1$ vs.~ViT for families of increasing size. The top two rows show normalised improvements, see \cref{eq:norm_flops}, which take into consideration the increased difficulty of improving NLL for larger ViT families whose performance is beginning to saturate. The bottom two rows are the original percentage improvements without normalisation.}
    \label{tab:relative_gains}
    \begin{tabular}{ccccccc}
    \toprule
     & & \textsc{S/32} & \textsc{B/32} & \textsc{L/32} & \textsc{L/16} & \textsc{H/14} \\ \midrule
    \multirow{2}*{Normalised} & \ack vs.~ViT & 0.02\% & 0.09\% & 0.24\% & 2.35\% & 4.27\% \\
     & V-MoE vs.~ViT & 0.01\% & 0.06\% & -0.04\% & 0.89\% & 0.02\% \\ \cmidrule{1-7}
    \multirow{2}*{Not normalised} & \ack vs.~ViT & 9.82\% & 9.53\% & 3.76\% & 5.38\% & 4.27\% \\
     & V-MoE vs.~ViT & 7.98\% & 6.62\% & -0.60\% & 2.05\% & 0.02\% \\ \bottomrule
    \end{tabular}
\end{table}

\cref{tab:relative_gains} shows both the difficulty-normalised and original improvements (without normalisation). Looking first at the original improvements, we can see that while both \ack and V-MoE have smaller improvements over ViT for larger families, \ack's improvements decrease more slowly. Furthermore, by comparing the normalised improvements we see that \ack's improvements actually \emph{grow} monotonically when taking difficulty into account. This is not the case for V-MoE.

\section{From Batch Ensembles to Sparse MoEs}\label{sec:batch_ensembles_as_sparse_moes}

\citet{wen2020batchensemble} have shown that, given a batch of $B$ inputs $\mX \in \R^{B \times P}$, a single forward pass can efficiently compute the predictions of all the ensemble members $\{f(\mX; \vtheta_m)\}_{m=1}^M$. By appropriately tiling the inputs of the network $\mX_\text{tiled} \in \R^{(M\cdot B) \times P}$ by a factor $M$, each internal operation per ensemble member can then be vectorized.

We take the previous example of a dense layer with parameters $\mU\in \R^{D \times L}$ and we assume the layer receives the tiled inputs $\{\mH_m\}_{m=1}^M$ where $\mH_m \in \R^{B \times D}$. We need to compute for each ensemble member $\mH_m \mU_m = \mH_m [\mU \circ (\vr_m \vs_m^\top)]$. Denoting by $\vh_{i,m} \in \R^D$ the $i$-th input in $\mH_m$, we have
\begin{equation}\label{eq:be_as_sparse_moe}
    \vh_{i,m}^\top \mU_m = \sum_{e=1}^E g_{e}(\vh_{i,m}) \cdot \texttt{expert}_{e}(\vh_{i,m})
    \ \ \
    \text{with}
    \ M=E,\ \ \
    \begin{cases}
    g_{e}(\vh_{i,m}) = 1\ \text{if}\ e=m,\\
    g_{e}(\vh_{i,m}) = 0\ \text{otherwise}
    \end{cases}
\end{equation}
and $\texttt{expert}_{e}(\vz) = \vz^\top [\mU \circ (\vr_e \vs_e^\top)]$. Although \cref{eq:be_as_sparse_moe} may appear as a convoluted way of writing the operations in batch ensembles, it unveils a connection with~\cref{eq:gating}. Indeed, operations in batch ensembles can be seen as a specific sparse MoE, e.g., with binary routing weights depending only on the position in the tiled inputs. While \citet{wen2020batchensemble} primarily tiled the inputs for the sake of efficiency, it also induces some form of conditional computation, an insight that we exploit in \ack.

\section{Batch Ensembles versus \MNAME}\label{sec:e3_vs_batch_ensembles}

\begin{table}[ht]
    \centering
    \caption{Summary of the differences between BE and \ack.}
    \label{tab:e3_vs_batch_ensembles}
    \resizebox{\linewidth}{!}{
    \begin{tabular}{cllll}
    \toprule
         &  \textsc{Shared Parameters} & \textsc{Unique Parameters} & \textsc{Tiling} & \textsc{Training Epochs} \\ \midrule
    BE   & Kernels in each linear layer & ``Fast-weights'' and biases & Start of the model & Typically $\approx$50\% more \\
    \ack & All parameters outside MoE layer & Experts in subset $\mathcal{E}_m$ & Before $1^\text{st}$ MoE layer & Unchanged \\
    \bottomrule
    \end{tabular}
    }
\end{table}

While \ack is inspired in several ways by BE, it has also been specialised to sparse MoEs. This leads to a number of significant differences between the two algorithms. These differences are summarised in \cref{tab:e3_vs_batch_ensembles}, and we elaborate upon them here:
\begin{itemize}
    \item \textbf{Shared parameters:} In BE, the shared parameters are the kernels in each linear (e.g. dense or convolution) layer. This makes up the majority of the parameters in BE. In \ack, the shared parameters are all of those not found in the expert layers.
    \item \textbf{Ensemble member specific parameters:} BE uses pairs of ``fast-weights'' $\{\vr_m, \vs_m\}$ and bias terms for each linear layer. \ack uses the parameters of all the $E/M$ experts in the $m$-th subset $\mathcal{E}_m$ and the router for those expert, in each MoE layer. This results in a much larger set of ensemble member specific parameters, and thus a higher level of predictive diversity (see \cref{sec:importance_of_partitioning}).
    \item \textbf{Tiling:} In BE, tiling is performed at the very beginning of the model, regardless of the model's internal structure. Tiling in \ack occurs before the first MoE layer, thus saving redundant computation for ``Last-n'' V-MoE.
    \item \textbf{Number of training epochs:} BE usually requires 50\% more training epochs than a vanilla model. \ack requires the same number of training epochs as V-MoE.
\end{itemize}

\section{Efficient Ensemble Comparisons}\label{sec:efficient_ensemble_comparison}

\begin{table}[ht]
\centering
\caption{ImageNet performance of different efficient ensemble approaches.
The table reports the means $\pm$ standard errors over 8 replications. All models have a ViT-B/32 architecture. $K$ stands for the sparsity in V-MoEs, $M$ denotes the ensemble size while ``BR'' corresponds to the batch repetition in MIMO~\citep{havasi2020training}.
}
\label{tab:full_comparison_efficient_ensembles_B32}
\begin{tabular}{cccccccc}
\toprule
 & \textsc{K} & \textsc{M} &                      \textsc{NLL} $\downarrow$ &                   \textsc{Error} $\downarrow$ &                      \textsc{ECE} $\downarrow$ &                       \textsc{KL} $\uparrow$ & \textsc{GFLOPs} $\downarrow$ \\
\midrule
ViT &         -- &         -- &  0.688 {\tiny \color{gray} $\pm$ 0.003} &  18.65 {\tiny \color{gray} $\pm$ 0.08} &  0.022 {\tiny \color{gray} $\pm$ 0.000} &                                -- &            78.0 \\
\cmidrule{1-8}
\multirow{2}{*}{BE ViT} &         -- &          2 &  0.682 {\tiny \color{gray} $\pm$ 0.003} &  18.47 {\tiny \color{gray} $\pm$ 0.05} &  0.021 {\tiny \color{gray} $\pm$ 0.000} &  0.040 {\tiny \color{gray} $\pm$ 0.001} &            97.1 \\
 &         -- &          4 &  0.675 {\tiny \color{gray} $\pm$ 0.003} &  18.40 {\tiny \color{gray} $\pm$ 0.09} &  0.017 {\tiny \color{gray} $\pm$ 0.000} &  0.035 {\tiny \color{gray} $\pm$ 0.001} &           135.4 \\
\cmidrule{1-8}
\multirow{4}{*}{V-MoE} &          1 &         -- &  0.642 {\tiny \color{gray} $\pm$ 0.002} &  16.90 {\tiny \color{gray} $\pm$ 0.05} &  0.029 {\tiny \color{gray} $\pm$ 0.001} &                                -- &            82.4 \\
 &          2 &         -- &  0.638 {\tiny \color{gray} $\pm$ 0.001} &  16.76 {\tiny \color{gray} $\pm$ 0.05} &  0.033 {\tiny \color{gray} $\pm$ 0.001} &                                -- &            94.9 \\
 &          4 &         -- &  0.636 {\tiny \color{gray} $\pm$ 0.001} &  16.70 {\tiny \color{gray} $\pm$ 0.04} &  0.034 {\tiny \color{gray} $\pm$ 0.001} &                                -- &           120.1 \\
 &          8 &         -- &  0.635 {\tiny \color{gray} $\pm$ 0.002} &  16.72 {\tiny \color{gray} $\pm$ 0.06} &  0.028 {\tiny \color{gray} $\pm$ 0.001} &                                -- &           170.4 \\
\cmidrule{1-8}
\multirow{5}{*}{MC Dropout V-MoE} &          1 &          2 &  0.648 {\tiny \color{gray} $\pm$ 0.002} &  17.10 {\tiny \color{gray} $\pm$ 0.05} &  0.019 {\tiny \color{gray} $\pm$ 0.001} &  0.046 {\tiny \color{gray} $\pm$ 0.000} &            97.2 \\
 &          1 &          4 &  0.641 {\tiny \color{gray} $\pm$ 0.002} &  16.96 {\tiny \color{gray} $\pm$ 0.05} &  0.017 {\tiny \color{gray} $\pm$ 0.001} &  0.046 {\tiny \color{gray} $\pm$ 0.001} &           135.6 \\
 &          2 &          2 &  0.642 {\tiny \color{gray} $\pm$ 0.002} &  16.94 {\tiny \color{gray} $\pm$ 0.04} &  0.021 {\tiny \color{gray} $\pm$ 0.001} &  0.046 {\tiny \color{gray} $\pm$ 0.001} &           113.7 \\
 &          2 &          4 &  0.634 {\tiny \color{gray} $\pm$ 0.001} &  16.80 {\tiny \color{gray} $\pm$ 0.03} &  0.020 {\tiny \color{gray} $\pm$ 0.000} &  0.046 {\tiny \color{gray} $\pm$ 0.001} &           168.6 \\
 &          4 &          2 &  0.639 {\tiny \color{gray} $\pm$ 0.002} &  16.91 {\tiny \color{gray} $\pm$ 0.06} &  0.022 {\tiny \color{gray} $\pm$ 0.001} &  0.045 {\tiny \color{gray} $\pm$ 0.001} &           146.7 \\
\cmidrule{1-8}
\multirow{2}{*}{MIMO V-MoE (BR=1)} &          2 &          2 &  0.636 {\tiny \color{gray} $\pm$ 0.002} &  16.97 {\tiny \color{gray} $\pm$ 0.04} &  0.028 {\tiny \color{gray} $\pm$ 0.001} &  0.000 {\tiny \color{gray} $\pm$ 0.000} &            96.3 \\
 &          2 &          4 &  0.672 {\tiny \color{gray} $\pm$ 0.001} &  17.72 {\tiny \color{gray} $\pm$ 0.04} &  0.037 {\tiny \color{gray} $\pm$ 0.000} &  0.001 {\tiny \color{gray} $\pm$ 0.000} &            99.0 \\
\multirow{2}{*}{MIMO V-MoE (BR=2)} &          2 &          2 &  0.638 {\tiny \color{gray} $\pm$ 0.001} &  17.14 {\tiny \color{gray} $\pm$ 0.03} &  0.031 {\tiny \color{gray} $\pm$ 0.000} &  0.001 {\tiny \color{gray} $\pm$ 0.000} &           192.6 \\
 &          2 &          4 &  0.665 {\tiny \color{gray} $\pm$ 0.002} &  17.38 {\tiny \color{gray} $\pm$ 0.04} &  0.038 {\tiny \color{gray} $\pm$ 0.000} &  0.000 {\tiny \color{gray} $\pm$ 0.000} &           198.1 \\
\cmidrule{1-8}
\multirow{5}{*}{\ack} &          1 &          2 &  0.622 {\tiny \color{gray} $\pm$ 0.001} &  16.70 {\tiny \color{gray} $\pm$ 0.03} &  0.018 {\tiny \color{gray} $\pm$ 0.000} &  0.217 {\tiny \color{gray} $\pm$ 0.003} &           105.9 \\
 &          1 &          4 &  0.624 {\tiny \color{gray} $\pm$ 0.001} &  16.99 {\tiny \color{gray} $\pm$ 0.03} &  0.013 {\tiny \color{gray} $\pm$ 0.000} &  0.164 {\tiny \color{gray} $\pm$ 0.001} &           153.0 \\
 &          2 &          2 &  0.612 {\tiny \color{gray} $\pm$ 0.001} &  16.49 {\tiny \color{gray} $\pm$ 0.02} &  0.013 {\tiny \color{gray} $\pm$ 0.000} &  0.198 {\tiny \color{gray} $\pm$ 0.003} &           131.1 \\
 &          2 &          4 &  0.620 {\tiny \color{gray} $\pm$ 0.001} &  16.86 {\tiny \color{gray} $\pm$ 0.02} &  0.015 {\tiny \color{gray} $\pm$ 0.000} &  0.170 {\tiny \color{gray} $\pm$ 0.001} &           203.3 \\
 &          4 &          2 &  0.611 {\tiny \color{gray} $\pm$ 0.001} &  16.45 {\tiny \color{gray} $\pm$ 0.03} &  0.014 {\tiny \color{gray} $\pm$ 0.000} &  0.193 {\tiny \color{gray} $\pm$ 0.003} &           181.4 \\
\bottomrule
\end{tabular}
\end{table}

In this section, we compare \mname (\ack) to several popular efficient ensemble approaches, namely MIMO~\citep{havasi2020training}, batch ensemble (BE)~\citep{wen2020batchensemble}, and MC Dropout~\citep{gal2016dropout}.

\Cref{tab:full_comparison_efficient_ensembles_B32} reports the ImageNet performance of those different techniques, when all models are based on a ViT-B/32 architecture.
We start by highlighting the most salient conclusions of the experiment and defer to the next subsections the descriptions of the different competing techniques.

We make the following observations:
\begin{itemize}
    \item BE built upon ViT improves the performance of ViT in terms of NLL, classification error and ECE. However, the resulting increase in FLOPs makes BE a less viable option compared to \ack.
    \item MC Dropout V-MoE is on par with standard V-MoE in terms of NLL and classification error, while it improves the ECE. For all values of $K$, we observe that the performance tends to improve as the number of samples, i.e., $M$, increases. However, already for $M$ in $\{2, 4\}$, the resulting increase in FLOPs makes MC Dropout V-MoE a less favorable option compared to \ack.
    \item Perhaps surprisingly (see detailed investigations in~\Cref{sec:MIMO_in_efficient_ensemble_comparison}), MIMO V-MoE does not lead to improvements compared with V-MoE. In fact, for higher ensembles sizes, MIMO V-MoE results in worse performance than standard V-MoE. Moreover, increasing the batch repetition parameter of MIMO (``BR'' in~\Cref{tab:full_comparison_efficient_ensembles_B32}) further worsens the results. Interestingly, we can see that MIMO does not manage to produce diverse predictions, as illustrated by the small values of KL.
    \item \ack offers the best performance vs.~FLOPs trade-offs, e.g., when looking at $(K, M)=(1, 2)$ and $(K, M)=(2, 2)$. We notably observe that the diversity of the predictions in \ack is orders of magnitude larger than that of the other ensemble approaches.
\end{itemize}

We briefly recall the optimization explained in~\Cref{sec:E3_architecture} to save redundant computations: In the ``last-$n$'' setting of \citet{riquelme2021scaling}, it is sufficient to  tile the representations only when entering the first MoE layer/dropout layer/batch-ensemble layer for respectively \ack/MC Dropout V-MoE/BE. We apply this optimization to all the efficient ensemble methods. 

\subsection{Batch Ensembles}\label{sec:BE_in_efficient_ensemble_comparison}

Following the design of V-MoEs, we place the batch-ensemble layers in the MLP layers of the Transformer, following the ``last-$n$'' setting of \citet{riquelme2021scaling}; see~\Cref{sec:vit_and_vmoe}.

The vectors of the rank-1 parametrization introduced at fine-tuning time (see~\Cref{sec:upstream_ckpt_compatibility}) need to be initialized and optimized.
Following the recommendation from~\citet{wen2020batchensemble}, we consider the following hyperparameters in addition to the common sweep described in~\Cref{tab:finetuning_hp_sweep}:
\begin{itemize}
    \item \textbf{Initialization:} Either a random sign vector with entries in $\{-1, 1\}$ independently drawn with probability $\frac{1}{2}$ or a random Gaussian vector with entries independently drawn from $\gN(1, 0.5)$.
    \item \textbf{Learning-rate scale factor:} The vectors of the rank-1 parametrization are updated with a learning rate scaled by a factor in $\{0.5, 1, 2\}$.
\end{itemize}

\subsection{MC Dropout V-MoEs}

For MC Dropout V-MoE, we take the available fine-tuned V-MoEs and enable dropout at prediction time. Indeed, as described in~\Cref{tab:finetuning_hp_sweep}, all V-MoE models already have a 0.1 dropout rate in the experts.

\subsection{MIMO V-MoEs}\label{sec:MIMO_in_efficient_ensemble_comparison}

Following \citet{havasi2020training} we consider two MIMO-specific hyperparameters, in addition to the hyperparameters listed in \Cref{tab:finetuning_hp_sweep}:
\begin{itemize}
    \item \textbf{Input replication probability:} $\{0.5, 0.625, 0.75\}$
    \item \textbf{Batch repetitions:} $\{1, 2\}$
\end{itemize}
Our preliminary investigations also considered lower input repetition probabilities and higher batch repetitions. However, lower input repetition probabilities tended to result in poorer performance. While higher batch repetitions did help to some extent, the additional computational cost made it impractical. 

Given the surprising result that an ensemble size of $M=2$ provides no performance improvement over the standard V-MoE and that increasing $M$ further provides worse performance, there seems to be some incompatibility between MIMO and V-MoE. In fact, our investigations revealed that ViT is the source of the problems since applying MIMO to vanilla ViT without experts resulted in the same trends as for V-MoE. 
Thus we hypothesise that the differences between ViT and ResNet---the architecture to which MIMO was originally applied by \citet{havasi2020training}---are responsible for MIMO's poor performance when applied to ViT.

\paragraph{Difference 1: Class token.} One of the differences between ViT and ResNet is that ViT makes use of a special learnable class token to classify an image (see \citet{dosovitskiy2020image} for details). ResNet on the other hand makes use of the representation from an entire image for classification. We tried two strategies to mitigate this difference:
\begin{enumerate}
    \item We applied the global average pooling (GAP) and multi-head attention pooling (MAP) classification strategies introduced in \citet{dosovitskiy2020image} and \citet{zhai2021scaling}, respectively. In short, both of these methods make use of all the tokens from an image for classification. However, neither of these strategies made a significant difference to the relative performance of MIMO and ViT. In fact, the choice of classification method was the least impactful hyperparameter in our sweep.
    \item Rather than learning a single class token, we learnt $M$ class tokens. This strategy resulted in MIMO with $M=2$ outperforming ViT. However, for $M > 2$ the improvement was small enough that ViT still outperformed MIMO.
\end{enumerate}

\paragraph{Difference 2: Attention.} The other major difference between ViT and ResNet is the building block for each model. While ResNets are primarily composed of convolution operations, ViT makes heavy used of attention. We hypothesised that attention is less suited to separating the information for $M$ input images stored in the channel dimension of a single image. We tried two strategies to mitigate this potential issue:
\begin{enumerate}
\item We applied the hybrid architecture, described in \citet{dosovitskiy2020image}, in which the input sequence to ViT is formed by CNN feature maps. We used ResNet14 and ResNet50. In both cases, we found that strategy boosted the performance of ViT and MIMO equally.
\item Rather than concatenating images in the channel dimension, we concatenated them in the width dimension, resulting in 3 times as many patches for ViT to process. This strategy was successful in the sense that the MIMO performance for $M>2$ improved significantly. However, the significant additional computational cost made it an infeasible solution.
\end{enumerate}

Our findings suggest that MIMO and ViT are indeed somewhat incompatible. Unfortunately, none of our proposed solutions to this problem provided high enough predictive performance increases (or indeed low enough computational cost increases in some cases) to warrant immediate further investigation.

\section{Sensitivity Analyses} \label{sec:sensitivity}

\subsection{Load Balancing}

\begin{figure}[ht]
    \centering
    \includegraphics[width=\textwidth]{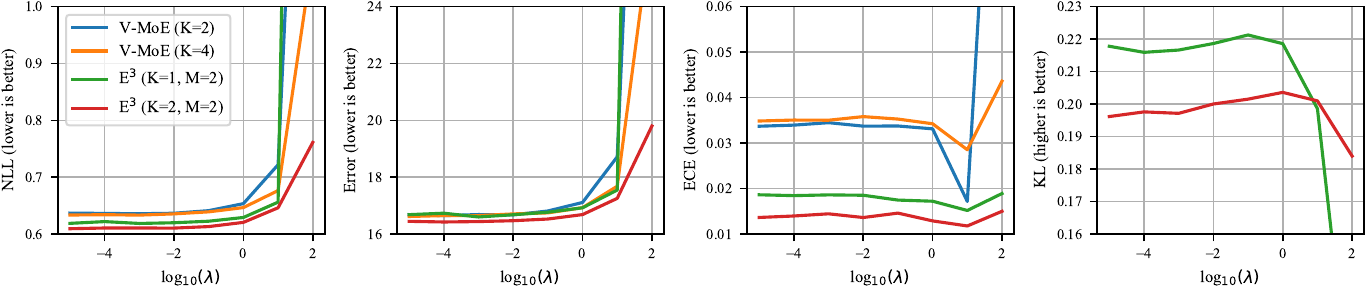}
    \caption{The effect of $\lambda$---the strength of the load balancing, see appendix A of \citet{riquelme2021scaling}---on NLL, Error, ECE and diversity, for \ack and V-MoE. Results are averaged over three random seeds. All models have a ViT-B/32 architecture.}
    \label{fig:example_balancing}
    \vspace{-0.2cm}
\end{figure}

\cref{fig:example_balancing} explores the effect of the load balancing loss on diversity and predictive performance. We see that for both V-MoE and \ack the predictive performance, as measured by NLL, Error, and ECE are largely insensitive to the strength of the load balancing loss. Similarly, the diversity of the predictions of \ack---as measured by the KL---mildly depends on that regularization. Only when the load balancing loss strength becomes excessively large (4 orders of magnitude larger than standard values) do all the performance metrics plummet. In other words, this hyperparameter does not allow us to \emph{increase} the diversity and thereby the predictive performance of our models.

\subsection{Expert Initialization}

\begin{figure}[ht]
    \centering
    \includegraphics[width=\textwidth]{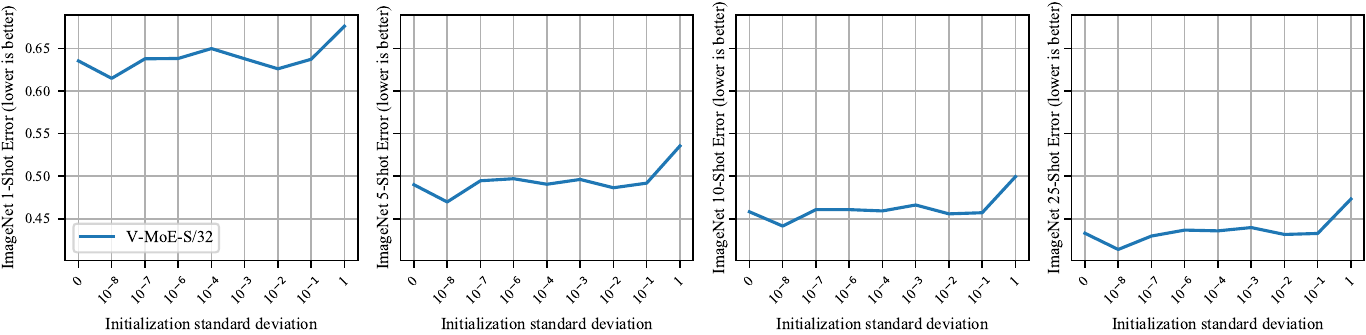}
    \caption{The influence of noise in the expert MLPs' initial random weights. The models are trained on JFT-300M and we measure the ImageNet few-shot Error as in \citet{riquelme2021scaling}. Results are for a single random seed.}
    \label{fig:expert_init}
    \vspace{-0.2cm}
\end{figure}

\cref{fig:expert_init} explores the influence of the initialization of the experts. In particular, we add Gaussian noise with varying standard deviations to the initial weights of the expert MLPs. We notably show that more diverse initializations (larger standard deviations) do not translate to any clear performance gain. Note that this experiment takes place \emph{upstream}, since downstream, the experts are already initialized (we fine-tune them from the upstream checkpoint).

\subsection{Expert Group Permutation}

\cref{tab:permutation_results} investigates the performance of \ack where the ordering of the upstream V-MoE experts has been randomly permuted before fine-tuning\footnote{We also permute the columns of the routing matrix $\mW_m$ accordingly to keep the routing consistent.}. We see that the permutation of the experts has little-to-no significant impact on performance (as measured by NLL and error). The small discrepancies between error for $K=1$ and NLL for $K=2$ can be explained by noting that \ack aggregates results with different upstream checkpoints while \ack (\texttt{permuted}) only takes one of these checkpoints and varies the permutations; this was done to isolate the effect of the permutations only. Note also that the standard errors for \ack and \ack (\texttt{permuted}) are very close to each other, which indicates that permutation of experts has a similar impact to random initialization.

\begin{table}[h]
    \centering
    \caption{Comparison of \ack-B/32 with $E=32$ total experts and $M=2$ partitions, and \ack where the order of the experts has been \texttt{permuted}. For \ack results are aggregated over 8 random seeds whereas for \ack (\texttt{permuted}) results are aggregated over 5 permutations.}
    \label{tab:permutation_results}
    \begin{tabular}{cccc}
    \toprule
     & K &  \textsc{NLL} $\downarrow$ &   \textsc{Error} $\downarrow$ \\
    \midrule
    \multirow{1}{*}{\ack} & 1 &   \textbf{0.622} {\tiny \color{gray} $\pm$ 0.001} &    \textbf{16.70} {\tiny \color{gray} $\pm$ 0.03}\\
    \multirow{1}{*}{\ack (\texttt{permuted})} & 1 &   \textbf{0.620} {\tiny \color{gray} $\pm$ 0.001} &    16.80 {\tiny \color{gray} $\pm$ 0.04}\\ \cmidrule{1-4}
    \multirow{1}{*}{\ack} & 2 &   0.612 {\tiny \color{gray} $\pm$ 0.001} &  \textbf{16.49} {\tiny \color{gray} $\pm$ 0.02} \\
    \multirow{1}{*}{\ack (\texttt{permuted})} & 2 &   \textbf{0.608} {\tiny \color{gray} $\pm$ 0.000} &  \textbf{16.47} {\tiny \color{gray} $\pm$ 0.03} \\
    \bottomrule
    \end{tabular}
\end{table}

\section{The Roles of Ensemble Diversity and Individual Model Performance} \label{sec:diversity_and_ens_members}

\cref{tab:diversity_and_ens_members} shows the individual member performance for each of our efficient ensemble variants as well as upstream and downstream deep ensembles, for sizes $M=2$ and $M=4$. For each method and ensemble size, the first row shows the combined ensemble performance, and the following rows show the performance of the individual ensemble members.

For \ack, the gap between single members and their ensemble is considerable. This is reminiscent of deep ensembles (both variants) and in stark contrast with what we observe for the other efficient ensembles: BE ViT and MIMO V-MoE.
The diversity of \ack is comparable to that of upstream deep ensembles and much larger than downstream deep ensembles.
For example, if we compare MIMO V-MoE (M=2) and \ack (M=2), the single members of \ack are all worse in NLL, ACC, and ECE than the single members of MIMO. However, the performance gap between the individual members and the ensemble is much larger for \ack than MIMO (where there is almost no difference). The diversity of the individual models in \ack is key to its strong performance. Thus, \ack outperforms MIMO. 

In short, this suggests that \ack approximates a deep ensemble of smaller V-MoE models. That is, with \ack, we are able to take advantage of the overparameterization of the experts to create a rich set of ensemble members within a single model. Each ensemble member has a large number of non-shared parameters, thus high induced diversity. In comparison, BE only has a few vectors specific to each member.

\begin{table}[htbp]
    \centering
    \caption{Comparison of the individual ensemble member performance and combined ensemble performance for BE ViT, MIMO V-MoE ($K=2$, BR=1), \ack ($K=1$), as well as upstream and downstream V-MoE ($K=1$) ensembles.}
    \label{tab:diversity_and_ens_members}
    \begin{tabular}{ccccccc}
    \toprule
         & & & \textsc{NLL} $\downarrow$ &                   \textsc{Error} $\downarrow$ &                      \textsc{ECE} $\downarrow$ &                       \textsc{KL} $\uparrow$ \\ \midrule
        \multirow{8}{*}{BE ViT} &   \multirow{3}{*}{M=2} & ensemble & 0.682 {\tiny \color{gray} $\pm$ 0.003} & 18.47 {\tiny \color{gray} $\pm$ 0.05} & 0.021 {\tiny \color{gray} $\pm$ 0.000} & 0.040 {\tiny \color{gray} $\pm$ 0.001} \\ 
        & & member 0 & 0.693 {\tiny \color{gray} $\pm$ 0.003} & 18.68 {\tiny \color{gray} $\pm$ 0.05} & 0.025 {\tiny \color{gray} $\pm$ 0.000} & --- \\ 
        & & member 1 & 0.693 {\tiny \color{gray} $\pm$ 0.003} & 18.68 {\tiny \color{gray} $\pm$ 0.04} & 0.025 {\tiny \color{gray} $\pm$ 0.001} & --- \\ \cmidrule{2-7}
        & \multirow{5}{*}{M=4} & ensemble & 0.675 {\tiny \color{gray} $\pm$ 0.003} & 18.40 {\tiny \color{gray} $\pm$ 0.09} & 0.017 {\tiny \color{gray} $\pm$ 0.000} & 0.035 {\tiny \color{gray} $\pm$ 0.001} \\
        & & member 0 & 0.690 {\tiny \color{gray} $\pm$ 0.003} & 18.70 {\tiny \color{gray} $\pm$ 0.09} & 0.022 {\tiny \color{gray} $\pm$ 0.000} & --- \\
        & & member 1 & 0.690 {\tiny \color{gray} $\pm$ 0.003} & 18.70 {\tiny \color{gray} $\pm$ 0.08} & 0.023 {\tiny \color{gray} $\pm$ 0.000} & --- \\
        & & member 2 & 0.690 {\tiny \color{gray} $\pm$ 0.003} & 18.70 {\tiny \color{gray} $\pm$ 0.07} & 0.023 {\tiny \color{gray} $\pm$ 0.000} & --- \\
        & & member 3 & 0.691 {\tiny \color{gray} $\pm$ 0.003} & 18.70 {\tiny \color{gray} $\pm$ 0.07} & 0.023 {\tiny \color{gray} $\pm$ 0.000} & --- \\ \midrule
        \multirow{8}{*}{MIMO V-MoE} & \multirow{3}{*}{M=2} & ensemble & 0.636 {\tiny \color{gray} $\pm$ 0.002} & 16.97 {\tiny \color{gray} $\pm$ 0.04} & 0.028 {\tiny \color{gray} $\pm$ 0.001} & 0.001 {\tiny \color{gray} $\pm$ 0.000} \\ 
        & & member 0 & 0.636 {\tiny \color{gray} $\pm$ 0.002} & 16.97 {\tiny \color{gray} $\pm$ 0.04} & 0.028 {\tiny \color{gray} $\pm$ 0.001} & --- \\ 
        & & member 1 & 0.636 {\tiny \color{gray} $\pm$ 0.002} & 16.97 {\tiny \color{gray} $\pm$ 0.04} & 0.028 {\tiny \color{gray} $\pm$ 0.000} & --- \\ \cmidrule{2-7}
        & \multirow{5}{*}{M=4} & ensemble & 0.672 {\tiny \color{gray} $\pm$ 0.001} & 17.72 {\tiny \color{gray} $\pm$ 0.04} & 0.037 {\tiny \color{gray} $\pm$ 0.000} & 0.001 {\tiny \color{gray} $\pm$ 0.000} \\
        & & member 0 & 0.672 {\tiny \color{gray} $\pm$ 0.001} & 17.74 {\tiny \color{gray} $\pm$ 0.05} & 0.037 {\tiny \color{gray} $\pm$ 0.000} & --- \\
        & & member 1 & 0.672 {\tiny \color{gray} $\pm$ 0.001} & 17.71 {\tiny \color{gray} $\pm$ 0.05} & 0.037 {\tiny \color{gray} $\pm$ 0.000} & --- \\
        & & member 2 & 0.672 {\tiny \color{gray} $\pm$ 0.001} & 17.72 {\tiny \color{gray} $\pm$ 0.05} & 0.037 {\tiny \color{gray} $\pm$ 0.000} & --- \\
        & & member 3 & 0.673 {\tiny \color{gray} $\pm$ 0.001} & 17.73 {\tiny \color{gray} $\pm$ 0.05} & 0.037 {\tiny \color{gray} $\pm$ 0.000} & --- \\ \midrule
        \multirow{8}{*}{\ack} & \multirow{3}{*}{M=2} & ensemble & 0.622 {\tiny \color{gray} $\pm$ 0.001} & 16.70 {\tiny \color{gray} $\pm$ 0.03} & 0.018 {\tiny \color{gray} $\pm$ 0.000} & 0.217 {\tiny \color{gray} $\pm$ 0.003} \\
        & & member 0 & 0.671{\tiny \color{gray} $\pm$ 0.003} & 17.74 {\tiny \color{gray} $\pm$ 0.06} & 0.038 {\tiny \color{gray} $\pm$ 0.001} & --- \\ 
        & & member 1 & 0.683 {\tiny \color{gray} $\pm$ 0.002} & 17.94 {\tiny \color{gray} $\pm$ 0.005} & 0.038 {\tiny \color{gray} $\pm$ 0.001} & --- \\ \cmidrule{2-7}
        & \multirow{5}{*}{M=4} & ensemble & 0.624 {\tiny \color{gray} $\pm$ 0.001} & 16.99 {\tiny \color{gray} $\pm$ 0.03} & 0.013 {\tiny \color{gray} $\pm$ 0.000} & 0.164 {\tiny \color{gray} $\pm$ 0.001} \\ 
        & & member 0 & 0.677 {\tiny \color{gray} $\pm$ 0.002} & 18.04 {\tiny \color{gray} $\pm$ 0.05} & 0.034 {\tiny \color{gray} $\pm$ 0.001} & --- \\
        & & member 1 & 0.685 {\tiny \color{gray} $\pm$ 0.001} & 18.23 {\tiny \color{gray} $\pm$ 0.05} & 0.034 {\tiny \color{gray} $\pm$ 0.001} & --- \\
        & & member 2 & 0.691 {\tiny \color{gray} $\pm$ 0.002} & 18.35 {\tiny \color{gray} $\pm$ 0.07} & 0.035 {\tiny \color{gray} $\pm$ 0.001} & --- \\
        & & member 3 & 0.697 {\tiny \color{gray} $\pm$ 0.002} & 18.47 {\tiny \color{gray} $\pm$ 0.08} & 0.035 {\tiny \color{gray} $\pm$ 0.001} & --- \\ \midrule
        \multirow{8}{*}{Up-DE} & \multirow{3}{*}{M=2} & ensemble & 0.588 {\tiny \color{gray} $\pm$ 0.001} & 15.74 {\tiny \color{gray} $\pm$ 0.05} & 0.017 {\tiny \color{gray} $\pm$ 0.001} & 0.214 {\tiny \color{gray} $\pm$ 0.001} \\
        & & member 0 & 0.640 {\tiny \color{gray} $\pm$ 0.001} & 16.82 {\tiny \color{gray} $\pm$ 0.04} & 0.030 {\tiny \color{gray} $\pm$ 0.000} & --- \\
        & & member 1 & 0.645 {\tiny \color{gray} $\pm$ 0.002} & 16.97 {\tiny \color{gray} $\pm$ 0.06} & 0.029 {\tiny \color{gray} $\pm$ 0.002} & --- \\ \cmidrule{2-7}
        & \multirow{5}{*}{M=4} & ensemble & 0.561 {\tiny \color{gray} $\pm$ 0.001} & 15.10 {\tiny \color{gray} $\pm$ 0.03} & 0.020 {\tiny \color{gray} $\pm$ 0.000} & 0.214 {\tiny \color{gray} $\pm$ 0.001} \\
        & & member 0 & 0.639 {\tiny \color{gray} $\pm$ 0.001} & 16.80 {\tiny \color{gray} $\pm$ 0.03} & 0.030 {\tiny \color{gray} $\pm$ 0.000} & --- \\
        & & member 1 & 0.641 {\tiny \color{gray} $\pm$ 0.001} & 16.84 {\tiny \color{gray} $\pm$ 0.05} & 0.030 {\tiny \color{gray} $\pm$ 0.000} & --- \\
        & & member 2 & 0.642 {\tiny \color{gray} $\pm$ 0.001} & 16.90 {\tiny \color{gray} $\pm$ 0.03} & 0.031 {\tiny \color{gray} $\pm$ 0.000} & --- \\
        & & member 3 & 0.647 {\tiny \color{gray} $\pm$ 0.002} & 17.05 {\tiny \color{gray} $\pm$ 0.06} & 0.028 {\tiny \color{gray} $\pm$ 0.002} & --- \\ \midrule
        \multirow{8}{*}{Down-DE} & \multirow{3}{*}{M=2} & ensemble & 0.620 {\tiny \color{gray} $\pm$ 0.001} & 16.44 {\tiny \color{gray} $\pm$ 0.04} & 0.023 {\tiny \color{gray} $\pm$ 0.000} & 0.073 {\tiny \color{gray} $\pm$ 0.001} \\
        & & member 0 & 0.642 {\tiny \color{gray} $\pm$ 0.001} & 16.83 {\tiny \color{gray} $\pm$ 0.03} & 0.030 {\tiny \color{gray} $\pm$ 0.000} & --- \\
        & & member 1 & 0.643 {\tiny \color{gray} $\pm$ 0.001} & 16.84 {\tiny \color{gray} $\pm$ 0.03} & 0.030 {\tiny \color{gray} $\pm$ 0.000} & --- \\ \cmidrule{2-7}
        & \multirow{5}{*}{M=4} & ensemble& 0.607 {\tiny \color{gray} $\pm$ 0.000} & 16.17 {\tiny \color{gray} $\pm$ 0.02} & 0.021 {\tiny \color{gray} $\pm$ 0.001} & 0.073 {\tiny \color{gray} $\pm$ 0.000}\\
        & & member 0 & 0.641 {\tiny \color{gray} $\pm$ 0.001} & 16.82 {\tiny \color{gray} $\pm$ 0.03} & 0.030 {\tiny \color{gray} $\pm$ 0.000} & --- \\
        & & member 1 & 0.642 {\tiny \color{gray} $\pm$ 0.001} & 16.85 {\tiny \color{gray} $\pm$ 0.03} & 0.030 {\tiny \color{gray} $\pm$ 0.000} & --- \\
        & & member 2 & 0.643 {\tiny \color{gray} $\pm$ 0.001} & 16.89 {\tiny \color{gray} $\pm$ 0.04} & 0.031 {\tiny \color{gray} $\pm$ 0.000} & --- \\
        & & member 3 & 0.643 {\tiny \color{gray} $\pm$ 0.001} & 16.93 {\tiny \color{gray} $\pm$ 0.07}& 0.031 {\tiny \color{gray} $\pm$ 0.000} & --- \\
        \bottomrule
    \end{tabular}
\end{table}

\begin{landscape}
\section{Upstream \& Downstream versus Downstream-only Ensembles} \label{sec:up_and_down_vs_down_only_ens}

In \cref{sec:experiments}, and \cref{sec:additional_results} include \emph{downstream} deep ensembles (down-DE) of V-MoE, and in some cases ViT, as a baseline. This choice was motivated by the fact that like ViT, V-MoE, and \ack, down-DE requires a only a single upstream checkpoint, which all of the methods more comparable. However, it is clear that using different upstream checkpoints and then further fine-tuning each of these with different random seeds to construct an \emph{upstream} deep ensemble (up-DE) would result in more varied ensemble members and as a result, a better performing ensemble. This idea has recently been explored by \citet{mustafa2020deep}. 

Thus, for completeness, we also investigate the effects of upstream ensembling on V-MoE. \cref{tab:upstream_vs_downstream_ensembles} compares the performance of upstream and downstream V-MoE ($K=1$) ensembles of sizes $M=2$ and $M=4$. Across the range of metrics, for both ImageNet and ImageNet-C, for all ViT families, and for both values of $M$, we see that up-DE outperforms down-DE. In fact, up-DE with $M=2$ is very often better than or equal to down-DE with $M=4$.
This is especially true for the diversity metrics, which indicates that diversity is indeed the driver for improved performance in up-DE. 
Not shown in the table is the very large computational cost associated with training upstream ensembles. 

\begin{table}[h]
    \centering
    \caption{Comparison of upstream and downstream ensembles of V-MoE with ($K=1$).}
    \label{tab:upstream_vs_downstream_ensembles}
    \resizebox{\linewidth}{!}{
\begin{tabular}{ccccccccc|cccccc}
\toprule
     &       &         \; & \multicolumn{6}{c}{\textsc{ImageNet}} & \multicolumn{6}{c}{\textsc{ImageNet-C}} \\
     &       & \textsc{M} &                        \textsc{NLL} $\downarrow$ &                     \textsc{Error} $\downarrow$ &                        \textsc{ECE} $\downarrow$ &                           \textsc{KL} $\uparrow$ &                  \textsc{Cos. Sim.} $\downarrow$ &                   \textsc{Norm. Dis.} $\uparrow$ &                        \textsc{NLL} $\downarrow$ &                     \textsc{Error} $\downarrow$ &                        \textsc{ECE} $\downarrow$ &                           \textsc{KL} $\uparrow$ &                  \textsc{Cos. Sim.} $\downarrow$ &                   \textsc{Norm. Dis.} $\uparrow$ \\
\midrule
\multirow{4}{*}{H/14} & down-DE &          2 &           0.403 {\tiny \color{gray} $\pm$ 0.000} &  11.35 {\tiny \color{gray} $\pm$ 0.05} &  0.018 {\tiny \color{gray} $\pm$ 0.001} &           0.079 {\tiny \color{gray} $\pm$ 0.003} &           0.974 {\tiny \color{gray} $\pm$ 0.001} &           0.488 {\tiny \color{gray} $\pm$ 0.006} &  0.871 {\tiny \color{gray} $\pm$ 0.012} &  21.37 {\tiny \color{gray} $\pm$ 0.20} &  \textbf{0.021} {\tiny \color{gray} $\pm$ 0.001} &           0.218 {\tiny \color{gray} $\pm$ 0.002} &           0.925 {\tiny \color{gray} $\pm$ 0.001} &           0.628 {\tiny \color{gray} $\pm$ 0.003} \\
     & up-DE &          2 &  0.391 {\tiny \color{gray} $\pm$ 0.000} &  11.12 {\tiny \color{gray} $\pm$ 0.10} &  0.016 {\tiny \color{gray} $\pm$ 0.001} &  \textbf{0.126} {\tiny \color{gray} $\pm$ 0.008} &  \textbf{0.963} {\tiny \color{gray} $\pm$ 0.002} &  0.625 {\tiny \color{gray} $\pm$ 0.012} &  0.839 {\tiny \color{gray} $\pm$ 0.011} &  20.66 {\tiny \color{gray} $\pm$ 0.22} &  \textbf{0.022} {\tiny \color{gray} $\pm$ 0.000} &  \textbf{0.355} {\tiny \color{gray} $\pm$ 0.007} &  \textbf{0.892} {\tiny \color{gray} $\pm$ 0.002} &  0.809 {\tiny \color{gray} $\pm$ 0.006} \\
     & down-DE &          4 &             0.392 {\tiny \color{gray} $\pm$ 0.000} &            11.20 {\tiny \color{gray} $\pm$ 0.000} &             0.014 {\tiny \color{gray} $\pm$ 0.000} &             0.083 {\tiny \color{gray} $\pm$ 0.000} &             0.973 {\tiny \color{gray} $\pm$ 0.000} &             0.509 {\tiny \color{gray} $\pm$ 0.000} &             0.851 {\tiny \color{gray} $\pm$ 0.000} &            20.97 {\tiny \color{gray} $\pm$ 0.000} &             \textbf{0.021} {\tiny \color{gray} $\pm$ 0.000} &             0.221 {\tiny \color{gray} $\pm$ 0.000} &             0.923 {\tiny \color{gray} $\pm$ 0.000} &             0.650 {\tiny \color{gray} $\pm$ 0.000} \\
     & up-DE &          4 &             \textbf{0.375} {\tiny \color{gray} $\pm$ 0.000} &            \textbf{10.66} {\tiny \color{gray} $\pm$ 0.000} &             \textbf{0.013} {\tiny \color{gray} $\pm$ 0.000} &             \textbf{0.129} {\tiny \color{gray} $\pm$ 0.000} &             \textbf{0.963} {\tiny \color{gray} $\pm$ 0.000} &             \textbf{0.652} {\tiny \color{gray} $\pm$ 0.000} &             \textbf{0.792} {\tiny \color{gray} $\pm$ 0.000} &            \textbf{19.61} {\tiny \color{gray} $\pm$ 0.000} &             0.032 {\tiny \color{gray} $\pm$ 0.000} &             \textbf{0.361} {\tiny \color{gray} $\pm$ 0.000} &             \textbf{0.892} {\tiny \color{gray} $\pm$ 0.000} &             \textbf{0.850} {\tiny \color{gray} $\pm$ 0.000} \\
\cmidrule{1-15}
\multirow{4}{*}{L/16} & down-DE &          2 &           0.450 {\tiny \color{gray} $\pm$ 0.002} &           12.62 {\tiny \color{gray} $\pm$ 0.04} &           0.016 {\tiny \color{gray} $\pm$ 0.000} &           0.061 {\tiny \color{gray} $\pm$ 0.001} &           0.979 {\tiny \color{gray} $\pm$ 0.000} &           0.419 {\tiny \color{gray} $\pm$ 0.002} &           1.010 {\tiny \color{gray} $\pm$ 0.006} &           24.43 {\tiny \color{gray} $\pm$ 0.12} &  \textbf{0.021} {\tiny \color{gray} $\pm$ 0.000} &           0.168 {\tiny \color{gray} $\pm$ 0.002} &           0.936 {\tiny \color{gray} $\pm$ 0.001} &           0.539 {\tiny \color{gray} $\pm$ 0.003} \\
     & up-DE &          2 &           0.434 {\tiny \color{gray} $\pm$ 0.000} &           12.23 {\tiny \color{gray} $\pm$ 0.04} &  \textbf{0.014} {\tiny \color{gray} $\pm$ 0.000} &  \textbf{0.118} {\tiny \color{gray} $\pm$ 0.001} &  \textbf{0.964} {\tiny \color{gray} $\pm$ 0.000} &           0.584 {\tiny \color{gray} $\pm$ 0.001} &           0.961 {\tiny \color{gray} $\pm$ 0.001} &           23.46 {\tiny \color{gray} $\pm$ 0.03} &           0.023 {\tiny \color{gray} $\pm$ 0.000} &  \textbf{0.342} {\tiny \color{gray} $\pm$ 0.001} &  \textbf{0.890} {\tiny \color{gray} $\pm$ 0.000} &           0.766 {\tiny \color{gray} $\pm$ 0.001} \\
     & down-DE &          4 &           0.440 {\tiny \color{gray} $\pm$ 0.002} &           12.39 {\tiny \color{gray} $\pm$ 0.06} &           0.015 {\tiny \color{gray} $\pm$ 0.000} &           0.061 {\tiny \color{gray} $\pm$ 0.001} &           0.979 {\tiny \color{gray} $\pm$ 0.000} &           0.425 {\tiny \color{gray} $\pm$ 0.002} &           0.983 {\tiny \color{gray} $\pm$ 0.006} &           23.95 {\tiny \color{gray} $\pm$ 0.12} &  \textbf{0.020} {\tiny \color{gray} $\pm$ 0.000} &           0.166 {\tiny \color{gray} $\pm$ 0.001} &           0.937 {\tiny \color{gray} $\pm$ 0.000} &           0.547 {\tiny \color{gray} $\pm$ 0.002} \\
     & up-DE &          4 &  \textbf{0.418} {\tiny \color{gray} $\pm$ 0.000} &  \textbf{11.86} {\tiny \color{gray} $\pm$ 0.01} &  \textbf{0.013} {\tiny \color{gray} $\pm$ 0.000} &  \textbf{0.118} {\tiny \color{gray} $\pm$ 0.000} &  \textbf{0.964} {\tiny \color{gray} $\pm$ 0.000} &  \textbf{0.603} {\tiny \color{gray} $\pm$ 0.001} &  \textbf{0.916} {\tiny \color{gray} $\pm$ 0.001} &  \textbf{22.45} {\tiny \color{gray} $\pm$ 0.02} &           0.034 {\tiny \color{gray} $\pm$ 0.000} &  \textbf{0.341} {\tiny \color{gray} $\pm$ 0.000} &  \textbf{0.890} {\tiny \color{gray} $\pm$ 0.000} &  \textbf{0.800} {\tiny \color{gray} $\pm$ 0.001} \\
\cmidrule{1-15}
\multirow{4}{*}{L/32} & down-DE &          2 &           0.533 {\tiny \color{gray} $\pm$ 0.002} &           14.55 {\tiny \color{gray} $\pm$ 0.04} &           0.025 {\tiny \color{gray} $\pm$ 0.001} &           0.092 {\tiny \color{gray} $\pm$ 0.001} &           0.969 {\tiny \color{gray} $\pm$ 0.000} &           0.479 {\tiny \color{gray} $\pm$ 0.004} &           1.184 {\tiny \color{gray} $\pm$ 0.003} &           27.98 {\tiny \color{gray} $\pm$ 0.04} &           0.029 {\tiny \color{gray} $\pm$ 0.000} &           0.199 {\tiny \color{gray} $\pm$ 0.002} &           0.925 {\tiny \color{gray} $\pm$ 0.001} &           0.556 {\tiny \color{gray} $\pm$ 0.002} \\
     & up-DE &          2 &           0.511 {\tiny \color{gray} $\pm$ 0.001} &           14.07 {\tiny \color{gray} $\pm$ 0.02} &           0.019 {\tiny \color{gray} $\pm$ 0.000} &  \textbf{0.191} {\tiny \color{gray} $\pm$ 0.001} &  \textbf{0.945} {\tiny \color{gray} $\pm$ 0.000} &           0.694 {\tiny \color{gray} $\pm$ 0.005} &           1.133 {\tiny \color{gray} $\pm$ 0.002} &           26.97 {\tiny \color{gray} $\pm$ 0.04} &  \textbf{0.022} {\tiny \color{gray} $\pm$ 0.000} &  \textbf{0.449} {\tiny \color{gray} $\pm$ 0.005} &  \textbf{0.861} {\tiny \color{gray} $\pm$ 0.001} &           0.820 {\tiny \color{gray} $\pm$ 0.003} \\
     & down-DE &          4 &           0.518 {\tiny \color{gray} $\pm$ 0.002} &           14.29 {\tiny \color{gray} $\pm$ 0.03} &           0.022 {\tiny \color{gray} $\pm$ 0.000} &           0.092 {\tiny \color{gray} $\pm$ 0.001} &           0.969 {\tiny \color{gray} $\pm$ 0.000} &           0.487 {\tiny \color{gray} $\pm$ 0.003} &           1.154 {\tiny \color{gray} $\pm$ 0.004} &           27.47 {\tiny \color{gray} $\pm$ 0.05} &           0.023 {\tiny \color{gray} $\pm$ 0.000} &           0.199 {\tiny \color{gray} $\pm$ 0.002} &           0.925 {\tiny \color{gray} $\pm$ 0.001} &           0.567 {\tiny \color{gray} $\pm$ 0.002} \\
     & up-DE &          4 &  \textbf{0.486} {\tiny \color{gray} $\pm$ 0.000} &  \textbf{13.52} {\tiny \color{gray} $\pm$ 0.02} &  \textbf{0.016} {\tiny \color{gray} $\pm$ 0.000} &  \textbf{0.190} {\tiny \color{gray} $\pm$ 0.001} &  \textbf{0.946} {\tiny \color{gray} $\pm$ 0.000} &  \textbf{0.722} {\tiny \color{gray} $\pm$ 0.001} &  \textbf{1.073} {\tiny \color{gray} $\pm$ 0.001} &  \textbf{25.74} {\tiny \color{gray} $\pm$ 0.02} &           0.030 {\tiny \color{gray} $\pm$ 0.000} &  \textbf{0.446} {\tiny \color{gray} $\pm$ 0.001} &  \textbf{0.862} {\tiny \color{gray} $\pm$ 0.000} &  \textbf{0.857} {\tiny \color{gray} $\pm$ 0.000} \\
\cmidrule{1-15}
\multirow{4}{*}{B/16} & down-DE &          2 &           0.519 {\tiny \color{gray} $\pm$ 0.002} &           14.09 {\tiny \color{gray} $\pm$ 0.02} &           0.021 {\tiny \color{gray} $\pm$ 0.001} &           0.048 {\tiny \color{gray} $\pm$ 0.000} &           0.982 {\tiny \color{gray} $\pm$ 0.000} &           0.351 {\tiny \color{gray} $\pm$ 0.002} &           1.316 {\tiny \color{gray} $\pm$ 0.008} &           30.02 {\tiny \color{gray} $\pm$ 0.18} &           0.030 {\tiny \color{gray} $\pm$ 0.000} &           0.132 {\tiny \color{gray} $\pm$ 0.001} &           0.943 {\tiny \color{gray} $\pm$ 0.000} &           0.448 {\tiny \color{gray} $\pm$ 0.002} \\
     & up-DE &          2 &           0.489 {\tiny \color{gray} $\pm$ 0.001} &           13.40 {\tiny \color{gray} $\pm$ 0.03} &  \textbf{0.015} {\tiny \color{gray} $\pm$ 0.000} &  \textbf{0.169} {\tiny \color{gray} $\pm$ 0.002} &  \textbf{0.951} {\tiny \color{gray} $\pm$ 0.000} &           0.668 {\tiny \color{gray} $\pm$ 0.004} &           1.231 {\tiny \color{gray} $\pm$ 0.004} &           28.41 {\tiny \color{gray} $\pm$ 0.09} &  \textbf{0.023} {\tiny \color{gray} $\pm$ 0.000} &  \textbf{0.481} {\tiny \color{gray} $\pm$ 0.006} &  \textbf{0.845} {\tiny \color{gray} $\pm$ 0.001} &           0.838 {\tiny \color{gray} $\pm$ 0.003} \\
     & down-DE &          4 &           0.511 {\tiny \color{gray} $\pm$ 0.002} &           13.95 {\tiny \color{gray} $\pm$ 0.01} &           0.019 {\tiny \color{gray} $\pm$ 0.001} &           0.048 {\tiny \color{gray} $\pm$ 0.000} &           0.982 {\tiny \color{gray} $\pm$ 0.000} &           0.354 {\tiny \color{gray} $\pm$ 0.002} &           1.293 {\tiny \color{gray} $\pm$ 0.008} &           29.67 {\tiny \color{gray} $\pm$ 0.18} &           0.026 {\tiny \color{gray} $\pm$ 0.000} &           0.132 {\tiny \color{gray} $\pm$ 0.001} &           0.943 {\tiny \color{gray} $\pm$ 0.000} &           0.453 {\tiny \color{gray} $\pm$ 0.002} \\
     & up-DE &          4 &  \textbf{0.468} {\tiny \color{gray} $\pm$ 0.000} &  \textbf{12.89} {\tiny \color{gray} $\pm$ 0.03} &  \textbf{0.016} {\tiny \color{gray} $\pm$ 0.000} &  \textbf{0.168} {\tiny \color{gray} $\pm$ 0.000} &  \textbf{0.951} {\tiny \color{gray} $\pm$ 0.000} &  \textbf{0.690} {\tiny \color{gray} $\pm$ 0.001} &  \textbf{1.166} {\tiny \color{gray} $\pm$ 0.002} &  \textbf{27.08} {\tiny \color{gray} $\pm$ 0.05} &           0.037 {\tiny \color{gray} $\pm$ 0.000} &  \textbf{0.479} {\tiny \color{gray} $\pm$ 0.001} &  \textbf{0.846} {\tiny \color{gray} $\pm$ 0.000} &  \textbf{0.879} {\tiny \color{gray} $\pm$ 0.001} \\
\cmidrule{1-15}
\multirow{4}{*}{B/32} & down-DE &          2 &           0.620 {\tiny \color{gray} $\pm$ 0.001} &           16.44 {\tiny \color{gray} $\pm$ 0.04} &           0.023 {\tiny \color{gray} $\pm$ 0.000} &           0.073 {\tiny \color{gray} $\pm$ 0.001} &           0.973 {\tiny \color{gray} $\pm$ 0.000} &           0.414 {\tiny \color{gray} $\pm$ 0.002} &           1.510 {\tiny \color{gray} $\pm$ 0.005} &           33.79 {\tiny \color{gray} $\pm$ 0.08} &           0.032 {\tiny \color{gray} $\pm$ 0.000} &           0.175 {\tiny \color{gray} $\pm$ 0.001} &           0.925 {\tiny \color{gray} $\pm$ 0.000} &           0.498 {\tiny \color{gray} $\pm$ 0.002} \\
     & up-DE &          2 &           0.588 {\tiny \color{gray} $\pm$ 0.001} &           15.74 {\tiny \color{gray} $\pm$ 0.05} &  \textbf{0.017} {\tiny \color{gray} $\pm$ 0.001} &  \textbf{0.214} {\tiny \color{gray} $\pm$ 0.001} &  \textbf{0.937} {\tiny \color{gray} $\pm$ 0.000} &           0.709 {\tiny \color{gray} $\pm$ 0.001} &           1.430 {\tiny \color{gray} $\pm$ 0.003} &           32.37 {\tiny \color{gray} $\pm$ 0.05} &  \textbf{0.022} {\tiny \color{gray} $\pm$ 0.000} &  \textbf{0.537} {\tiny \color{gray} $\pm$ 0.002} &  \textbf{0.824} {\tiny \color{gray} $\pm$ 0.001} &           0.844 {\tiny \color{gray} $\pm$ 0.002} \\
     & down-DE &          4 &           0.607 {\tiny \color{gray} $\pm$ 0.000} &           16.17 {\tiny \color{gray} $\pm$ 0.02} &  \textbf{0.021} {\tiny \color{gray} $\pm$ 0.001} &           0.073 {\tiny \color{gray} $\pm$ 0.000} &           0.973 {\tiny \color{gray} $\pm$ 0.000} &           0.418 {\tiny \color{gray} $\pm$ 0.005} &           1.483 {\tiny \color{gray} $\pm$ 0.008} &           33.36 {\tiny \color{gray} $\pm$ 0.13} &           0.027 {\tiny \color{gray} $\pm$ 0.000} &           0.174 {\tiny \color{gray} $\pm$ 0.001} &           0.926 {\tiny \color{gray} $\pm$ 0.001} &           0.504 {\tiny \color{gray} $\pm$ 0.002} \\
     & up-DE &          4 &  \textbf{0.561} {\tiny \color{gray} $\pm$ 0.001} &  \textbf{15.10} {\tiny \color{gray} $\pm$ 0.03} &           0.020 {\tiny \color{gray} $\pm$ 0.000} &  \textbf{0.214} {\tiny \color{gray} $\pm$ 0.001} &  \textbf{0.937} {\tiny \color{gray} $\pm$ 0.000} &  \textbf{0.739} {\tiny \color{gray} $\pm$ 0.001} &  \textbf{1.357} {\tiny \color{gray} $\pm$ 0.002} &  \textbf{30.92} {\tiny \color{gray} $\pm$ 0.03} &           0.036 {\tiny \color{gray} $\pm$ 0.000} &  \textbf{0.537} {\tiny \color{gray} $\pm$ 0.001} &  \textbf{0.824} {\tiny \color{gray} $\pm$ 0.000} &  \textbf{0.884} {\tiny \color{gray} $\pm$ 0.001} \\
\cmidrule{1-15}
\multirow{4}{*}{S/32} & down-DE &          2 &           0.807 {\tiny \color{gray} $\pm$ 0.003} &           20.90 {\tiny \color{gray} $\pm$ 0.10} &  \textbf{0.018} {\tiny \color{gray} $\pm$ 0.001} &           0.102 {\tiny \color{gray} $\pm$ 0.001} &           0.962 {\tiny \color{gray} $\pm$ 0.000} &           0.458 {\tiny \color{gray} $\pm$ 0.003} &           2.106 {\tiny \color{gray} $\pm$ 0.010} &           44.52 {\tiny \color{gray} $\pm$ 0.18} &           0.038 {\tiny \color{gray} $\pm$ 0.001} &           0.223 {\tiny \color{gray} $\pm$ 0.003} &           0.900 {\tiny \color{gray} $\pm$ 0.001} &           0.521 {\tiny \color{gray} $\pm$ 0.002} \\
     & up-DE &          2 &           0.763 {\tiny \color{gray} $\pm$ 0.001} &           19.85 {\tiny \color{gray} $\pm$ 0.04} &  \textbf{0.016} {\tiny \color{gray} $\pm$ 0.000} &  \textbf{0.305} {\tiny \color{gray} $\pm$ 0.002} &  \textbf{0.911} {\tiny \color{gray} $\pm$ 0.000} &           0.773 {\tiny \color{gray} $\pm$ 0.002} &           2.004 {\tiny \color{gray} $\pm$ 0.004} &           42.92 {\tiny \color{gray} $\pm$ 0.08} &  \textbf{0.025} {\tiny \color{gray} $\pm$ 0.000} &  \textbf{0.683} {\tiny \color{gray} $\pm$ 0.003} &  \textbf{0.767} {\tiny \color{gray} $\pm$ 0.001} &           0.856 {\tiny \color{gray} $\pm$ 0.002} \\
     & down-DE &          4 &           0.795 {\tiny \color{gray} $\pm$ 0.003} &           20.66 {\tiny \color{gray} $\pm$ 0.13} &  \textbf{0.015} {\tiny \color{gray} $\pm$ 0.001} &           0.102 {\tiny \color{gray} $\pm$ 0.002} &           0.962 {\tiny \color{gray} $\pm$ 0.001} &           0.462 {\tiny \color{gray} $\pm$ 0.004} &           2.076 {\tiny \color{gray} $\pm$ 0.012} &           44.16 {\tiny \color{gray} $\pm$ 0.21} &           0.031 {\tiny \color{gray} $\pm$ 0.000} &           0.222 {\tiny \color{gray} $\pm$ 0.003} &           0.900 {\tiny \color{gray} $\pm$ 0.001} &           0.526 {\tiny \color{gray} $\pm$ 0.003} \\
     & up-DE &          4 &  \textbf{0.728} {\tiny \color{gray} $\pm$ 0.001} &  \textbf{19.06} {\tiny \color{gray} $\pm$ 0.04} &           0.025 {\tiny \color{gray} $\pm$ 0.000} &  \textbf{0.304} {\tiny \color{gray} $\pm$ 0.001} &  \textbf{0.911} {\tiny \color{gray} $\pm$ 0.000} &  \textbf{0.808} {\tiny \color{gray} $\pm$ 0.002} &  \textbf{1.914} {\tiny \color{gray} $\pm$ 0.003} &  \textbf{41.38} {\tiny \color{gray} $\pm$ 0.05} &           0.034 {\tiny \color{gray} $\pm$ 0.000} &  \textbf{0.682} {\tiny \color{gray} $\pm$ 0.003} &  \textbf{0.767} {\tiny \color{gray} $\pm$ 0.001} &  \textbf{0.891} {\tiny \color{gray} $\pm$ 0.002} \\
\bottomrule
\end{tabular}
}
\end{table}
\end{landscape}

\section{Preliminary ImageNet Results without Pre-training} \label{sec:imagenet_from_scratch}

Training large-scale sparse MoEs on datasets such as ImageNet and in absence of any pre-training is a difficult task.
Indeed, the massive number of parameters causes models to severely overfit in that regime. In practice, we need to combine various regularization techniques to control this overfitting.

To obtain preliminary results of \ack trained from scratch on ImageNet, we adapt the recipe proposed in the code released by~\citet{riquelme2021scaling}. The recipe is itself based on the regularisation protocol of~\citet{steiner2021train} (``AugReg''), which trained performant dense ViT models from scratch on ImageNet, without pre-training. Overall, compared to the default training schema used for this paper, we add:
\begin{itemize}
    \item Mixup~\citep{zhang2018mixup},
    \item Weight decay,
    \item Dropout~\citep{srivastava2014dropout} on expert MLPs.
\end{itemize}
\begin{table}[t]
    \centering
        \caption{
        ImageNet performance (means $\pm$ standard errors over 3 seeds) of V-MoE and \ack for a S/32 backbone, without pre-training.
        }
    \label{tab:imagenet_from_scratch}
\begin{tabular}{cccc}
\toprule
     & \textsc{K}&                        \textsc{NLL} $\downarrow$ &                     \textsc{Error} $\downarrow$  \\
\midrule
 \ack ($M=2$)&          1 &  \textbf{1.420} {\tiny \color{gray} $\pm$ 0.007} &           \textbf{23.78} {\tiny \color{gray} $\pm$ 0.22}  \\ 
 V-MoE &          2 &  1.478 {\tiny \color{gray} $\pm$ 0.003} &  {24.45} {\tiny \color{gray} $\pm$ 0.08}  \\
\bottomrule
\end{tabular}
\end{table}
The baseline configuration for V-MoE was tuned according to the hyperparameter search space defined by~\citet{steiner2021train}, with an extra sweep over expert MLP dropout in $\{0.1, 0.2\}$. Selecting according to validation accuracy, the optimal V-MoE setting was~\texttt{medium2}, i.e., mixup ratio 0.5 and RandAugment~\citep{cubuk2020randaugment} parameters 2 and 15 (2 augmentations applied of magnitude 15), alongside expert dropout 0.2.
For \ack, these regularization-related hyperparameters were kept fixed and were not further tuned. More precisely, we keep \texttt{medium2} and tune only the learning rate in $\{0.001, 0.003\}$; we thus suspect we could improve our current results for \ack by considering a broader sweep of hyperparameters (e.g., re-tuning all the regularization-related hyperparameters).

In \Cref{tab:imagenet_from_scratch}, we report the performance for both V-MoE and \ack with a S/32 backbone. In terms of both NLL and classification error, \ack outperforms V-MoE. The model checkpoint, along with our code, can be found at \url{https://github.com/google-research/vmoe}.

Code can be found at \url{https://github.com/google-research/vmoe}.

\section{Additional Experimental Results} \label{sec:additional_results}

In this section we expand on various experiments presented in \cref{sec:sparse_moes_meet_efficient_ensembles,sec:method,sec:experiments}. In experiments considering multiple ViT families we also include B/16 which was excluded from the main text for clarity.

\subsection{Static versus Adaptive Combination} \label{sec:sva_part2}

Here we continue the investigation into static versus adaptive combination from \cref{sec:sparse_moes_meet_efficient_ensembles}.

\begin{figure}[h]
    \centering
    \includegraphics{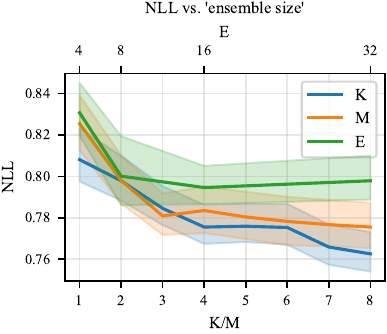}
    \caption{Comparison for the impact on ImageNet NLL of variations in $K$, $E$ and $M$. The underlying model is ViT-S/32.}
    \label{fig:e_k_m_plot}
\end{figure}

\paragraph{Individual gains with respect to $E, K$ and $M$.} \cref{fig:e_k_m_plot} shows the effect of increasing the the various `ensemble size' parameters for a deep ensemble of V-MoEs. In particular, we investigate the static combination axis $M$ (the number of ensemble members), as well as the two adaptive axes---$K$ (the number of experts chosen per patch) and $E$ (the total number of experts). 

When investigating the effect of $K$, we fix $E=32$ and average over $M \in \{1,..,8\}$. Similarly, when investigating $M$, we fix $E=32$ and average over $K \in \{1,..,8\}$. When investigating the effect of $E$ we fix $K = 2$ and average over $M \in \{1,..,8\}$. As a result of this procedure the exact values of the curves are not directly comparable. However, we can still examine the relative trends. 

Specifically, we note that while the variation in $K$ and $M$ curves is roughly of the same size, the variation in the $E$ curve is smaller. We also note that there is very little variation beyond $E=8$ (note the difference in the scales of the axes for the curves). These observations motivate the design of \ack, where we split the sub-models along the $E$ axis, in order to better take advantage of the experts.

\paragraph{Extended Results for the Cumulative Effects of Static and Adaptive Combination.}

\begin{figure}[ht]
    \centering
    \includegraphics[width=0.9\textwidth]{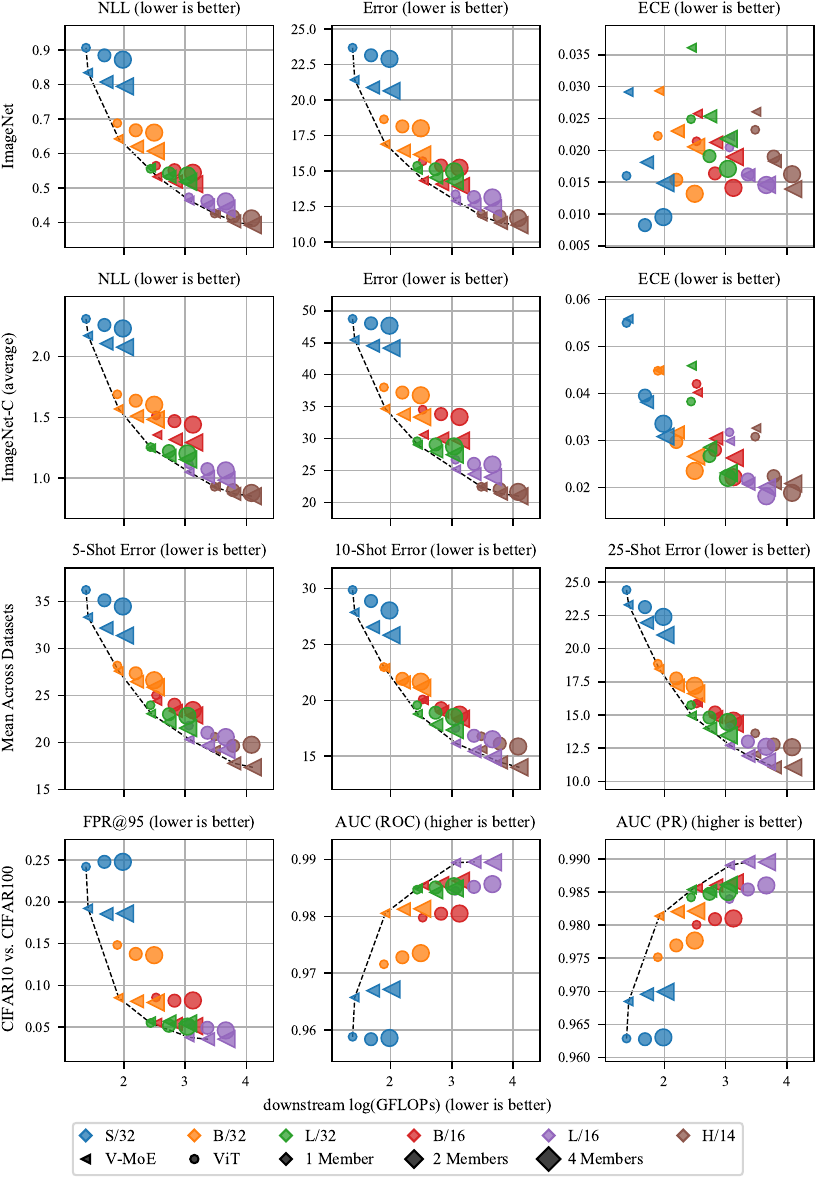}
    \caption{Extended results for \cref{fig:vit_vmoe_ens} to a selection of other tasks and metrics. We see that in most cases, ensembles tend to help ViT and V-MoE ($K = 1$) equally.}
    \label{fig:vit_vmoe_ens_extended}
\end{figure}

In \cref{fig:vit_vmoe_ens_extended} we extend the ImageNet NLL results, presented in \cref{fig:vit_vmoe_ens}, to a range of other datasets and performance metrics. We see that in most cases, the trends are the same as before. That is, (static) ensembling tends to improve the performance of ViT and V-MoE equally. The two exceptions are ECE and OOD detection for ViT-S/32 where we see that larger ensemble sizes can result in decreased performance. 
These results indicate that the for small ViT models, larger ensembles can have slightly lower quality uncertainty estimates.
The trend for ImageNet-C performance is also not as consistent with ensembling sometimes helping ViT or V-MoE less (as indicated by the changes in ordering on the y-axis).

\subsection{An Additional Motivating Experiment -- Deep Ensembles of V-MoEs with Fewer Experts} \label{sec:more_motivation}

As an additional motivation for combining sparse MoEs and ensembles, \cref{tab:ensemble_small_vmoes} compares the performance of a V-MoE with $E=32$ total experts and ensembles of V-MoEs with ($M=2$, $E=16$) and ($M=4$, $E=8$), for both $K=1$ and $K=2$. We see that, in terms of NLL, ($M=4$, $E=8$) is better than ($M=2$, $E=16$) which is in turn better than ($M=1$, $E=32$). We see similar results for Error and ECE.\\~\\
We note that this result is especially remarkable since the \emph{individual} upstream (and later, downstream) models are such that $E=8$ is worse than $E=16$ which in turn performs worse than $E=32$; ensembling thus manages to counterbalance the poorer individual model performance. This suggests that the efficient ensembling---i.e., the combination of multiple models within a single model---of sparse MoEs could lead to strong performance while reducing computational costs.

\begin{table}[h]
    \centering
    \caption{Comparison of upstream deep ensembles of V-MoE-B/32 models with fewer experts. %
    }
    \label{tab:ensemble_small_vmoes}
    \begin{tabular}{ccccllll}
\toprule
& K & M  & E &  \textsc{NLL} $\downarrow$ &    \textsc{Error} $\downarrow$ &       \textsc{ECE} $\downarrow$ & \textsc{KL} $\uparrow$  \\
\midrule
\multirow{6}{*}{V-MoE} & \multirow{3}{*}{1} & 1 & 32 &  0.642 {\tiny \color{gray} $\pm$ 0.002} &  16.90 {\tiny \color{gray} $\pm$ 0.05} &  0.029 {\tiny \color{gray} $\pm$ 0.001} &                                -- \\
&  & 2 & 16 &  0.588 &  15.97 &  0.015 & 0.211 \\
& & 4 & 8  &  0.577 &  15.82  &  0.017 & 0.228 \\ \cmidrule{2-8}
& \multirow{3}{*}{2} & 1 & 32 &  0.638 {\tiny \color{gray} $\pm$ 0.001} &  16.76 {\tiny \color{gray} $\pm$ 0.05} &  0.033 {\tiny \color{gray} $\pm$ 0.001} &                                --   \\
&  & 2 & 16 &  0.583 &  15.75  &  0.016 & 0.211   \\
& & 4 & 8  &  0.580 &  15.94 &  0.015  &  0.184  \\
\bottomrule
\end{tabular}
\end{table}

\subsection{Extended Results for the Tiling with Increasing Parameter Sharing Ablation} \label{sec:importance_of_partitioning_part2}

In \cref{tab:partitioning_importance_part2}, we extend the results for our parameter sharing ablation in \cref{sec:importance_of_partitioning}, from $K=2$ to $K=1$. We see that the results remain the same in this case.

\begin{table}[h]
    \centering
    \caption{Extension of \cref{tab:partitioning_importance}, showing the impact of parameter sharing in \ack, for $K = 1$.}
    \label{tab:partitioning_importance_part2}
    \begin{tabular}{ccccc}
\toprule
 \textsc{Overlap} & \textsc{NLL} $\downarrow$ & \textsc{Error} $\downarrow$ &  \textsc{ECE} $\downarrow$ & \textsc{KL} $\uparrow$ \\
\midrule
  0 (=\ack) & \textbf{0.622} {\tiny \color{gray} $\pm$ 0.001} &  \textbf{16.70} {\tiny \color{gray} $\pm$ 0.03} &  \textbf{0.018} {\tiny \color{gray} $\pm$ 0.000} &  \textbf{0.217} {\tiny \color{gray} $\pm$ 0.003} \\
  2 &  0.627 {\tiny \color{gray} $\pm$ 0.003} &  16.83 {\tiny \color{gray} $\pm$ 0.07} &  0.022 {\tiny \color{gray} $\pm$ 0.001} &  0.194 {\tiny \color{gray} $\pm$ 0.005} \\
  4 &  0.634 {\tiny \color{gray} $\pm$ 0.002} &  16.92 {\tiny \color{gray} $\pm$ 0.07} &  0.024 {\tiny \color{gray} $\pm$ 0.001} &  0.178 {\tiny \color{gray} $\pm$ 0.004} \\
  8 &  0.642 {\tiny \color{gray} $\pm$ 0.001} &  17.04 {\tiny \color{gray} $\pm$ 0.10} &  0.028 {\tiny \color{gray} $\pm$ 0.001} &  0.151 {\tiny \color{gray} $\pm$ 0.009} \\
 16 &  0.659 {\tiny \color{gray} $\pm$ 0.004} &  17.28 {\tiny \color{gray} $\pm$ 0.12} &  0.036 {\tiny \color{gray} $\pm$ 0.001} &  0.103 {\tiny \color{gray} $\pm$ 0.009} \\
\bottomrule
\end{tabular}
\end{table}

\subsection{Extended Results for Few-shot Learning} \label{sec:extended_fewshot}

\begin{figure}[ht]
    \centering
    \includegraphics[width=\textwidth]{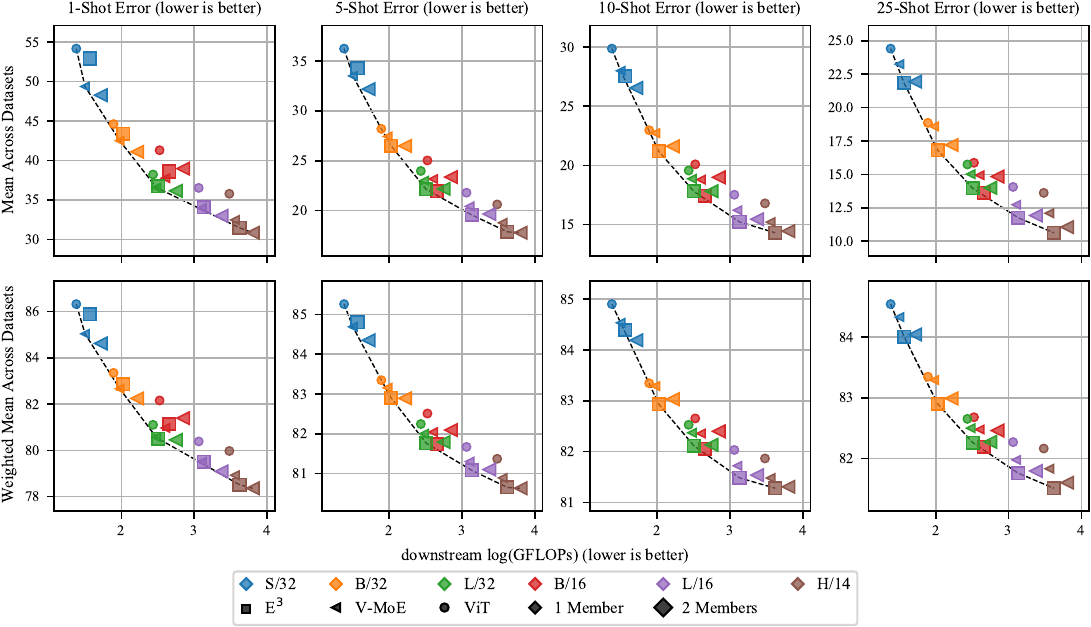}
    \caption{Extended few-shot results from \cref{fig:nll_and_fewshot_zooms} with an additional aggregation method and numbers of shots.}
    \label{fig:extended_fewshot}
\end{figure}

In \cref{fig:extended_fewshot}, we extend the few-shot learning results of \cref{fig:nll_and_fewshot_zooms} to also include 1, 5, and 25-shot. Additionally, we show results for the weighted aggregation strategy mentioned in \cref{sec:linear_few_shot_evaluation}.

We confirm the result that few-shot performance for \ack gets better, relative to the other baselines, with larger ViT families. Additionally, we see that \ack performance seems to get better, again relative to the other baselines, with more shots. This phenomenon can most easily be noticed by comparing the results for S/32 across the different numbers of shots.
Finally, we see that the trends with and without the weighted mean are the same.

\subsection{Extended Results for OOD Detection}

\begin{figure}[ht]
    \centering
    \includegraphics[width=0.9\textwidth]{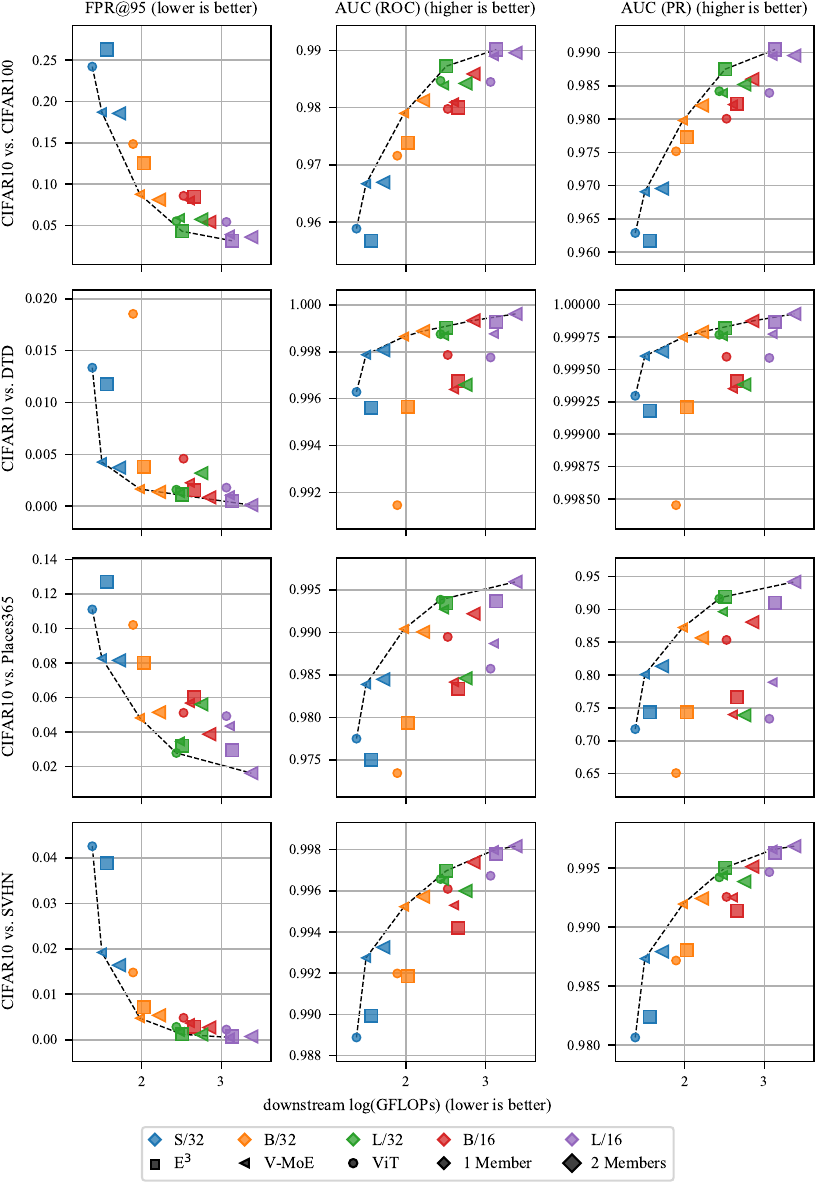}
    \caption{Extended OOD detection the results from \cref{fig:ood_cifar10} with an additional OOD dataset and more metrics.}
    \label{fig:c10_ood_extended}
\end{figure}

\begin{figure}[ht]
    \centering
    \includegraphics[width=0.9\textwidth]{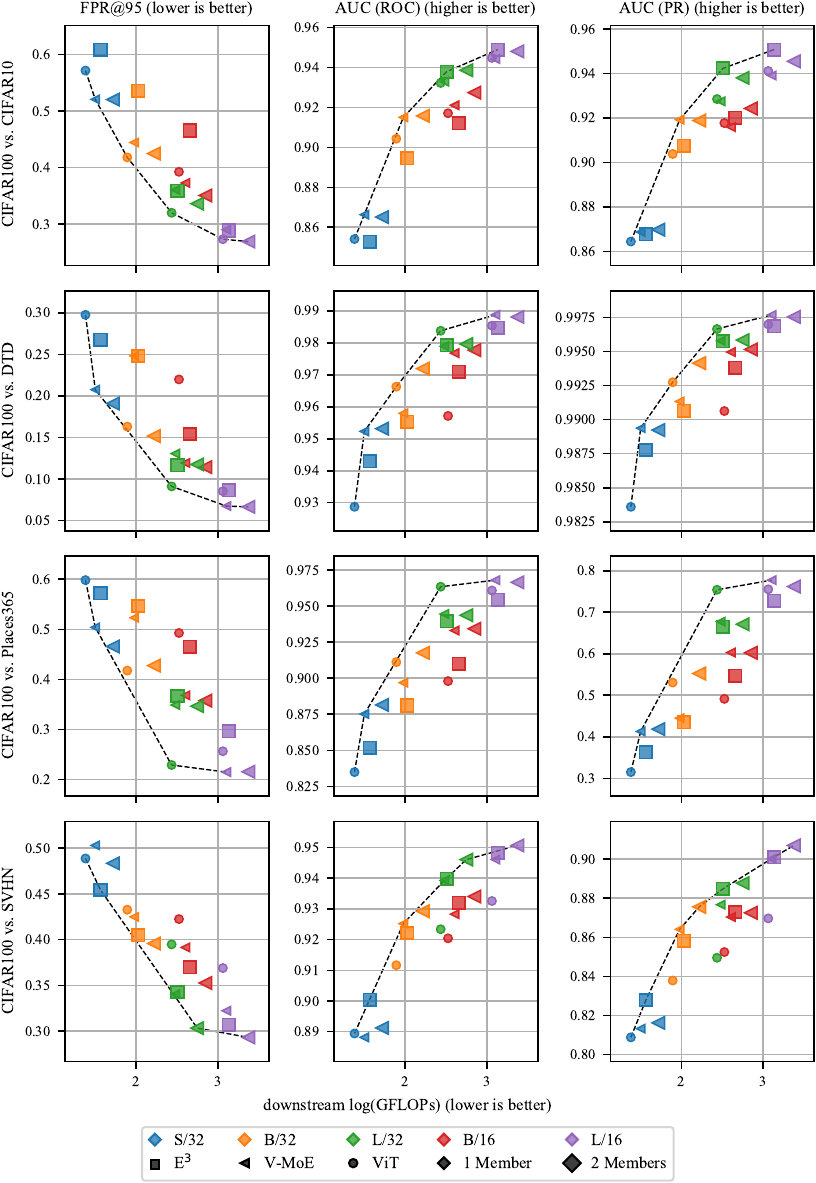}
    \caption{Extended OOD detection the results from \cref{fig:ood_cifar10} with CIFAR100 as the in-distribution dataset, an additional OOD dataset, and more metrics.}
    \label{fig:c100_ood_extended}
\end{figure}

Here we extended the OOD results of \cref{fig:ood_cifar10}. Specifically, we add CIFAR100 as an in-distribution dataset and Describable Textures Dataset (DTD) \citep{cimpoi14describing} as an OOD dataset. We also add area under the receiver operating characteristic (AUC~(ROC)) and area under the precision-recall curve (AUC~(PR)) as metrics. \Cref{fig:c10_ood_extended,fig:c100_ood_extended} contain the results with CIFAR10 and CIFAR100 as the in-distribution datasets, respectively.

As in \cref{fig:ood_cifar10}, we see that \ack performs better (relative to the other baselines) for larger ViT families.
Furthermore, \ack seems to perform better in near OOD detection (i.e. CIFAR10 versus CIFAR100, and vice versa) than far OOD detection. 
Finally, we see that these trends are consistent across OOD metrics.

\subsection{Extended Results for ImageNet} \label{sec:imagenet_extended}

\begin{figure}[ht]
    \centering
    \includegraphics[width=0.9\textwidth]{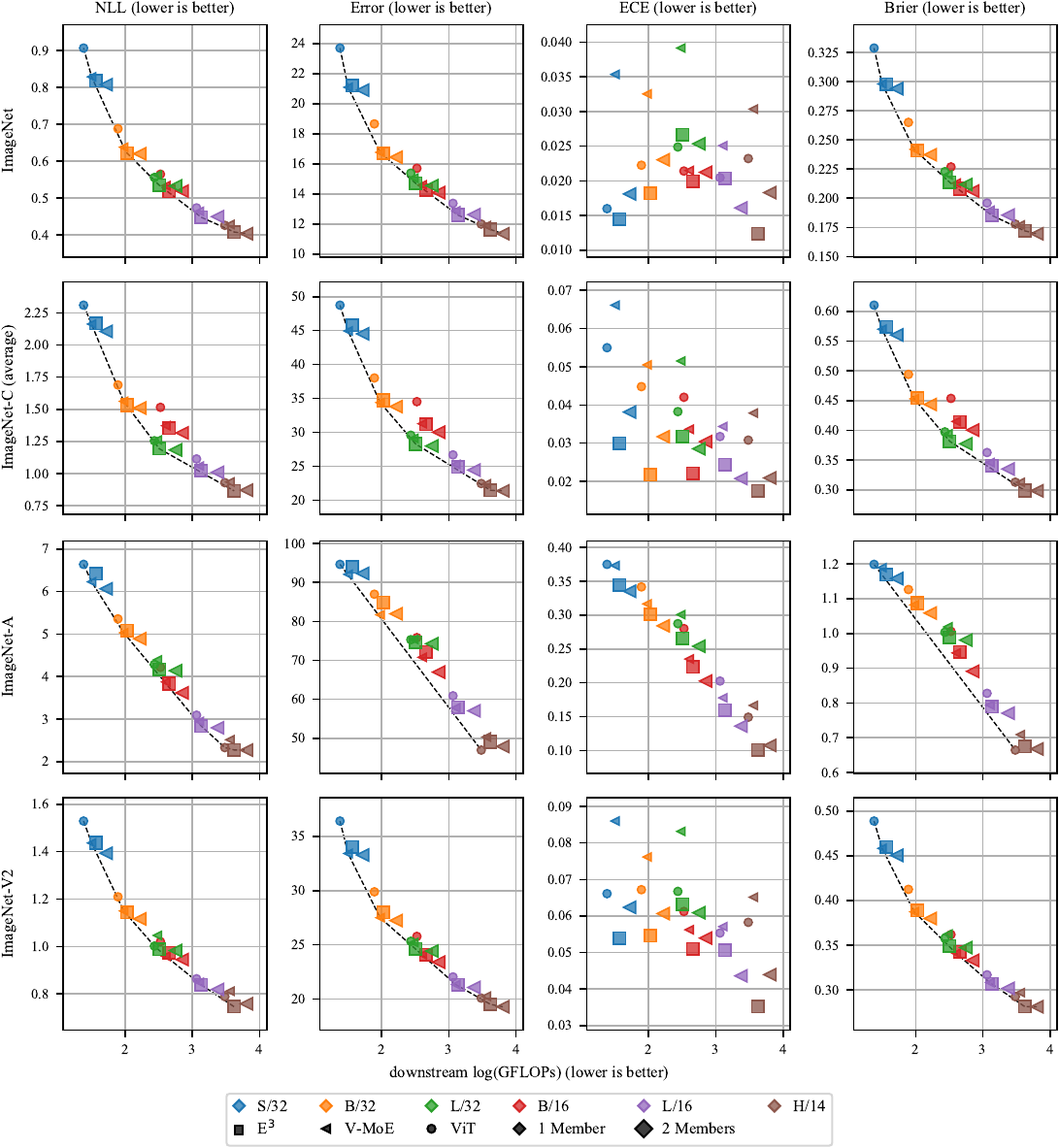}
    \caption{Extended results from \Cref{fig:nll_and_fewshot_zooms,fig:ece_dist_shift,fig:dist_shift} with additional metrics.}
    \label{fig:imagenet_extended}
\end{figure}

In this section we extend the results for ImageNet and the corrupted variants presented in \Cref{fig:nll_and_fewshot_zooms,fig:ece_dist_shift,fig:dist_shift}.
In addition to NLL, classification error, ECE (for standard ImageNet), and Brier score, \cref{fig:imagenet_extended} provides classification error and ECE for all ImageNet variants.

Most of the trends observed in \cref{sec:experiments} remain true:
\begin{itemize}
    \item \ack tends to be Pareto efficient in the presence of distribution shift.
    \item For smaller ViT families, V-MoE outperforms ViT in the presence of distribution shift.
    \item \ack improves ECE over ViT and V-MoE.
    \item \ack improves classification performance.
    \item ViT consistently provides better ECE than V-MoE.
\end{itemize}

However, there are some exceptions:
\begin{itemize}
    \item \textbf{ImageNet-A classification error.} All models (including \ack) under-perform relative to ViT-S/32 and ViT-H/14.
    \item \textbf{ECE for ImageNet-C, ImageNet-A, and ImageNet-V2.} Interestingly for the non-standard ImageNet variants, and in particular for ImageNet-A, there is a strong correlation between lower ECE and larger ViT families.
\end{itemize}

We also find that the results for classification error and Brier score follow those for NLL closely.

\subsection{Additional CIFAR10, CIFAR100, Flowers, and Pets Results}

\begin{figure}[ht]
    \centering
    \includegraphics[width=\textwidth]{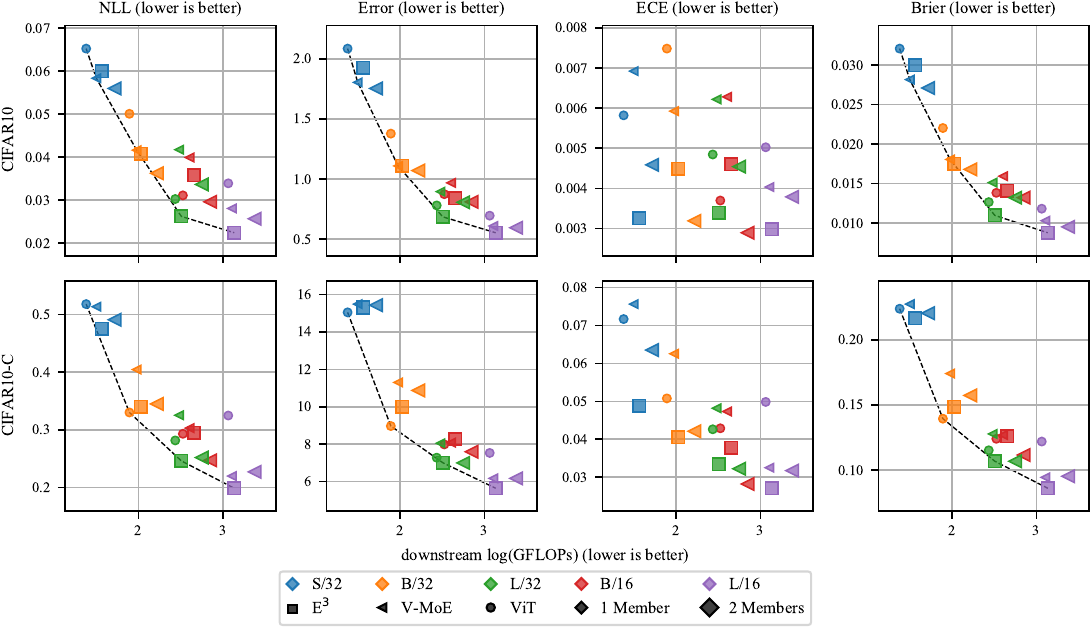}
    \caption{Results for CIFAR10 and CIFAR10-C.}
    \label{fig:cifar10_add_results}
\end{figure}

\begin{figure}[ht]
    \centering
    \includegraphics[width=\textwidth]{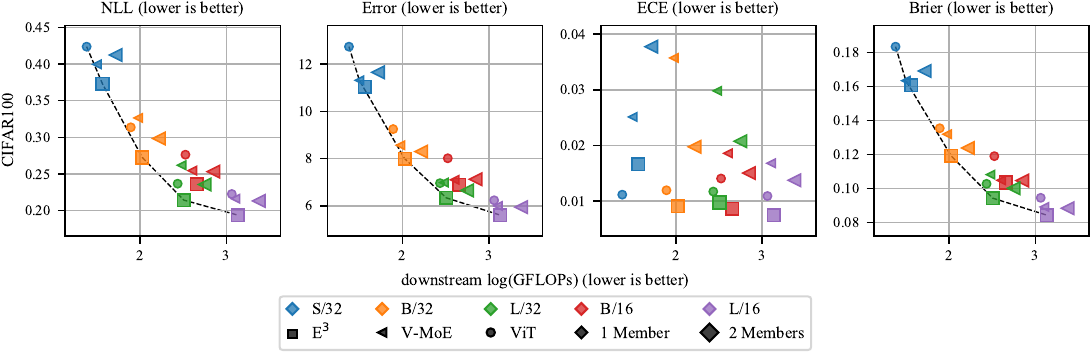}
    \caption{Results for CIFAR100.}
    \label{fig:cifar100_add_results}
\end{figure}

\begin{figure}[ht]
    \centering
    \includegraphics[width=\textwidth]{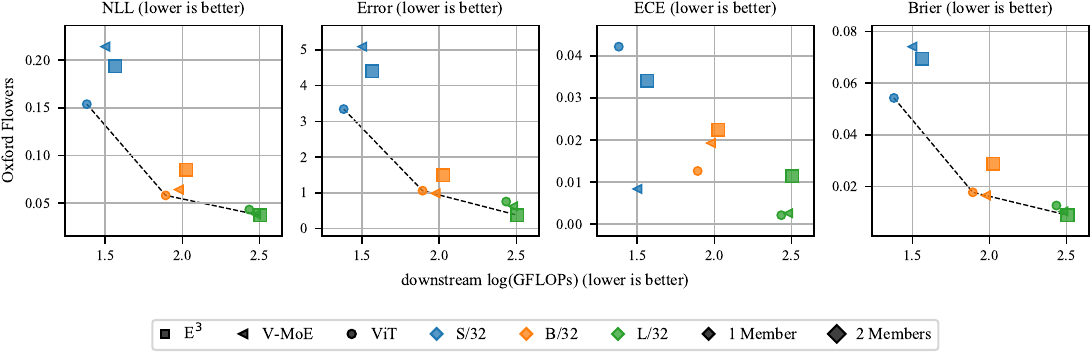}
    \caption{Results for Oxford Flowers 102.}
    \label{fig:flowers_add_results}
\end{figure}

\begin{figure}[ht]
    \centering
    \includegraphics[width=\textwidth]{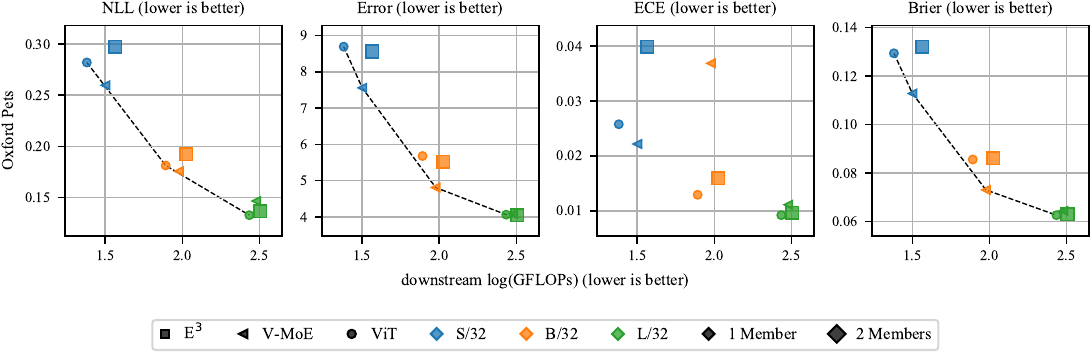}
    \caption{Results for Oxford IIIT Pet.}
    \label{fig:pets_add_results}
\end{figure}

Here we extend the results for ImageNet and the corrupted variants presented in \Cref{fig:nll_and_fewshot_zooms,fig:ece_dist_shift,fig:dist_shift} to four additional datasets. \Cref{fig:cifar10_add_results,fig:cifar100_add_results,,fig:flowers_add_results,fig:pets_add_results} provide results for CIFAR10, CIFAR100, Oxford Flowers 102, and Oxford IIIT Pet, respectively. As in \cref{sec:imagenet_extended}, we find that the results are similar to those in \cref{sec:experiments}. 

Compared to ImageNet, for CIFAR10, CIFAR10-C, and CIFAR100, \ack seems to perform even better relative to the other baselines. Note, for example, that \ack is Pareto efficient (even for S/32) in the cases of CIFAR10-C and CIFAR100 NLL.  As in \cref{sec:imagenet_extended}, we see that the ECE has a stronger downward trend with respect to increased ViT family size for shifted test data.

For Flowers and Pets, where we only have results for smaller ViT families, \ack seems to under perform. However, the performance for L/32 is better than for S/32 and B/32 which suggests that the results for these datasets are consistent with the other datasets presented in this work and, therefore, that we should expect \ack's predictive performance to keep improving with larger models.

\subsection{\MName and V-MoE with larger values of $K$ and $M$}

\begin{figure}[ht]
    \centering
    \includegraphics[width=0.9\textwidth]{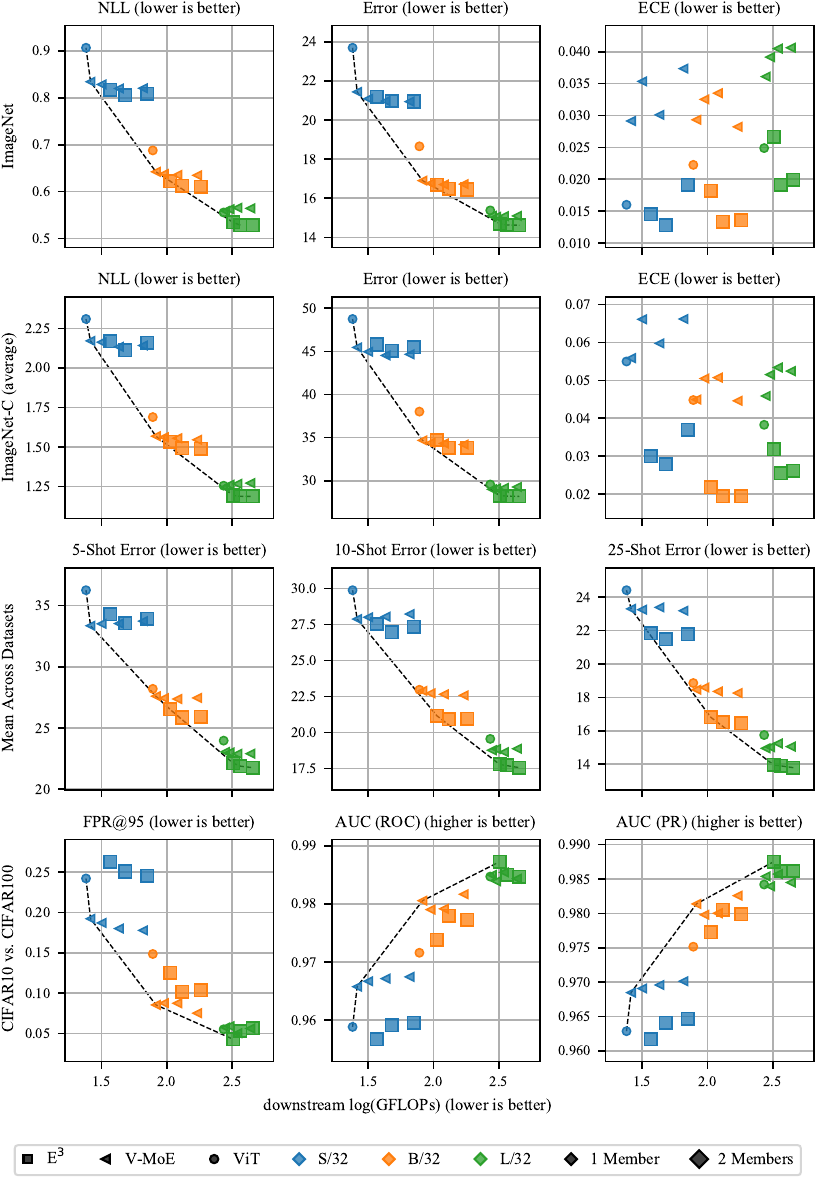}
    \caption{Results for V-MoE with $K \in \{1, 2, 4, 8\}$ and \ack with  $K \in \{1, 2, 4\}$. Models with larger values of $K$ have larger FLOPs.}
    \label{fig:varying_k}
\end{figure}

\begin{figure}[ht]
    \centering
    \includegraphics[width=0.9\textwidth]{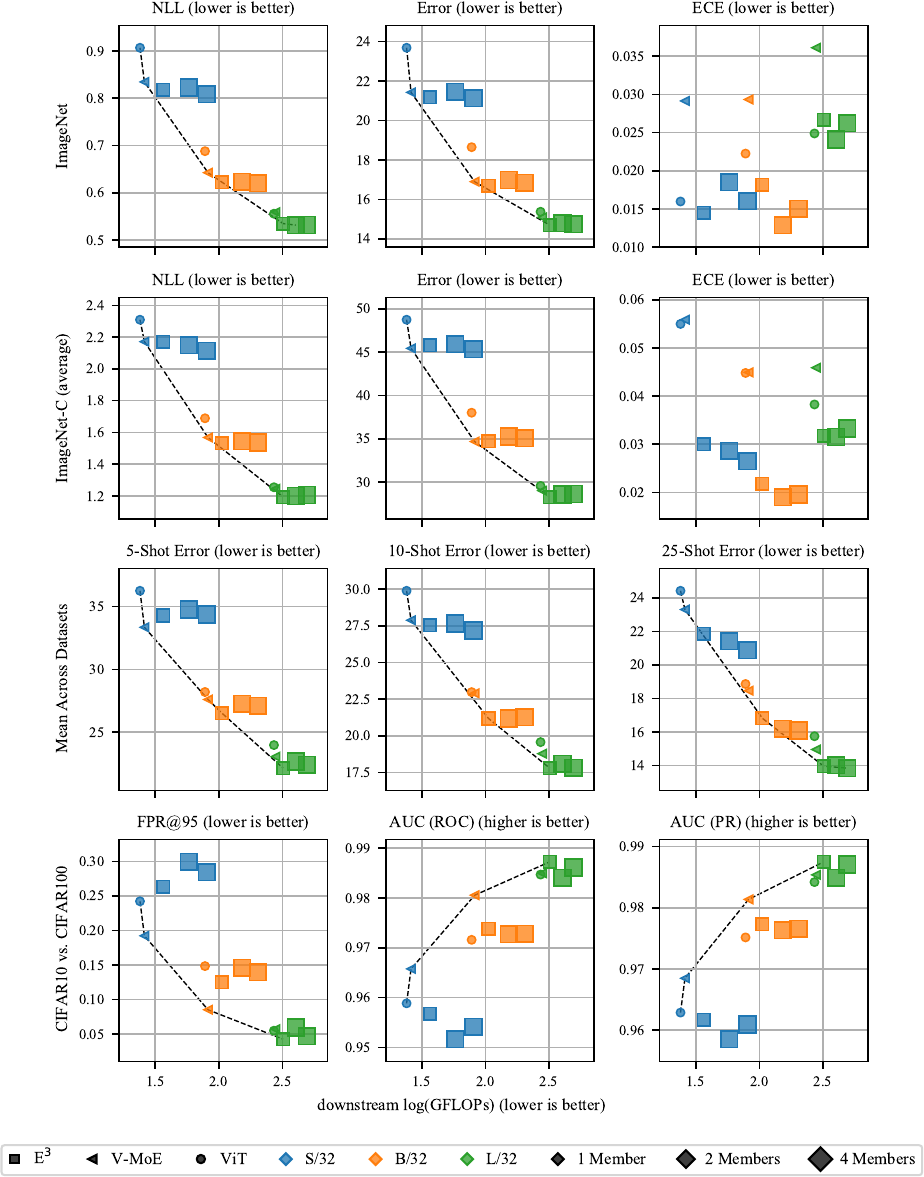}
    \caption{Results for \ack with $M=4$ and  $K \in \{1, 2\}$.}
    \label{fig:pbatch_m4}
\end{figure}

\cref{fig:varying_k} and \cref{fig:pbatch_m4} show the effect of varying $K$ on \ack and V-MoE, and the effect of varying $M$ on \ack, respectively. We make the following observations:
\begin{itemize}
    \item In almost all cases, increasing $K$ or $M$ does not result in Pareto efficient models. 
    \item For V-MoE, increasing $K$ seems to help in most cases, except for ECE performance where it usually hurts.
    \item For \ack, going from $K=1$ to $K = 2$ seems to help in most cases but going from $K = 2$ to $K = 4$ usually hurts. Going from $K = 1$ to $K = 4$ still helps but to a lesser extent than from $K = 1$ to $K = 2$.
    \item For \ack, increasing $M$ either doesn't make a consistent and significant difference or hurts (e.g. in OOD detection).
\end{itemize}

These conclusions should, however, be considered with caution. Recall that the upstream checkpoints used for fine-tuning all V-MoE and \ack models in this work are V-MoE models with $K=2$. Thus, the results in this experiment are confounded by upstream and downstream checkpoint mismatch for all \ack models and all V-MoE models with $K\neq2$. 
This phenomenon was observed in \citet{riquelme2021scaling} for V-MoE models. I.e., performance of downstream V-MoE models with mismatched values of $K$ were relatively worse than those with matched values of $K$; see their Appendix E.4 (Figures 33-35).
We also hypothesise that it is more difficult to train downstream \ack models with larger values of $M$ from upstream V-MoE models because in each subset of the experts some common expert specialisations will need to be duplicated.
Our ensemble members are fine-tuned from an upstream V-MoE with a predefined total number of experts ($E=32$) meaning that increasing $M$ decreases the number of experts available to each ensemble member (with $E/M$ experts per member). This could also impact the performance of \ack with larger sizes of $M$.

\subsubsection{Upstream vs. Downstream Mismatch}

\begin{table}[h]
    \centering
    \caption{ImageNet performance of \ack models fine-tuned from V-MoE-B/32 checkpoints with $K=2$ or $K=4$, and $E=32$. $K_\text{upstream} = 2$ results are averaged over 8 random seeds, while $K_\text{upstream} = 4$ results are averaged over 3 seeds.}
    \label{tab:up_down_mismatch}
    \begin{tabular}{cccccc}
\toprule
                        & K & M & K$_\text{upstream}$ &   NLL $\downarrow$ &   Error $\downarrow$ \\
\midrule
 \multirow{4}{*}{\ack} &    \multirow{2}{*}{1} &   \multirow{2}{*}{2} & 2&  \textbf{0.622} {\tiny \color{gray} $\pm$ 0.001} &  \textbf{16.70} {\tiny \color{gray} $\pm$ 0.03} \\
 &  &  & 4 & \textbf{0.622} {\tiny \color{gray} $\pm$ 0.001} &    16.81 {\tiny \color{gray} $\pm$ 0.05} \\ \cmidrule{2-6}
 &  \multirow{2}{*}{1} &   \multirow{2}{*}{4} &  2 &  0.624 {\tiny \color{gray} $\pm$ 0.001} &  16.99 {\tiny \color{gray} $\pm$ 0.03}  \\ 
 &   &  & 4 &  \textbf{0.622} {\tiny \color{gray} $\pm$ 0.000} &    \textbf{16.93} {\tiny \color{gray} $\pm$ 0.01}\\ 
\bottomrule
\end{tabular}
\end{table}

Here we investigate whether upstream versus downstream mismatch partially explains the counter-intuitive result that increasing $M$ can result in worse performance for \ack. We train downstream \ack models with ($K=1$, $M=2$) and ($K=1$, $M=4$) from upstream V-MoE checkpoints with $K=2$ and $K=4$. \cref{tab:up_down_mismatch} shows the results. 
We see that ($K=1$, $M=2$) does not seem to benefit from increasing $K_\text{upstream}$ from 2 to 4, despite the fact that the upstream $K=4$ model is better than the upstream $K=2$ model (NLL upstream is 8.18 for $K=4$ and 8.27 for $K=2$). On the other hand, ($K=1$, $M=4$) does benefit from having K=4 upstream.

This confirms that the mismatch between upstream and downstream models is one of the factors explaining the results of \ack for growing $M$. However, we also see that the best performing model is ($K_\text{upstream} = 2$, $K=1$, $M=2$), which suggests that there are other confounding factors, such as those mentioned above.

\FloatBarrier
\section{Results Tables} \label{sec:tabular_results}

\subsection{FLOPs Numbers} \label{sec:flops_tables}

\cref{tab:raw_gflops} provides the downstream training FLOPs for various \ack, V-MoE, and ViT configurations. These numbers correspond to the x-values of the points in the figures presented in \Cref{sec:experiments,sec:additional_results}.
\cref{tab:relative_flops} provides the percentage difference in FLOPs between the \ack, V-MoE and down-DE models most commonly used in this work.
Note that the percentage differences for H/14 do not follow the trend of the other sizes, e.g. that the percentage difference between \ack and V-MoE gets smaller for larger sizes, due to the fact that for H/15 we use a last-5 configuration rather than the last-2 configuration used for the other ViT families.
\cref{tab:ablation_flops} provides the downstream training FLOPs for the various ablation study models presented in \cref{sec:pandt_ablations}.\footnote{Note that the \ack and V-MoE results here are different to those in \cref{tab:raw_gflops} due to a difference in implementation. We used a simplified, but less computationally efficient expert-routing implementation for all of the  ablation studies. As a result the V-MoE and \ack FLOPs in \cref{tab:raw_gflops} are lower and cannot be fairly compared with the ablation models presented here. We thus re-benchmarked \ack and V-MoE to obtain comparable results.}

\begin{table}[h]
    \centering
    \caption{Downstream training GFLOPs for the various \ack, V-MoE, and ViT baselines used in this work.}
    \label{tab:raw_gflops}
    \begin{tabular}{ccccccccc}
\toprule
{} &  K &  M &    S/32 &    B/32 &     B/16 &     L/32 &     L/16 &      H/14 \\
\midrule
\multirow{5}{*}{\ack} &  1 &  2 &   36.69 &  105.89 &   452.62 &   320.92 &  1356.46 &   4183.28 \\
 &  1 &  4 &   58.03 &  152.98 &      ---    &   403.77 &      ---    &       ---    \\
 &  2 &  2 &   47.98 &  131.06 &   552.65 &   365.42 &  1533.74 &       ---    \\
 &  2 &  4 &   80.66 &  203.31 &      ---    &   492.77 &      ---    &       ---    \\
 &  4 &  2 &   70.61 &  181.40 &      ---    &   454.43 &      ---    &       ---    \\ \cmidrule{1-9}
\multirow{3}{*}{ViT}   &  - &  1 &   24.01 &   77.97 &   334.48 &   271.83 &  1151.71 &   3033.60 \\
   &  - &  2 &   48.02 &  155.93 &   668.95 &   543.66 &  2303.43 &   6067.21 \\
   &  - &  4 &   96.03 &  311.87 &  1337.90 &  1087.33 &  4606.85 &  12134.41 \\ \cmidrule{1-9}
\multirow{6}{*}{V-MoE} &  1 &  1 &   26.02 &   82.35 &   351.91 &   279.55 &  1182.08 &   3179.90 \\
 &  1 &  2 &   52.04 &  164.70 &   703.81 &   559.10 &  2364.16 &   6359.81 \\
 &  1 &  4 &  104.07 &  329.41 &  1407.63 &  1118.21 &  4728.32 &  12719.61 \\
 &  2 &  1 &   31.66 &   94.93 &   401.98 &   301.75 &  1270.78 &   3617.94 \\
 &  4 &  1 &   42.95 &  120.11 &   501.99 &   346.25 &  1448.06 &       ---    \\
 &  8 &  1 &   65.59 &  170.44 &      ---    &   435.26 &      ---    &       ---    \\
\bottomrule
\end{tabular}

\end{table}

\begin{table}[h]
    \centering
        \caption{Percentage difference in downstream training FLOPs for \ack with $(K, M) = (1, 2)$ compared with V-MoE with $K = 1$ and an ensemble of two such V-MoE members. }
    \label{tab:relative_flops}
\begin{tabular}{ccccccc}
\toprule
{} &    S/32 &    B/32 &    B/16 &    L/32 &    L/16 &    H/14 \\
\midrule
\ack vs. V-MoE    &   41.01 &   28.58 &   28.62 &   14.80 &   14.75 &   31.55 \\
\ack vs. down-DE &  -29.49 &  -35.71 &  -35.69 &  -42.60 &  -42.62 &  -34.22 \\
\bottomrule
\end{tabular}
\end{table}

\begin{table}[h]
    \centering
    \caption{Downstream training GFLOPs comparison for the ablation study models in \cref{sec:pandt_ablations}.}
    \label{tab:ablation_flops}
    \begin{tabular}{cccc}
\toprule
                            &  \textsc{K} & \textsc{M} & \textsc{GFLOPs} \\
\midrule
             V-MoE & 2 & --- &   96.895 \\
             V-MoE & 4 & ---  &  123.644 \\
             V-MoE & 8 & ---  &  178.133 \\ \cmidrule{1-4}
        \ack & 1 & 2  &  109.781 \\
        \ack & 2 & 2  &  138.457 \\
        \ack & 4 & 2  &  196.870 \\
        \cmidrule{1-4}
 
    Tiling & 2 & 2  &  138.460 \\ \cmidrule{1-4}
 Partitioning & 2 & --- &  97.885 \\ \cmidrule{1-4}
     Overlap = 2 & 2 & 2 &  138.457 \\
     Overlap = 4 & 2 & 2  &  138.458 \\
     Overlap = 8 & 2 & 2  &  138.459 \\
    Overlap = 6 & 2 & 2  &  138.460 \\ \cmidrule{1-4}
   Multi-pred & 2 & ---  &   96.330 \\
   Multi-pred & 4 & ---   &  122.526 \\
   Multi-pred & 8 & ---   &  175.889 \\

\bottomrule
\end{tabular}
\end{table}

\subsection{Parameter Counts}

\cref{tab:param_counts} compares the parameter counts for ViT and V-MoE/\ack models in each ViT family.

\begin{table}[htbp]
    \centering
    \caption{Parameter counts for ViT vs V-MoE and \ack.}
    \label{tab:param_counts}
    \begin{tabular}{ccccccc}
    \toprule
         &  S/32 & B/32 & B/16 & L/32 & L16 & H/14 \\ \midrule
      ViT   & 36.5M & 102.1M & 100.5M & 325.3M & 323.1M & 655.8M \\
      V-MoE/\ack & 166.7M & 395.0M & 393.3M & 845.8M & 843.6M & 2688.6M \\
    \bottomrule
    \end{tabular}
\end{table}

\subsection{Summary for NLL under Distribution Shift} \label{sec:nll_shift_summary}

\cref{tab:nll_shift_summary} shows the percentage improvement for \ack versus V-MoE in NLL (i.e., $\frac{\text{NLL}_\text{V-MoE} - \text{NLL}_\text{\ack}}{\text{NLL}_\text{V-MoE}} \times 100$, with positive values thus indicating that \ack improves upon V-MoE) averaged over ImageNet-C, ImageNet-A, and ImageNet-V2, for a given ViT family and ($K$, $M$) in \{(1, 2), (2, 2)\} (compared to V-MoE with $K=2$ and $K=4$, respectively). We see that for all ViT families except S/32, \ack outperforms V-MoE. This result is not unexpected, since ensembles tend to provide improved performance under distribution shift relative to single models \citep{Lakshminarayanan2017,snoek2019can,havasi2020training}.

\begin{table}[h]
    \centering
    \caption{Percentage improvement for \ack vs V-MoE in NLL under distribution shift, averaged over ImageNet-C, ImageNet-A, and ImageNet-V2.}
    \label{tab:nll_shift_summary}
\begin{tabular}{ccccccc}
    \toprule
         &  S/32 & B/32 & B/16 & L/32 & L/16 & H/14 \\ \midrule
        \ack ($K=1,M=2$) vs. V-MoE ($K=2$) & -1.15 & 0.52 & 0.31 & 5.34 & 2.78 & 8.33 \\
        \ack ($K=2,M=2$) vs. V-MoE ($K=4$) & -0.21 & 3.91 & 1.33 & 7.37 & 2.27 & --- \\
        \bottomrule
    \end{tabular}
\end{table}

\subsection{Auxiliary Tables with Standard Errors.} \label{sec:std_err_tables}

The tables in this section provide the mean values and corresponding standard errors for many of the results depicted in figures throughout \cref{sec:experiments,sec:additional_results}.

\begin{landscape}
\begin{table}[]
    \centering
    \caption{ImageNet comparison of V-MoE, downstream ensembles there-of, and \ack with 2 experts per input in each case.}
    \label{tab:imagenet_k2}
    \resizebox{\linewidth}{!}{
\begin{tabular}{ccccccccccccccccc}
\toprule
     &       & \multicolumn{2}{c}{\;} & \multicolumn{3}{c}{\textsc{ImageNet}} & \multicolumn{3}{c}{\textsc{ImageNet-C (average)}} & \multicolumn{3}{c}{\textsc{ImageNet-A}} & \multicolumn{3}{c}{\textsc{ImageNet-V2}} &                      \textsc{Downstream} \\
     &       & \textsc{K} & \textsc{M} &                        \textsc{NLL} $\downarrow$ &                     \textsc{Error} $\downarrow$ &                        \textsc{ECE} $\downarrow$ &                        \textsc{NLL} $\downarrow$ &                     \textsc{Error} $\downarrow$ &                        \textsc{ECE} $\downarrow$ &                        \textsc{NLL} $\downarrow$ &                     \textsc{Error} $\downarrow$ &                        \textsc{ECE} $\downarrow$ &                        \textsc{NLL} $\downarrow$ &                     \textsc{Error} $\downarrow$ &                        \textsc{ECE} $\downarrow$ & $\Delta$\textsc{FLOPs} (\%) $\downarrow$ \\
\midrule
\multirow{3}{*}{H/14} & \ack &          1 &          2 &           0.408 {\tiny \color{gray} $\pm$ 0.001} &           11.63 {\tiny \color{gray} $\pm$ 0.05} &  \textbf{0.012} {\tiny \color{gray} $\pm$ 0.000} &  \textbf{0.865} {\tiny \color{gray} $\pm$ 0.010} &  \textbf{21.46} {\tiny \color{gray} $\pm$ 0.16} &  \textbf{0.018} {\tiny \color{gray} $\pm$ 0.000} &  \textbf{2.276} {\tiny \color{gray} $\pm$ 0.042} &  \textbf{49.09} {\tiny \color{gray} $\pm$ 0.78} &  \textbf{0.101} {\tiny \color{gray} $\pm$ 0.003} &  \textbf{0.745} {\tiny \color{gray} $\pm$ 0.001} &  \textbf{19.50} {\tiny \color{gray} $\pm$ 0.08} &  \textbf{0.035} {\tiny \color{gray} $\pm$ 0.001} &                           \textbf{15.63} \\
\bottomrule
\end{tabular}
}
\end{table}

\begin{table}[]
    \centering
    \caption{ImageNet comparison of V-MoE, downstream ensembles there-of, and \ack with 4 experts per input in each case.}
    \label{tab:imagenet_k4}
    \resizebox{\linewidth}{!}{
\begin{tabular}{ccccccccccccccccc}
\toprule
     &       & \multicolumn{2}{c}{\;} & \multicolumn{3}{c}{\textsc{ImageNet}} & \multicolumn{3}{c}{\textsc{ImageNet-C (average)}} & \multicolumn{3}{c}{\textsc{ImageNet-A}} & \multicolumn{3}{c}{\textsc{ImageNet-V2}} &                      \textsc{Downstream} \\
     &       & \textsc{K} & \textsc{M} &                        \textsc{NLL} $\downarrow$ &                     \textsc{Error} $\downarrow$ &                        \textsc{ECE} $\downarrow$ &                        \textsc{NLL} $\downarrow$ &                     \textsc{Error} $\downarrow$ &                        \textsc{ECE} $\downarrow$ &                        \textsc{NLL} $\downarrow$ &                     \textsc{Error} $\downarrow$ &                        \textsc{ECE} $\downarrow$ &                        \textsc{NLL} $\downarrow$ &                     \textsc{Error} $\downarrow$ &                        \textsc{ECE} $\downarrow$ & $\Delta$\textsc{FLOPs} (\%) $\downarrow$ \\
\midrule
\multirow{3}{*}{L/16} & \ack &          2 &          2 &           0.451 {\tiny \color{gray} $\pm$ 0.001} &           12.61 {\tiny \color{gray} $\pm$ 0.04} &           0.023 {\tiny \color{gray} $\pm$ 0.000} &           1.028 {\tiny \color{gray} $\pm$ 0.003} &           24.90 {\tiny \color{gray} $\pm$ 0.07} &           0.028 {\tiny \color{gray} $\pm$ 0.000} &           2.886 {\tiny \color{gray} $\pm$ 0.016} &           58.26 {\tiny \color{gray} $\pm$ 0.24} &           0.171 {\tiny \color{gray} $\pm$ 0.002} &           0.849 {\tiny \color{gray} $\pm$ 0.002} &           21.40 {\tiny \color{gray} $\pm$ 0.11} &           0.055 {\tiny \color{gray} $\pm$ 0.001} &                            \textbf{5.92} \\
\bottomrule
\end{tabular}
}
\end{table}

\begin{table}[]
    \centering
    \caption{ImageNet comparison of V-MoE and ViT.}
    \label{tab:imagenet_vmoe_vit}
    \resizebox{\linewidth}{!}{
\begin{tabular}{cccccccccccccccc}
\toprule
     &     & \multicolumn{2}{c}{\;} & \multicolumn{3}{c}{\textsc{ImageNet}} & \multicolumn{3}{c}{\textsc{ImageNet-C}} & \multicolumn{3}{c}{\textsc{ImageNet-A}} & \multicolumn{3}{c}{\textsc{ImageNet-V2}} \\
     &     & \textsc{K} & \textsc{M} &                        \textsc{NLL} $\downarrow$ &                     \textsc{Error} $\downarrow$ &                        \textsc{ECE} $\downarrow$ &                        \textsc{NLL} $\downarrow$ &                     \textsc{Error} $\downarrow$ &                        \textsc{ECE} $\downarrow$ &                        \textsc{NLL} $\downarrow$ &                     \textsc{Error} $\downarrow$ &                        \textsc{ECE} $\downarrow$ &                        \textsc{NLL} $\downarrow$ &                     \textsc{Error} $\downarrow$ &                        \textsc{ECE} $\downarrow$ \\
\midrule
\multirow{2}{*}{H/14} & V-MoE &          1 &          1 &  \textbf{0.426} {\tiny \color{gray} $\pm$ 0.002} &  \textbf{11.81} {\tiny \color{gray} $\pm$ 0.08} &           0.026 {\tiny \color{gray} $\pm$ 0.000} &  \textbf{0.932} {\tiny \color{gray} $\pm$ 0.009} &  \textbf{22.43} {\tiny \color{gray} $\pm$ 0.16} &           0.033 {\tiny \color{gray} $\pm$ 0.000} &           2.494 {\tiny \color{gray} $\pm$ 0.057} &           50.22 {\tiny \color{gray} $\pm$ 1.02} &  \textbf{0.153} {\tiny \color{gray} $\pm$ 0.004} &  \textbf{0.806} {\tiny \color{gray} $\pm$ 0.005} &  \textbf{20.08} {\tiny \color{gray} $\pm$ 0.13} &  \textbf{0.059} {\tiny \color{gray} $\pm$ 0.001} \\
\bottomrule
\end{tabular}
}
\end{table}
\end{landscape}

\begin{table}[]
    \centering
    \caption{CIFAR10 comparison of V-MoE, downstream ensembles there-of, and \ack with 2 experts per input in each case.}
    \label{tab:cifar10_k2}
    \resizebox{\linewidth}{!}{
\begin{tabular}{ccccccccccc}
\toprule
     &       & \multicolumn{2}{c}{\;} & \multicolumn{3}{c}{\textsc{Cifar10}} & \multicolumn{3}{c}{\textsc{Cifar10-C}} &                      \textsc{Downstream} \\
     &       & \textsc{K} & \textsc{M} &                        \textsc{NLL} $\downarrow$ &                    \textsc{Error} $\downarrow$ &                        \textsc{ECE} $\downarrow$ &                        \textsc{NLL} $\downarrow$ &                     \textsc{Error} $\downarrow$ &                        \textsc{ECE} $\downarrow$ & $\Delta$\textsc{FLOPs} (\%) $\downarrow$ \\
\midrule
\multirow{3}{*}{L/16} & E$^3$ &          1 &          2 &  \textbf{0.022} {\tiny \color{gray} $\pm$ 0.000} &  \textbf{0.55} {\tiny \color{gray} $\pm$ 0.02} &  \textbf{0.003} {\tiny \color{gray} $\pm$ 0.000} &  \textbf{0.198} {\tiny \color{gray} $\pm$ 0.012} &   \textbf{5.65} {\tiny \color{gray} $\pm$ 0.20} &  \textbf{0.027} {\tiny \color{gray} $\pm$ 0.002} &                            \textbf{6.74} \\
     & down-DE &          1 &          2 &           0.026 {\tiny \color{gray} $\pm$ 0.001} &  \textbf{0.59} {\tiny \color{gray} $\pm$ 0.04} &           0.004 {\tiny \color{gray} $\pm$ 0.000} &  \textbf{0.227} {\tiny \color{gray} $\pm$ 0.011} &   \textbf{6.18} {\tiny \color{gray} $\pm$ 0.21} &  \textbf{0.032} {\tiny \color{gray} $\pm$ 0.003} &                                    86.04 \\
     & V-MoE &          2 &          1 &           0.028 {\tiny \color{gray} $\pm$ 0.000} &  \textbf{0.61} {\tiny \color{gray} $\pm$ 0.01} &           0.004 {\tiny \color{gray} $\pm$ 0.000} &  \textbf{0.220} {\tiny \color{gray} $\pm$ 0.012} &   \textbf{6.17} {\tiny \color{gray} $\pm$ 0.23} &  \textbf{0.032} {\tiny \color{gray} $\pm$ 0.003} &                                      --- \\
\bottomrule
\end{tabular}
}
\end{table}

\begin{table}[]
    \centering
    \caption{CIFAR10 comparison of V-MoE, downstream ensembles there-of, and \ack with 4 experts per input in each case.}
    \label{tab:cifar10_k4}
    \resizebox{\linewidth}{!}{
\begin{tabular}{ccccccccccc}
\toprule
     &       & \multicolumn{2}{c}{\;} & \multicolumn{3}{c}{\textsc{Cifar10}} & \multicolumn{3}{c}{\textsc{Cifar10-C}} &                      \textsc{Downstream} \\
     &       & \textsc{K} & \textsc{M} &                        \textsc{NLL} $\downarrow$ &                    \textsc{Error} $\downarrow$ &                        \textsc{ECE} $\downarrow$ &                        \textsc{NLL} $\downarrow$ &                     \textsc{Error} $\downarrow$ &                        \textsc{ECE} $\downarrow$ & $\Delta$\textsc{FLOPs} (\%) $\downarrow$ \\
\midrule
\multirow{3}{*}{L/16} & E$^3$ &          2 &          2 &  \textbf{0.026} {\tiny \color{gray} $\pm$ 0.001} &  \textbf{0.58} {\tiny \color{gray} $\pm$ 0.02} &  \textbf{0.003} {\tiny \color{gray} $\pm$ 0.000} &  \textbf{0.247} {\tiny \color{gray} $\pm$ 0.016} &   \textbf{6.56} {\tiny \color{gray} $\pm$ 0.26} &  \textbf{0.036} {\tiny \color{gray} $\pm$ 0.003} &                            \textbf{5.92} \\
     & down-DE &          1 &          4 &  \textbf{0.025} {\tiny \color{gray} $\pm$ 0.001} &  \textbf{0.57} {\tiny \color{gray} $\pm$ 0.03} &  \textbf{0.004} {\tiny \color{gray} $\pm$ 0.000} &  \textbf{0.219} {\tiny \color{gray} $\pm$ 0.010} &   \textbf{6.12} {\tiny \color{gray} $\pm$ 0.19} &  \textbf{0.030} {\tiny \color{gray} $\pm$ 0.003} &                                   226.53 \\
     & V-MoE &          4 &          1 &           0.034 {\tiny \color{gray} $\pm$ 0.001} &           0.75 {\tiny \color{gray} $\pm$ 0.02} &           0.005 {\tiny \color{gray} $\pm$ 0.000} &           0.278 {\tiny \color{gray} $\pm$ 0.010} &            7.71 {\tiny \color{gray} $\pm$ 0.19} &           0.043 {\tiny \color{gray} $\pm$ 0.002} &                                      --- \\
\bottomrule
\end{tabular}
}
\end{table}

\begin{table}[]
    \centering
    \caption{CIFAR10 comparison of V-MoE and ViT.}
    \label{tab:cifar10_vmoe_vit}
    \begin{tabular}{cccccccccc}
\toprule
     &     & \multicolumn{2}{c}{\;} & \multicolumn{3}{c}{\textsc{Cifar10}} & \multicolumn{3}{c}{\textsc{Cifar10-C}} \\
     &     & \textsc{K} & \textsc{M} &                        \textsc{NLL} $\downarrow$ &                    \textsc{Error} $\downarrow$ &                        \textsc{ECE} $\downarrow$ &                        \textsc{NLL} $\downarrow$ &                     \textsc{Error} $\downarrow$ &                        \textsc{ECE} $\downarrow$ \\
\midrule
\multirow{2}{*}{L/16} & V-MoE &          1 &          1 &  \textbf{0.029} {\tiny \color{gray} $\pm$ 0.001} &  \textbf{0.59} {\tiny \color{gray} $\pm$ 0.02} &  \textbf{0.004} {\tiny \color{gray} $\pm$ 0.000} &  \textbf{0.236} {\tiny \color{gray} $\pm$ 0.011} &   \textbf{6.25} {\tiny \color{gray} $\pm$ 0.20} &  \textbf{0.034} {\tiny \color{gray} $\pm$ 0.002} \\
     & ViT &          - &          1 &           0.034 {\tiny \color{gray} $\pm$ 0.001} &  \textbf{0.69} {\tiny \color{gray} $\pm$ 0.03} &  \textbf{0.005} {\tiny \color{gray} $\pm$ 0.000} &           0.325 {\tiny \color{gray} $\pm$ 0.020} &            7.53 {\tiny \color{gray} $\pm$ 0.40} &           0.050 {\tiny \color{gray} $\pm$ 0.004} \\
\cmidrule{1-10}
\multirow{2}{*}{L/32} & V-MoE &          1 &          1 &           0.040 {\tiny \color{gray} $\pm$ 0.001} &  \textbf{0.86} {\tiny \color{gray} $\pm$ 0.02} &           0.006 {\tiny \color{gray} $\pm$ 0.000} &  \textbf{0.290} {\tiny \color{gray} $\pm$ 0.013} &   \textbf{7.35} {\tiny \color{gray} $\pm$ 0.22} &  \textbf{0.043} {\tiny \color{gray} $\pm$ 0.003} \\
     & ViT &          - &          1 &  \textbf{0.030} {\tiny \color{gray} $\pm$ 0.001} &  \textbf{0.78} {\tiny \color{gray} $\pm$ 0.02} &  \textbf{0.005} {\tiny \color{gray} $\pm$ 0.000} &  \textbf{0.281} {\tiny \color{gray} $\pm$ 0.010} &   \textbf{7.28} {\tiny \color{gray} $\pm$ 0.13} &  \textbf{0.043} {\tiny \color{gray} $\pm$ 0.002} \\
\bottomrule
\end{tabular}

\end{table}

\begin{landscape}
\begin{table}[]
    \centering
        \caption{CIFAR10 OOD comparison of V-MoE, downstream ensembles there-of, and \ack with 2 experts per input in each case.}
    \label{tab:cifar10_ood}
    \resizebox{\linewidth}{!}{
\begin{tabular}{cccccccccccccccc}
\toprule
     &       & \multicolumn{2}{c}{\;} & \multicolumn{3}{c}{\textsc{CIFAR10 vs. CIFAR100}} & \multicolumn{3}{c}{\textsc{CIFAR10 vs. DTD}} & \multicolumn{3}{c}{\textsc{CIFAR10 vs. Places365}} & \multicolumn{3}{c}{\textsc{CIFAR10 vs. SVHN}} \\
     &       & \textsc{K} & \textsc{M} &                       \textsc{AUC (PR)} $\uparrow$ &                      \textsc{AUC (ROC)} $\uparrow$ &                       \textsc{FPR@95} $\downarrow$ &                       \textsc{AUC (PR)} $\uparrow$ &                      \textsc{AUC (ROC)} $\uparrow$ &                       \textsc{FPR@95} $\downarrow$ &                       \textsc{AUC (PR)} $\uparrow$ &                      \textsc{AUC (ROC)} $\uparrow$ &                       \textsc{FPR@95} $\downarrow$ &                       \textsc{AUC (PR)} $\uparrow$ &                      \textsc{AUC (ROC)} $\uparrow$ &                       \textsc{FPR@95} $\downarrow$ \\
\midrule
\multirow{3}{*}{L/16} & \ack &          1 &          2 &  \textbf{0.9905} {\tiny \color{gray} $\pm$ 0.0003} &  \textbf{0.9902} {\tiny \color{gray} $\pm$ 0.0003} &  \textbf{0.0313} {\tiny \color{gray} $\pm$ 0.0010} &           0.9999 {\tiny \color{gray} $\pm$ 0.0000} &           0.9993 {\tiny \color{gray} $\pm$ 0.0000} &           0.0005 {\tiny \color{gray} $\pm$ 0.0001} &           0.9103 {\tiny \color{gray} $\pm$ 0.0068} &           0.9936 {\tiny \color{gray} $\pm$ 0.0003} &           0.0294 {\tiny \color{gray} $\pm$ 0.0013} &  \textbf{0.9963} {\tiny \color{gray} $\pm$ 0.0002} &  \textbf{0.9978} {\tiny \color{gray} $\pm$ 0.0001} &  \textbf{0.0007} {\tiny \color{gray} $\pm$ 0.0002} \\
\bottomrule
\end{tabular}
}
\end{table}

\begin{table}[]
    \centering
        \caption{CIFAR10 OOD comparison of ViT and V-MoE}
    \label{tab:cifar10_ood_vmoe_vit}
    \resizebox{\linewidth}{!}{
\begin{tabular}{cccccccccccccccc}
\toprule
     &     & \multicolumn{2}{c}{\;} & \multicolumn{3}{c}{\textsc{CIFAR10 vs. CIFAR100}} & \multicolumn{3}{c}{\textsc{CIFAR10 vs. DTD}} & \multicolumn{3}{c}{\textsc{CIFAR10 vs. Places365}} & \multicolumn{3}{c}{\textsc{CIFAR10 vs. SVHN}} \\
     &     & \textsc{K} & \textsc{M} &                       \textsc{AUC (PR)} $\uparrow$ &                      \textsc{AUC (ROC)} $\uparrow$ &                       \textsc{FPR@95} $\downarrow$ &                       \textsc{AUC (PR)} $\uparrow$ &                      \textsc{AUC (ROC)} $\uparrow$ &                       \textsc{FPR@95} $\downarrow$ &                       \textsc{AUC (PR)} $\uparrow$ &                      \textsc{AUC (ROC)} $\uparrow$ &                       \textsc{FPR@95} $\downarrow$ &                       \textsc{AUC (PR)} $\uparrow$ &                      \textsc{AUC (ROC)} $\uparrow$ &                       \textsc{FPR@95} $\downarrow$ \\
\midrule
\multirow{2}{*}{L/16} & V-MoE &          1 &          1 &  \textbf{0.9891} {\tiny \color{gray} $\pm$ 0.0004} &  \textbf{0.9895} {\tiny \color{gray} $\pm$ 0.0004} &  \textbf{0.0379} {\tiny \color{gray} $\pm$ 0.0022} &  \textbf{0.9999} {\tiny \color{gray} $\pm$ 0.0000} &  \textbf{0.9997} {\tiny \color{gray} $\pm$ 0.0000} &  \textbf{0.0001} {\tiny \color{gray} $\pm$ 0.0001} &  \textbf{0.9400} {\tiny \color{gray} $\pm$ 0.0057} &  \textbf{0.9960} {\tiny \color{gray} $\pm$ 0.0002} &  \textbf{0.0162} {\tiny \color{gray} $\pm$ 0.0013} &  \textbf{0.9972} {\tiny \color{gray} $\pm$ 0.0001} &  \textbf{0.9984} {\tiny \color{gray} $\pm$ 0.0001} &  \textbf{0.0007} {\tiny \color{gray} $\pm$ 0.0001} \\
\bottomrule
\end{tabular}
}
\end{table}

\begin{table}[]
    \centering
    \caption{CIFAR100 OOD comparison of V-MoE, downstream ensembles there-of, and \ack with 2 experts per input in each case.}
    \label{tab:cifar100_ood}
    \resizebox{\linewidth}{!}{
\begin{tabular}{cccccccccccccccc}
\toprule
     &       & \multicolumn{2}{c}{\;} & \multicolumn{3}{c}{\textsc{CIFAR100 vs. CIFAR10}} & \multicolumn{3}{c}{\textsc{CIFAR100 vs. DTD}} & \multicolumn{3}{c}{\textsc{CIFAR100 vs. Places365}} & \multicolumn{3}{c}{\textsc{CIFAR100 vs. SVHN}} \\
     &       & \textsc{K} & \textsc{M} &                       \textsc{AUC (PR)} $\uparrow$ &                      \textsc{AUC (ROC)} $\uparrow$ &                       \textsc{FPR@95} $\downarrow$ &                       \textsc{AUC (PR)} $\uparrow$ &                      \textsc{AUC (ROC)} $\uparrow$ &                       \textsc{FPR@95} $\downarrow$ &                       \textsc{AUC (PR)} $\uparrow$ &                      \textsc{AUC (ROC)} $\uparrow$ &                       \textsc{FPR@95} $\downarrow$ &                       \textsc{AUC (PR)} $\uparrow$ &                      \textsc{AUC (ROC)} $\uparrow$ &                       \textsc{FPR@95} $\downarrow$ \\
\midrule
\multirow{3}{*}{L/16} & \ack &          1 &          2 &  \textbf{0.9507} {\tiny \color{gray} $\pm$ 0.0012} &  \textbf{0.9490} {\tiny \color{gray} $\pm$ 0.0011} &  \textbf{0.2889} {\tiny \color{gray} $\pm$ 0.0058} &           0.9969 {\tiny \color{gray} $\pm$ 0.0001} &           0.9847 {\tiny \color{gray} $\pm$ 0.0002} &           0.0862 {\tiny \color{gray} $\pm$ 0.0010} &           0.7281 {\tiny \color{gray} $\pm$ 0.0023} &           0.9542 {\tiny \color{gray} $\pm$ 0.0007} &           0.2969 {\tiny \color{gray} $\pm$ 0.0036} &  \textbf{0.9011} {\tiny \color{gray} $\pm$ 0.0057} &  \textbf{0.9482} {\tiny \color{gray} $\pm$ 0.0024} &  \textbf{0.3071} {\tiny \color{gray} $\pm$ 0.0093} \\
\bottomrule
\end{tabular}
}
\end{table}

\begin{table}[]
    \centering
    \caption{CIFAR100 OOD comparison of V-MoE and ViT}
    \label{tab:cifar100_ood_vmoe_vit}
    \resizebox{\linewidth}{!}{
\begin{tabular}{cccccccccccccccc}
\toprule
     &     & \multicolumn{2}{c}{\;} & \multicolumn{3}{c}{\textsc{CIFAR100 vs. CIFAR10}} & \multicolumn{3}{c}{\textsc{CIFAR100 vs. DTD}} & \multicolumn{3}{c}{\textsc{CIFAR100 vs. Places365}} & \multicolumn{3}{c}{\textsc{CIFAR100 vs. SVHN}} \\
     &     & \textsc{K} & \textsc{M} &                       \textsc{AUC (PR)} $\uparrow$ &                      \textsc{AUC (ROC)} $\uparrow$ &                       \textsc{FPR@95} $\downarrow$ &                       \textsc{AUC (PR)} $\uparrow$ &                      \textsc{AUC (ROC)} $\uparrow$ &                       \textsc{FPR@95} $\downarrow$ &                       \textsc{AUC (PR)} $\uparrow$ &                      \textsc{AUC (ROC)} $\uparrow$ &                       \textsc{FPR@95} $\downarrow$ &                       \textsc{AUC (PR)} $\uparrow$ &                      \textsc{AUC (ROC)} $\uparrow$ &                       \textsc{FPR@95} $\downarrow$ \\
\midrule
\multirow{2}{*}{L/16} & V-MoE &          1 &          1 &  \textbf{0.9454} {\tiny \color{gray} $\pm$ 0.0013} &  \textbf{0.9481} {\tiny \color{gray} $\pm$ 0.0015} &  \textbf{0.2682} {\tiny \color{gray} $\pm$ 0.0073} &  \textbf{0.9976} {\tiny \color{gray} $\pm$ 0.0000} &  \textbf{0.9882} {\tiny \color{gray} $\pm$ 0.0002} &  \textbf{0.0658} {\tiny \color{gray} $\pm$ 0.0024} &  \textbf{0.7631} {\tiny \color{gray} $\pm$ 0.0043} &  \textbf{0.9667} {\tiny \color{gray} $\pm$ 0.0007} &  \textbf{0.2141} {\tiny \color{gray} $\pm$ 0.0040} &  \textbf{0.9115} {\tiny \color{gray} $\pm$ 0.0035} &  \textbf{0.9533} {\tiny \color{gray} $\pm$ 0.0020} &  \textbf{0.2794} {\tiny \color{gray} $\pm$ 0.0123} \\
\bottomrule
\end{tabular}
}
\end{table}

\begin{table}[]
    \centering
    \caption{Few-shot comparison of \ack, V-MoE and ensembles thereof.}
    \label{tab:fewshot_pbe_vmoe}
    \resizebox{\linewidth}{!}{
\begin{tabular}{ccccccccccccc}
\toprule
     &       & \multicolumn{2}{c}{\;} & \multicolumn{4}{c}{\textsc{Mean Across Datasets}} & \multicolumn{4}{c}{\textsc{Weighted Mean Across Datasets}} &                      \textsc{Downstream} \\
     &       & \textsc{K} & \textsc{M} &              \textsc{1-Shot Error} $\downarrow$ &              \textsc{5-Shot Error} $\downarrow$ &             \textsc{10-Shot Error} $\downarrow$ &             \textsc{25-Shot Error} $\downarrow$ &              \textsc{1-Shot Error} $\downarrow$ &              \textsc{5-Shot Error} $\downarrow$ &             \textsc{10-Shot Error} $\downarrow$ &             \textsc{25-Shot Error} $\downarrow$ & $\Delta$\textsc{FLOPs} (\%) $\downarrow$ \\
\midrule
\multirow{3}{*}{H/14} & E$^3$ &          1 &          2 &  \textbf{31.47} {\tiny \color{gray} $\pm$ 0.72} &  \textbf{17.87} {\tiny \color{gray} $\pm$ 0.32} &  \textbf{14.29} {\tiny \color{gray} $\pm$ 0.22} &  \textbf{10.64} {\tiny \color{gray} $\pm$ 0.25} &  \textbf{78.51} {\tiny \color{gray} $\pm$ 0.24} &  \textbf{80.65} {\tiny \color{gray} $\pm$ 0.08} &  \textbf{81.28} {\tiny \color{gray} $\pm$ 0.05} &  \textbf{81.51} {\tiny \color{gray} $\pm$ 0.06} &                           \textbf{15.63} \\
     & down-DE &          1 &          2 &  \textbf{30.84} {\tiny \color{gray} $\pm$ 0.42} &  \textbf{17.77} {\tiny \color{gray} $\pm$ 0.51} &  \textbf{14.43} {\tiny \color{gray} $\pm$ 0.03} &  \textbf{11.06} {\tiny \color{gray} $\pm$ 0.40} &  \textbf{78.36} {\tiny \color{gray} $\pm$ 0.20} &  \textbf{80.63} {\tiny \color{gray} $\pm$ 0.14} &  \textbf{81.31} {\tiny \color{gray} $\pm$ 0.02} &  \textbf{81.60} {\tiny \color{gray} $\pm$ 0.09} &                                    75.79 \\
\bottomrule
\end{tabular}
}
\end{table}

\begin{table}[]
    \centering
    \caption{Few-shot comparison of V-MoE and ViT.}
    \label{tab:fewshot_vmoe_vs_vit}
    \resizebox{\linewidth}{!}{
\begin{tabular}{cccccccccccc}
\toprule
     &     & \multicolumn{2}{c}{\;} & \multicolumn{4}{c}{\textsc{Mean Across Datasets}} & \multicolumn{4}{c}{\textsc{Weighted Mean Across Datasets}} \\
     &     & \textsc{K} & \textsc{M} &              \textsc{1-Shot Error} $\downarrow$ &              \textsc{5-Shot Error} $\downarrow$ &             \textsc{10-Shot Error} $\downarrow$ &             \textsc{25-Shot Error} $\downarrow$ &              \textsc{1-Shot Error} $\downarrow$ &              \textsc{5-Shot Error} $\downarrow$ &             \textsc{10-Shot Error} $\downarrow$ &             \textsc{25-Shot Error} $\downarrow$ \\
\midrule
\multirow{2}{*}{H/14} & V-MoE &          1 &          1 &  \textbf{33.04} {\tiny \color{gray} $\pm$ 0.32} &  \textbf{19.23} {\tiny \color{gray} $\pm$ 0.35} &  \textbf{15.63} {\tiny \color{gray} $\pm$ 0.25} &  \textbf{12.06} {\tiny \color{gray} $\pm$ 0.36} &  \textbf{79.05} {\tiny \color{gray} $\pm$ 0.12} &  \textbf{81.00} {\tiny \color{gray} $\pm$ 0.09} &  \textbf{81.59} {\tiny \color{gray} $\pm$ 0.06} &  \textbf{81.82} {\tiny \color{gray} $\pm$ 0.08} \\
     & ViT &          - &          1 &           35.78 {\tiny \color{gray} $\pm$ 0.41} &           20.61 {\tiny \color{gray} $\pm$ 0.15} &           16.78 {\tiny \color{gray} $\pm$ 0.12} &           13.62 {\tiny \color{gray} $\pm$ 0.24} &           79.97 {\tiny \color{gray} $\pm$ 0.16} &           81.37 {\tiny \color{gray} $\pm$ 0.04} &           81.87 {\tiny \color{gray} $\pm$ 0.03} &           82.16 {\tiny \color{gray} $\pm$ 0.05} \\
\cmidrule{1-12}
\multirow{2}{*}{L/16} & V-MoE &          1 &          1 &  \textbf{34.15} {\tiny \color{gray} $\pm$ 0.27} &  \textbf{20.32} {\tiny \color{gray} $\pm$ 0.12} &  \textbf{16.14} {\tiny \color{gray} $\pm$ 0.12} &  \textbf{12.71} {\tiny \color{gray} $\pm$ 0.15} &  \textbf{79.54} {\tiny \color{gray} $\pm$ 0.08} &  \textbf{81.27} {\tiny \color{gray} $\pm$ 0.03} &  \textbf{81.71} {\tiny \color{gray} $\pm$ 0.03} &  \textbf{81.97} {\tiny \color{gray} $\pm$ 0.03} \\
\bottomrule
\end{tabular}
}
\end{table}
\end{landscape}

\section{Additional Algorithm Overview Diagrams} \label{sec:additional_overview_diagrams}

\begin{figure}[htbp]
    \centering
    \resizebox{\textwidth}{!}{
    %
    }
    \caption{End-to-end overview of the \emph{Partitioning} method, from \cref{sec:partitioning_without_tiling}, with $E\!=\!6$ experts, partitioned into $M\!=\!2$ groups, with sparsity of $K\!=\!2$, and a ``last-2'' configuration. \textbf{Top}: \emph{Partitioning} contains a sequence of transformer blocks, followed by alternating transformer and p(artitioned)-MoE blocks. As in ViT, images are split into patches whose embeddings are processed by each block. Here, we show 1 embedding for each of three images (\patchmarker[color1], \patchmarker[color2], \patchmarker[color3]). \textcolor{mplred}{\textbf{Bottom left}}: In a MoE block, we replace the transformer block's MLP with parallel partitioned expert MLPs.
    The effect of the routing weights is not depicted. \textcolor{mplpurple}{\textbf{Bottom right}}: The classifier makes predictions from the final representations (\doublepatchmarker). Notice that without a tiling mechanism, there is only a single prediction per input.}
    \label{fig:partitioning_only_flow}
\end{figure}

\begin{figure}[htbp]
    \centering
    \resizebox{\textwidth}{!}{
    \begin{tikzpicture}[
        every node/.append style={font=\tiny}
    ]
    \input{tikz/tiling_only.tikz}
    \end{tikzpicture}
    }
    \caption{End-to-end overview of the \emph{Tiling} method, from \cref{sec:tiling_without_partitioning}, with $E\!=\!6$ experts, a sparsity of $K\!=\!2$, and a ``last-2'' configuration. \textbf{Top}: \emph{Tiling} contains a sequence of transformer blocks, followed by alternating transformer and MoE blocks. As in ViT, images are split into patches whose embeddings are processed by each block. Here, we show 1 embedding for each of three images (\patchmarker[color1], \patchmarker[color2], \patchmarker[color3]). \textcolor{mplred}{\textbf{Bottom left}}: In a MoE block, we replace the transformer block's MLP with parallel partitioned expert MLPs.
    The effect of the routing weights is not depicted.
    Embeddings are tiled (\tiledpatchmarker) in the first p-MoE block only. \textcolor{mplpurple}{\textbf{Bottom right}}: The classifier averages predictions from the final tiled representations (\doublepatchmarker). Notice that without partitioning, some patches and their corresponding tiled versions can be routed to the same experts, resulting in a reduced diversity in predictions.}
    \label{fig:tiling_only_flow}
\end{figure}

\begin{figure}[htbp]
    \centering
    \resizebox{\textwidth}{!}{
    \begin{tikzpicture}[
        every node/.append style={font=\tiny}
    ]
    \input{tikz/mulit-pred_diagram.tikz}
    \end{tikzpicture}
    }
    \caption{End-to-end overview of the simple \emph{Multi-pred} MoE, from \cref{sec:multipred_ablation}, with $E\!=\!3$ experts, sparsity of $K\!=\!2$, and a ``last-2'' configuration. \textbf{Top}: The multi-pred MoE contains a sequence of transformer blocks, followed by alternating transformer and MoE blocks. As in ViT, images are split into patches whose embeddings are processed by each block. Here, we show 1 embedding for each of three images (\patchmarker[color1], \patchmarker[color2], \patchmarker[color3]). \textcolor{mplred}{\textbf{Bottom left}}: In all but the last MoE block, we recombine the predictions as usual.
   \textcolor{mplpurple}{\textbf{Bottom right}}: The classifier averages predictions from the final representations (\doublepatchmarker).}
    \label{fig:naive_mh_flow}
\end{figure}

\end{document}